\documentclass[11pt]{article}

\usepackage{microtype}
\usepackage[margin=1in]{geometry}
\usepackage{graphicx}
\usepackage{subcaption}
\usepackage{booktabs} %
\usepackage{mathpazo}
\usepackage{pifont}
\usepackage{makecell}
\usepackage{tabularx}
\usepackage{pbox}

\usepackage{xcolor}
\usepackage{pgfplots}
\usepackage{pgfplotstable}
\pgfplotsset{compat=1.3}
\usepackage{tikz}
\usetikzlibrary{pgfplots.groupplots}
\usepackage{ragged2e}

\usepackage[toc,page,header]{appendix}
\usepackage{minitoc}

\usepackage[colorlinks=true,linkcolor=blue,urlcolor=blue,citecolor=blue]{hyperref}
\usepackage{xspace}
\usepackage{bm}
\usepackage{comment}
\usepackage{caption}
\usepackage{makecell}

\usepackage[natbib=true,style=alphabetic,minalphanames=3,maxbibnames=99,maxcitenames=2]{biblatex}
\newcommand{\trak}{\textsc{trak}\xspace}
\newcommand{\mtfive}{mT5\xspace}
\newcommand{\tracin}{TracIn\xspace}
\newcommand{\tracincp}{TracInCP\xspace}
\newcommand{\clip}{\textsc{CLIP}\xspace}
\newcommand{\bert}{\textsc{BERT}\xspace}
\newcommand{\bertbase}{\textsc{BERT-base}\xspace}
\newcommand{\cifarten}{\textsc{CIFAR-10}\xspace}
\newcommand{\cifartwo}{\textsc{CIFAR-2}\xspace}
\newcommand{\cifar}{\textsc{CIFAR}\xspace}
\newcommand{\qnli}{\textsc{QNLI}\xspace}
\newcommand{\glue}{\textsc{GLUE}\xspace}
\newcommand{\mscoco}{\textsc{MS COCO}\xspace}
\newcommand{\gas}{\textsc{GAS}\xspace}
\newcommand{\modeldiff}{\textsc{ModelDiff}\xspace}
\newcommand{\ftracetrex}{\textsc{Ftrace-TREx}\xspace}
\newcommand{\trex}{\textsc{TREx}\xspace}

\newcommand{\xmark}{\ding{55}}%

\newcommand{\spelledout}{Tracing with the Randomly-projected After Kernel}
\newcommand{\thetastar}{\theta^\star}
\newcommand{\thetahat}{\widehat{\theta}}

\newcommand{\lr}[1]{\left({#1}\right)}

\renewcommand{\hat}[1]{\widehat{#1}}

\newcommand{\modeleval}[2]{f({#1};{#2})}

\newcommand{\approxmodeleval}[2]{\hat{f}({#1};{#2})}

\newcommand{\mask}[1]{\bm{1}_{#1}}

\newcommand{\grad}{g}
\newcommand{\projgrad}{\phi}
\newcommand{\loss}[1]{L(#1; \theta)}
\newcommand{\lossstar}[1]{L(#1; \thetastar)}
\newcommand{\imagemb}[1]{\phi(#1; \theta)}
\newcommand{\textemb}[1]{\psi(#1; \theta)}

\newlength\myindent
\usepackage[normalem]{ulem}
\usepackage{xcolor}

\newcommand{\scrcmd}{\textsuperscript}
\newcommand{\scr}[1]{\scrcmd{#1}}

\usepackage{amsmath}
\usepackage{amssymb}
\usepackage{mathtools}
\usepackage{amsthm}
\usepackage{thmtools}
\usepackage{algorithm}
\usepackage{algpseudocode}
\usepackage[T1]{fontenc}
\usepackage{color}

\usepackage[capitalize,noabbrev]{cleveref}

\definecolor{mydarkblue}{rgb}{0,0.08,0.85}
\definecolor{mylightblue}{rgb}{0.06,0.56,1.0}
\definecolor{mylightorange}{rgb}{1.0,0.62,0.12}
\definecolor{mylightred}{rgb}{0.99,0.00,0.04}
\hypersetup{colorlinks=true,
linkcolor=mydarkblue,
citecolor=mydarkblue,
filecolor=mydarkblue,
urlcolor=mydarkblue,
bookmarksopen=true,
linktoc=page}

\theoremstyle{plain}
\newtheorem{theorem}{Theorem}[section]

\theoremstyle{definition}
\newtheorem{definition}[theorem]{Definition}

\newtheorem{example}[theorem]{Example}
\theoremstyle{remark}

\usepackage{marginnote}

\title{TRAK: Attributing Model Behavior at Scale}
\author{
\normalsize
Sung Min Park\footnote{Equal contribution.},\ \,Kristian Georgiev\footnotemark[1],\ \,Andrew Ilyas\footnotemark[1],\
\,Guillaume Leclerc, Aleksander M\k{a}dry \\
\normalsize MIT\\
\texttt{\{sp765,krisgrg,ailyas,leclerc,madry\}@mit.edu}}

\date{}

\makeatletter
\let\c@figure\c@table
\makeatother
\begin{document}
\setcounter{tocdepth}{2}
\doparttoc %
\renewcommand\ptctitle{}
\faketableofcontents %

\maketitle
\begin{abstract}
  The goal of {\em data attribution} is to trace model predictions back to training data.
Despite a long line of work towards this goal,
existing approaches to data attribution
tend to force users to choose between
computational tractability and efficacy.
That is, computationally tractable methods can struggle with accurately attributing
model predictions in non-convex settings (e.g., in the context of deep neural networks),
while methods that are effective in such regimes require training thousands of
models, which makes them impractical for large models or datasets.

In this work, we introduce \trak (\spelledout),
a data attribution method that is both effective {\em and}
computationally tractable for large-scale, differentiable models.
In particular, by leveraging only a handful of trained models,
\trak can match the performance of attribution methods that require training thousands of
models.
We demonstrate the utility of \trak across various modalities and scales:  image
classifiers trained on ImageNet, vision-language models (\clip), and
language models (\bert and \mtfive).
We provide code for using \trak
(and reproducing our work) at
\url{https://github.com/MadryLab/trak}.

\end{abstract}

\section{Introduction}
Training data is a key driver of model behavior in modern machine learning systems.
Indeed, model errors, biases, and capabilities can all stem from the training data \citep{ilyas2019adversarial,gu2017badnets,geirhos2019imagenet}.
Furthermore, improving the quality of training data generally improves the performance
of the resulting models \citep{huh2016makes,lee2022deduplicating}.
The importance of training data to model behavior has motivated extensive work
on {\em data attribution}, i.e., the task of tracing model predictions back to the
training examples that informed these predictions.
Recent work
demonstrates, in particular, the utility of data attribution methods in applications such as
explaining predictions \citep{koh2017understanding,ilyas2022datamodels},
debugging model behavior \citep{kong2022resolving,shah2022modeldiff},
assigning data valuations \citep{ghorbani2019data,jia2019towards},
detecting poisoned or mislabeled data \citep{lin2022measuring,hammoudeh2022identifying},
and curating data \citep{khanna2019interpreting,liu2021influence,jia2021scalability}.

However, a recurring tradeoff in the space of data attribution methods is that of
{\em computational demand} versus {\em efficacy}.
On the one hand, methods such as
influence approximation \citep{koh2017understanding, schioppa2022scaling}
or gradient agreement scoring \citep{pruthi2020estimating}
are computationally attractive
but can be unreliable
in non-convex settings
\citep{basu2021influence,ilyas2022datamodels,akyurek2022towards}.
On the other hand, sampling-based methods such as empirical influence
functions \citep{feldman2020neural}, Shapley value estimators \citep{ghorbani2019data,jia2019towards} or datamodels \citep{ilyas2022datamodels} are
more successful at accurately attributing predictions to training data
but require training thousands (or tens of thousands) of models
to be effective.
We thus ask:
\begin{center}
    {\em Are there data attribution methods that are both scalable and effective in large-scale non-convex settings?}
\end{center}

\begin{figure}[!htb]
    \centering
    \begin{tikzpicture}

\definecolor{darkgray141160203}{RGB}{141,160,203}
\definecolor{dimgray85}{RGB}{85,85,85}
\definecolor{gainsboro229}{RGB}{229,229,229}
\definecolor{lightgray204}{RGB}{204,204,204}
\definecolor{mediumaquamarine102194165}{RGB}{102,194,165}
\definecolor{orchid231138195}{RGB}{231,138,195}
\definecolor{salmon25214198}{RGB}{252,141,98}
\definecolor{amethyst}{rgb}{0.6, 0.4, 0.8}
\definecolor{bleudefrance}{rgb}{0.19, 0.55, 0.91}
\definecolor{blush}{rgb}{0.87, 0.36, 0.51}
\definecolor{brilliantrose}{rgb}{1.0, 0.33, 0.64}

\begin{groupplot}[
  group style={group size= 2 by 1},
  height={5cm},
  width=.5\linewidth]
\nextgroupplot[
axis background/.style={fill=gainsboro229},
axis line style={white},
legend cell align={left},
legend columns=4,
legend style={
  fill opacity=0.8,
  draw opacity=1,
  text opacity=1,
  at={(-0.2,1.6)},
  anchor=north west,
  draw=gainsboro229,
  fill=gainsboro229,
  /tikz/every even column/.append style={column sep=0.39cm},
  /tikz/every odd column/.append style={column sep=0.1cm}
},
log basis x={10},
tick align=outside,
tick pos=left,
x grid style={white},
align=center,
title={ResNet-9 on CIFAR-10},
xlabel={Computation time (mins) on 1xA100 \\ (\(\displaystyle \leftarrow\) more efficient)},
xmajorgrids,
xminorgrids,
xmin=0.6, xmax=110000,
xmode=log,
xtick style={color=dimgray85},
xtick={1,10,100,1000,10000,100000,1000000},
xticklabels={
  \(\displaystyle {10^{0}}\),
  \(\displaystyle {10^{1}}\),
  \(\displaystyle {10^{2}}\),
  \(\displaystyle {10^{3}}\),
  \(\displaystyle {10^{4}}\),
  \(\displaystyle {10^{5}}\),
  \(\displaystyle {10^{6}}\)
},
y grid style={white},
ylabel={Correlation \\ (more accurate \(\displaystyle \rightarrow\))},
ymajorgrids,
ymin=0, ymax=0.6,
ytick style={color=dimgray85}
]
\addplot [draw=black,
          fill=black,
          fill opacity=0,
          mark options={scale=1.5, line width=1pt},
          mark=star,
          only marks]
plot[error bars/.cd, y dir=both, y explicit]
table[x index={0}, y index={1}, y error index={2}]{%
x  y
1	0.058062	0.003878
5	0.116515	0.003609
20	0.270943	0.004159
100	0.401995	0.003637
1000 0.535    0.0040
};
\addlegendentry{TRAK}
\addplot [draw=mediumaquamarine102194165, fill opacity=0, color=mediumaquamarine102194165, mark options={line width=1pt}, mark=*, only marks]
plot[error bars/.cd, y dir=both, y explicit]
table[x index={0}, y index={1}, y error index={2}]{%
x  y
500	0.059707	0.004506
2500	0.199107	0.003671
10000	0.391663	0.003693
25000	0.481640	0.003031
50000	0.543450	0.002829
};
\addlegendentry{Datamodel [IPE+22]}
\addplot [draw=amethyst, fill opacity=0, color=amethyst, mark options={line width=1pt}, mark=diamond*, only marks]
plot[error bars/.cd, y dir=both, y explicit]
table[x index={0}, y index={1}, y error index={2}]{%
x  y
500	0.048068	0.004253
2500	0.100678	0.004457
10000	0.192958	0.004143
25000	0.284552	0.004109
};
\addlegendentry{Emp. Influence [FZ20]}
\addplot [draw=salmon25214198,
          color=salmon25214198,
          mark=square*,
          fill opacity=0,
          mark options={line width=1pt},
          only marks]
plot[error bars/.cd, y dir=both, y explicit]
table[x index={0}, y index={1}, y error index={2}]{%
x  y
120	0.020197	0.004771
};
\addlegendentry{IF-Arnoldi [SZV+22] }
\addplot [draw=bleudefrance,
          color=bleudefrance,
          mark=square*,
          fill opacity=0,
          mark options={line width=1pt},
          only marks]
plot[error bars/.cd, y dir=both, y explicit]
table[x index={0}, y index={1}, y error index={2}]{%
x  y
25003	0.036603	0.020784
};
\addlegendentry{IF [KL17] }
\addplot [draw=blush,
          color=blush,
          mark=pentagon*,
          fill opacity=0,
          mark options={line width=1pt},
          only marks]
plot[error bars/.cd, y dir=both, y explicit]
table[x index={0}, y index={1}, y error index={2}]{%
x  y
50.	0.029314	0.005953
};
\addlegendentry{Representation Sim.}
\addplot [draw=brilliantrose,
          color=brilliantrose,
          mark=triangle*,
          fill opacity=0,
          mark options={rotate=270, scale=1.5, line width=1pt},
          only marks]
plot[error bars/.cd, y dir=both, y explicit]
table[x index={0}, y index={1}, y error index={2}]{%
x  y
30	0.047173	0.007191
};
\addlegendentry{GAS [HL22]}
\addplot [draw=black, color=orchid231138195, mark options={rotate=90, scale=1.5,line width=1pt}, mark=triangle*, fill opacity=0, only marks]
plot[error bars/.cd, y dir=both, y explicit]
table[x index={0}, y index={1}, y error index={2}]{%
x  y
15 0.055579	0.007301
300	0.055225	0.007017
};
\addlegendentry{TracIn [PLS+20]}
\nextgroupplot[
axis background/.style={fill=gainsboro229},
axis line style={white},
log basis x={10},
tick align=outside,
tick pos=left,
x grid style={white},
align=center,
title={BERT-base on QNLI},
xlabel={Computation time (mins) on 1xA100 \\ (\(\displaystyle \leftarrow\) more efficient)},
xmajorgrids,
xminorgrids,
xmin=10, xmax=210000,
xmode=log,
xtick style={color=dimgray85},
xtick={1,10,100,1000,10000,100000,1000000},
xticklabels={
  \(\displaystyle {10^{0}}\),
  \(\displaystyle {10^{1}}\),
  \(\displaystyle {10^{2}}\),
  \(\displaystyle {10^{3}}\),
  \(\displaystyle {10^{4}}\),
  \(\displaystyle {10^{5}}\),
  \(\displaystyle {10^{6}}\)
},
y grid style={white},
ymajorgrids,
ymin=0, ymax=0.7,
ytick style={color=dimgray85}
]
\addplot [draw=mediumaquamarine102194165, fill opacity=0, color=mediumaquamarine102194165, mark options={line width=1pt}, mark=*, only marks]
plot[error bars/.cd, y dir=both, y explicit]
table[x index={0}, y index={1}, y error index={2}]{%
x  y
43200.0	0.178846	0.033482
85500.0	0.258097	0.033210
175500.0 0.344712	0.031926
};
\addplot [draw=black, color=orchid231138195, mark options={line width=1pt,rotate=90, scale=1.5}, mark=triangle*, only marks]
plot[error bars/.cd, y dir=both, y explicit]
table[x index={0}, y index={1}, y error index={2}]{%
x  y
284	0.072531	0.025567
};
\addplot [draw=amethyst, fill opacity=0, color=amethyst, mark options={line width=1pt}, mark=diamond*, only marks]
plot[error bars/.cd, y dir=both, y explicit]
table[x index={0}, y index={1}, y error index={2}]{%
x  y
43200.0	0.165813	0.041610
85500.0	0.202983	0.046267
175500.0	0.225179	0.044004
};
\addplot [draw=black,
          color=black,
          fill opacity=0,
          mark options={scale=1.5, line width=1pt},
          mark=star,
          only marks]
plot[error bars/.cd, y dir=both, y explicit]
table[x index={0}, y index={1}, y error index={2}]{%
x  y
64	0.178444	0.005628
640	0.416264	0.010370
6400	0.593538	0.014348
};
\addplot [draw=bleudefrance,
          color=bleudefrance,
          mark=square*,
          fill opacity=0,
          mark options={line width=1pt},
          only marks]
plot[error bars/.cd, y dir=both, y explicit]
table[x index={0}, y index={1}, y error index={2}]{%
x  y
18042.9		0.113670	0.043225
90214.5		0.155935	0.029792
};
\addplot [draw=blush,
          color=blush,
          mark=pentagon*,
          fill opacity=0,
          mark options={line width=1pt},
          only marks]
plot[error bars/.cd, y dir=both, y explicit]
table[x index={0}, y index={1}, y error index={2}]{%
x  y
90	0.050549	0.047340
180	0.050256	0.048500
};
\addplot [draw=brilliantrose,
          color=brilliantrose,
          mark=triangle*,
          fill opacity=0,
          mark options={line width=1pt, rotate=270, scale=1.5},
          only marks]
plot[error bars/.cd, y dir=both, y explicit]
table[x index={0}, y index={1}, y error index={2}]{%
x  y
284.0	0.077749	0.028504
};
\end{groupplot}
\end{tikzpicture}
\caption{Our data attribution method \trak achieves state-of-the-art tradeoffs between
speed and efficacy.
Here, we benchmark its performance relative to prior methods on
\cifarten-trained ResNet-9 models and \qnli-trained \bertbase models. The $x$-axis indicates
the time (in minutes) it takes to run each method on a single A100 GPU
(see \Cref{app:wall_time} for details).
The $y$-axis indicates the method's efficacy
as measured by its ability to make accurate counterfactual predictions
(see \Cref{def:attr_output} for the precise metric);
error bars indicate 95\% bootstrap confidence intervals.
}
\label{fig:headline}
\end{figure}

To properly answer this question,
we first need a unifying metric for evaluating data attribution methods.
To this end, we adopt the view that a data attribution method is useful
insofar as it can make accurate {\em counterfactual predictions}, i.e.,
answer questions of the form
``what would happen if I trained the model on a given subset $S'$ of my training set?''
This perspective motivates a benchmark---inspired by the datamodeling framework
\citep{ilyas2022datamodels}---that
measures the correlation between true model outputs
and attribution-derived predictions for those outputs.

With this benchmark in hand, in \cref{sec:method} we consider our motivating question and introduce \trak
(\spelledout),
a new data attribution method for parametric, differentiable models.
The key idea behind \trak is to first approximate models with a kernel machine
(e.g., through the empirical neural tangent kernel \citep{jacot2018neural})
and then to leverage our understanding of the resulting kernel domain to derive data attribution scores.

We demonstrate that %
\trak retains the efficacy of sampling-based
attribution methods while being several orders of magnitude cheaper computationally.
For example (Figure \ref{fig:headline}), on \textsc{CIFAR-10} (image classification)
and \textsc{QNLI} (natural language inference),
 \trak can be as
effective as datamodels \citep{ilyas2022datamodels} while being
100-1000x faster to compute.
Furthermore, \trak is as fast as existing gradient-based methods such as
TracIn \citep{pruthi2020estimating} or variations of influence functions \citep{koh2017understanding,schioppa2022scaling},
while being significantly more predictive of model behavior.

As a result, \trak enables us to study the connection between model predictions
and training data in large-scale settings.
For example,
we use \trak to study predictions of ImageNet classifiers (\cref{sec:eval});
to understand the shared image-text embedding space of \clip models \citep{radford2021learning} trained
on \mscoco \citep{lin2014microsoft} (\cref{sec:CLIP});
and to fact-trace language models
(a 300M-parameter \texttt{mT5-small} model \cite{raffel2020exploring,xue2021mt5})
finetuned on \ftracetrex (\cref{subsec:fact_trace}).

\section{Motivation and Setup}
\label{sec:prelim}
We begin with a focus on the supervised learning regime.
We will denote by $S = \{z_1, \ldots, z_n\}$ an
ordered training set of examples, where each $z_i = (x_i, y_i) \in \mathcal{Z}$ is an input-label pair.
We represent machine learning models (implicitly) using a {\em model output function}
$\modeleval{z}{\theta}$, which maps an example of interest $z$ and model parameters
$\theta$
to a real number.
There are a variety of model output functions that one can employ---for example,
the loss $\loss{z}$ of the model on the example $z$ is a natural choice.
Ultimately, though, the appropriate model output function to use
will depend on the setting that we are studying.

Throughout this work, we also assume that models are trained
to minimize the empirical training loss, i.e., that the parameters
of these models are given by
\begin{equation}
  \label{eq:erm}
  \thetastar(S) \coloneqq \arg\min_{\theta} \sum_{z_i \in S} \loss{z_i},
\end{equation}
where, again, $\loss{z_i}$ is the model training loss on example $z_i$.
We write $\thetastar$ as a function of $S$ as we will later consider varying $S$---but when $S$ is clear from the context, we omit it and just write $\thetastar$.

In this paper, our overarching goal is to trace model predictions back to the composition of training data.
This goal---which we refer to as {\em data attribution}---is not new.
Prior work has approached it using methods such as
influence functions and their many variants
\citep{hampel2011robust,koh2017understanding,feldman2020neural,hammoudeh2022identifying};
sampling-based estimators such as Shapley values \citep{lundberg2017unified},
empirical influences \citep{feldman2020neural},
and datamodels \citep{ilyas2022datamodels}; as well as
various other approaches
\citep{yeh2018representer,pruthi2020estimating,hammoudeh2022training}.  Each of
these methods implements a similar interface: given a model
and an output of interest (e.g., loss for a given prediction),
a data attribution method computes a {\em score} for each
training input indicating its importance to the output of interest.
\cref{def:attribution} below makes
this interface precise:

\begin{definition}[\em Data attribution]
  \label{def:attribution}
  Consider an ordered training set of examples $S = \{z_1, \ldots, z_n\}$ and a
  model output function $\modeleval{z}{\theta}$.
  A {\em data attribution method} $\tau(z, S)$
  is a function $\tau: \mathcal{Z} \times \mathcal{Z}^n \to \mathbb{R}^n$ that,
  for any example $z \in \mathcal{Z}$ and a training set $S$,
  assigns a (real-valued) score
  to each training input $z_i \in S$ indicating its importance\footnote{We make the notion of ``importance'' more
  precise later (in Definition \ref{def:attr_output}).}
  to the model output $\modeleval{z}{\thetastar(S)}$.
  When the second argument $S$ is clear from the context, we will omit the second argument and
  simply write $\tau(z)$.
\end{definition}
\begin{example}[\em Influence functions as an attribution method]
  \label{ex:infl_attr}
  An example of a data attribution method is the {\em influence function} approach,
  a concept from robust statistics.
  For a specific model output function $\modeleval{z}{\theta}$ on an example of interest $z$, an influence function assigns a score to
  each training example $z_i$ that approximates the effect on the output $\modeleval{z}{\theta}$ of infinitesimally up-weighting
  that training example. %
  A classic
  result from \citep{cook1982residuals} shows that this score can be computed as
  \[
    \tau_{\text{IF}}(z)_i = \nabla_\theta \modeleval{z}{\thetastar}^\top\cdot H_{\thetastar}^{-1}\cdot \nabla_\theta \lossstar{z_i},
  \]
  where, again, $\thetastar$ are the parameters that minimize the empirical risk, $\lossstar{z_i}$ is the training loss of example $z_i$, and $H_{\thetastar}$ is the Hessian
  $\nabla_\theta^2 \frac{1}{n} \sum_{z_i\in S} \lossstar{z_i}$ of the total training loss.
\end{example}

\paragraph{Evaluating attribution methods.}
Given the variety of existing
data attribution methods, we need a method %
to evaluate them in a consistent way.
One popular approach is to simply manually inspect the
training examples that the method identifies as most important
for a given prediction or set of predictions.
Such manual inspection can be a
useful sanity check, %
but
is
also often subjective and unreliable.
For example, in computer vision,
visual similarity between two images
does not
fully capture
the influence of one on the other in terms of model behavior~\citep{ilyas2022datamodels}.

A more objective alternative is to treat the
scores from a data attribution method as estimates of some
ground-truth parameters---such as leave-one-out influences
\citep{koh2017understanding,basu2021influence,koh2019accuracy}
or Shapley values \citep{lundberg2017unified}---and
then measure the accuracy of these estimates.
This approach to evaluation is not only more quantitative than visual inspection
but also inherits all favorable properties of the ground-truth parameter
being considered (e.g., additivity of Shapley values \citep{shapley1951notes}).
However, getting access to these ground-truth parameters can be prohibitively expensive in large-scale settings.

Finally, yet another possibility is to measure the
utility of data attribution scores for
an auxiliary task such as identifying mislabeled data
\citep{koh2017understanding,hammoudeh2022identifying} or active learning
\citep{jia2021scalability}.
This approach can indeed be a
useful proxy for evaluating data attribution methods, but the resulting
metrics may be too sensitive to the particulars of the
auxiliary task and thus make comparisons across different problems and settings difficult.

\subsection{The linear datamodeling score (LDS)}
\label{sub:lds}
Motivated by the above shortcomings of existing methodologies,
we propose a new metric for evaluating data attribution methods.
At the heart of our metric is the perspective that an effective data attribution
method should be able to make accurate {\em counterfactual predictions} about model outputs.
In other words, if a method can accurately quantify the importance of individual training examples to model outputs,
it should also be
able to predict how model outputs change when the training set is modified
in a particular way.

Inspired by \citet{ilyas2022datamodels},
we cast this counterfactual estimation task as that
of predicting the model output function
$\modeleval{z}{\thetastar(S')}$ given different subsets of the training set $S'$.
More precisely, consider---for a fixed example of interest $z \in \mathcal{Z}$---the model output
$\modeleval{z}{\thetastar(S')}$ arising from training on a subset $S'
\subset S$ of the training set $S$ (see \eqref{eq:erm}).\footnote{In many settings,
the non-determinism of training makes this model output function a random variable,
but we treat it as deterministic to simplify our notation.
We handle non-determinism explicitly in \cref{sec:estimator_algo}.}
Since $z$ is fixed and the learning algorithm $\thetastar(\cdot)$ is fixed,
we can view
this model output as a function of $S'$ alone.
A good data attribution method should help us {predict} the former
from the latter.

To operationalize this idea, we first need a way of converting a given data attribution
method $\tau(\cdot)$ into a counterfactual predictor.
We observe that the vast majority of data attribution methods are {\em additive}---that is,
they define the importance of a group of training examples to
be the sum of the importances of the examples in the group.\footnote{
  Note that this additivity assumption can be {explicit} or {implicit}.
  Shapley values \citep{shapley1951notes} and datamodels \citep{ilyas2022datamodels},
  for example, take additivity as an axiom.
  Meanwhile, attribution methods based on influence functions
  \citep{hampel2011robust,koh2017understanding,koh2019accuracy}
  implicitly use a first-order Taylor approximation of the loss function with respect
  to the vector of training example loss weights, which is precisely equivalent to
  an additivity assumption.
}
Motivated by this observation, we define an attribution method's
{\em prediction} of the model output for a subset $S' \subset S$
as the sum of the corresponding scores:
\begin{definition}[\em Attribution-based output predictions]
  \label{def:attr_output}
  Consider a training set $S$, a model output function $\modeleval{z}{\theta}$,
  and a corresponding data attribution method $\tau$
  (see Definition \ref{def:attribution}).
  The {\em attribution-based output prediction} of the model output
  $\modeleval{z}{\thetastar(S')}$ is defined as
  \begin{equation}
    \label{eq:data_attr_agg}
    g_\tau(z, S'; S) \coloneqq \sum_{i\ :\ z_i \in S'} \tau(z, S)_i = \tau(z, S) \cdot \bm{1}_{S'},
  \end{equation}
  where $\bm{1}_{S'}$ is the {\em indicator vector} of the subset $S'$ of $S$
  (i.e., $(\bm{1}_{S'})_i = \bm{1}\{z_i \in S'\}$).
\end{definition}
Intuitively, \cref{def:attr_output} turns any data attribution
method into a counterfactual predictor.
Specifically, for a given counterfactual training set
$S' \subset S$, the attribution method's prediction is simply the sum of the
scores of the examples contained in $S'$.

Now that we have defined how to derive predictions from an attribution method,
we can evaluate these predictions using the {\em linear datamodeling score},
defined as follows:
\begin{definition}[\em Linear datamodeling score]
  \label{def:datamodeling}
  Consider a training set $S$, a model output function $\modeleval{z}{\theta}$,
  and a corresponding data attribution method $\tau$
  (see Definition \ref{def:attribution}).
  Let $\{S_1, \ldots, S_m: S_i \subset S\}$ be $m$
  randomly sampled subsets of the training set $S$,
  each of size $\alpha \cdot n$
  for some $\alpha \in (0, 1)$.
  The {\em linear datamodeling score} (LDS) of a data attribution $\tau$
  for a specific example $z \in \mathcal{Z}$ is given by
  \[
    LDS(\tau, z) \coloneqq \bm{\rho}(\{\modeleval{z}{\thetastar(S_j)}: j \in [m]\},
    \{g_\tau(z, S_j;S): j \in [m]\}),
  \]
  where $\bm{\rho}$ denotes Spearman rank correlation \citep{spearman1904proof}.
  The attribution method's LDS for an entire test set is then simply the average
  per-example score.
\end{definition}

\noindent Note that the linear datamodeling score defined above
is quantitative,
simple to compute,\footnote{
  In practice, we can estimate the LDS with 100-500 models, as the average rank correlation (over a sufficient number of test examples) converges fairly quickly with sample size.
}
and not tied to a specific task or modality.

\subsection{An oracle for data attribution}
\label{subsec:oracle}
\cref{def:datamodeling} immediately suggests an ``optimal''
approach to data attribution (at least, in terms of optimizing LDS).
This approach simply samples random subsets $\{S_1,\ldots S_m\}$ of the training set;
trains a model on each subset (yielding $\{\thetastar(S_1), \ldots, \thetastar(S_m)\}$);
evaluates each corresponding model output function $\modeleval{z}{\thetastar(S_j)}$;
and then {\em fits} scores $\tau(z)$ that predict $\modeleval{z}{\thetastar(S_i)}$
from the indicator vector $\bm{1}_{S_i}$ using (regularized) empirical risk minimization.
Indeed, \citet{ilyas2022datamodels} take exactly this approach---the resulting
datamodel-based attribution for an example $z$ is then given by
\begin{align}
  \label{eq:testset_datamodels}
  \tau_{\textsc{dm}}(z) \coloneqq \min_{\beta \in \mathbb{R}^n}
  \frac{1}{m} \sum_{i=1}^m \lr{{\beta}^\top \mask{S_i} - \modeleval{z}{\thetastar(S_i)}}^2 + \lambda \|\beta\|_1.
\end{align}
The attributions $\tau_{\textsc{dm}}(z)$ turn out to indeed perform well according
to \cref{def:datamodeling}---that is, they yield counterfactual
predictions that are highly correlated with true model outputs (see \cref{fig:headline}).
Unfortunately, however, estimating accurate linear predictors \eqref{eq:testset_datamodels}
may require tens (or even hundreds) of thousands of samples
$(S_j, \modeleval{z}{\thetastar(S_j)})$.
Since each one of these samples involves training a model from scratch,
this direct estimator can be expensive to compute in large-scale settings.
More generally, this limitation applies to all sampling-based attribution methods, such as empirical influences \citep{feldman2020neural,carlini2022quantifying} and Shapley values \citep{ghorbani2019data,jia2019towards}.

In light of the above, we can view the approach of \citet{ilyas2022datamodels} as an ``oracle''
of sorts---it makes accurate counterfactual predictions
(and as a result has found downstream utility \citep{ilyas2022datamodels,shah2022modeldiff,chang2022careful}),
but is (often prohibitively) costly to compute.

\subsection{Data attribution methods beyond sampling}
How might we be able to circumvent the estimation cost of sampling-based attributions?
Let us start by examining the existing data attribution methods---specifically, the ones that
use only one (or a few) trained models---and evaluate them on our LDS benchmark.

\paragraph{Simulating re-training with influence functions.}
The bottleneck of the ``oracle'' datamodels attribution method \eqref{eq:testset_datamodels}
\citep{ilyas2022datamodels} is
that obtaining each sample $(S_j, \modeleval{z}{S_j})$ requires re-training our
model of interest from scratch on each subset $S_j$.
An alternative approach could be to {\em simulate} the effect of this re-training by
making some structural assumptions about the model being studied---e.g., that
its loss is locally well-approximated by a quadratic.
This idea has inspired a long line of work around {\em influence function estimation}
\citep{koh2017understanding,pruthi2020estimating,schioppa2022scaling}.
The resulting {\em influence function attributions}
(\cref{ex:infl_attr})
accurately approximate linear models and other
simple models, but
can
perform poorly in non-convex settings (e.g., in the context of deep neural networks) \citep{basu2021influence,ilyas2022datamodels,bae2022if}.
Indeed, as we can see in Figure~\ref{fig:headline}
(and as we later study in \cref{sec:eval}), estimators based on
influence functions \citep{koh2017understanding,schioppa2022scaling,hammoudeh2022identifying}
significantly underperform on our LDS benchmark (\cref{def:datamodeling}) when evaluated on neural networks on standard vision and natural language tasks.

\paragraph{Heuristic measures of example importance.}
There are also approaches that use more heuristic measures of
training example importance for data attribution.
These include methods based on, e.g.,
representation space similarity
\citep{zhang2018unreasonable,hanawa2021evaluation}
or gradient agreement
\citep{hammoudeh2022identifying}.
While such methods often yield qualitatively compelling
results, our experiments (again, see Figure~\ref{fig:headline})
indicate that, similarly to influence-based estimators,
they are unable to make meaningful counterfactual predictions
about model outputs in the large-scale, non-convex
settings we evaluate them on.

\section{\trak: Tracing with the Randomly-Projected After Kernel}
\label{sec:method}
We now present \trak,
a new data attribution method
which is designed to be both effective and scalable in large-scale differentiable settings.
(Recall from \cref{def:attribution} that a data attribution function is a function
mapping examples $z$ to a vector of per-training example scores in $\mathbb{R}^n$.)

As a warm-up,
and to illustrate the core primitive behind \trak,
we first study the simple case of logistic regression (\cref{ssec:logistic}).
In this setting, data attribution is well-understood---in particular, there is
a canonical attribution method \citep{pregibon1981logistic}
that is both easy-to-compute and highly effective
\citep{wojnowicz2016influence,koh2019accuracy}.
In \cref{sec:estimator_algo}, using this canonical attribution method
as a primitive, we derive our data attribution method
$\tau_\trak(\cdot)$
(Equation \eqref{eq:trak}, also summarized in \cref{alg:estimator_pseudo}
in \cref{sec:pseudocode}) which operates by reducing
complex models back to the logistic regression case.\footnote{
    Note that we focus on logistic regression for simplicity---more generally
    one can adapt \trak to any setting where the training loss is convex
    in the model output; see \Cref{app:general_model_output}.
}

\subsection{Warmup: Data attribution for logistic regression}
\label{ssec:logistic}
Consider the case where the model being studied is
(a generalized form of)
binary logistic regression.
In particular, adapting our notation from \cref{sec:prelim},
we consider a training set of $n$ examples
$$S = \{z_1, \ldots, z_n: z_i = (x_i \in \mathbb{R}^d,\, b_i \in \mathbb{R},\, y_i \in \{-1, 1\})\},$$
where each example comprises an input $x_i \in \mathbb{R}^d$, a bias $b_i \in \mathbb{R}$,
and a label $y_i \in \{-1, 1\}$.
The final model parameters $\thetastar(S)$ then minimize the log-loss over the training set, i.e.,
\begin{equation}
    \label{eq:binary_loss}
    \thetastar(S) \coloneqq \arg\min_{\theta}
        \sum_{(x_i, y_i) \in S}\log\left[1 + \exp(-y_i \cdot (\theta^\top x_i + b_i))\right].
\end{equation}
(Note that when the bias terms $b_i$ are identically zero, we recover
ordinary logistic regression.)
The natural choice of {\em model output function} in this case is then the ``raw logit'' function:
\begin{equation}
    \label{eq:modelout_logistic}
    \modeleval{z}{\theta} \coloneqq \theta^\top x + b, \qquad \text{ where } z = (x, b, y).
\end{equation}
Data attribution in this simple setting is a well-studied problem.
In particular,
the {\em one-step Newton approximation}
\citep{pregibon1981logistic,wojnowicz2016influence,rad2018scalable,koh2019accuracy},
which we present as a data attribution method $\tau_\text{NS}$ below,
is a standard tool for analyzing and understanding logistic regression models
in terms of their training data.
(We present the theoretical basis for this method in
\cref{app:theory_newton}.)
\begin{definition}[One-step Newton approximation \citep{pregibon1981logistic}]
    \label{lem:formal}
    For logistic regression, we define the Newton step
    data attribution method $\tau_\text{NS}$ as the
    approximate leave-one-out influence \citep{pregibon1981logistic}
    of training examples $z_i = (x_i, b_i, y_i)$
    on the model output function \eqref{eq:modelout_logistic}. That is,
    \begin{equation}
        \label{eq:loo_infl}
        \tau_\text{NS}(z)_i \coloneqq
        \frac{x^\top (X^\top R X)^{-1} x_i}{1 - x_i^\top (X^\top R X)^{-1} x_i \cdot p_i^\star (1 - p_i^\star)} (1 - p_i^\star)
        \approx \modeleval{z}{\thetastar(S)} - \modeleval{z}{\thetastar(S \setminus z_i)}
    \end{equation}
        where $X \in \mathbb{R}^{n \times k}$ is the matrix of stacked inputs $x_i$,
        $p_i^\star \coloneqq (1 + \exp(-y_i \cdot \modeleval{z_i}{\thetastar}))^{-1}$
        is the
        predicted correct-class probability at $\thetastar$ and
        $R$ is
        a diagonal $n \times n$ matrix
        with $R_{ii} = p_i^\star (1 - p_i^\star)$.
\end{definition}
If our model class of interest was binary logistic regression,
we could simply apply \cref{lem:formal} to perform data attribution.
As we discuss, however, our goal is precisely to scale
data attribution {\em beyond} such convex settings.
To this end, we next derive our data attribution method \trak
(\spelledout)
which leverages $\tau_\text{NS}$ (\cref{lem:formal})
as a primitive.

\subsection{\trak for binary (non-linear) classifiers}
\label{sec:estimator_algo}
We now present our method (\trak) for scaling data attribution to
non-convex differentiable settings.
More precisely, following \cref{def:attribution},
we describe how to compute a function
$\tau_\trak: \mathcal{Z} \to \mathbb{R}^n$
that maps examples of interest $z$ to vectors
of per-training example importance scores in $\mathbb{R}^n$.
The key primitive here will be \cref{lem:formal} from above---in
particular, we will show how to adapt our problem into one to which
we can apply the approximation \eqref{eq:loo_infl}.

For ease of exposition, we will first show how to compute
$\tau_\trak$ in the context of binary
classifiers trained with the negative log-likelihood loss.
We later generalize \trak to other types of models (e.g.,
to multi-class classifiers in \cref{ssec:multiclass},
to contrastive models in \cref{sec:CLIP},
and to language models in \cref{subsec:fact_trace}).
In this setting, let the model output function $\modeleval{z}{\theta}$
be the raw output (i.e., the logit)
of a binary classifier with parameters $\theta$.\footnote{
    Note that for the special case of binary classifiers,
    the model output function that we define (i.e.,
    $\modeleval{z}{\theta} = \modeleval{(x, y)}{\theta}$)
    depends only on the input $x$, and not on the label $y$.
    When we generalize \trak to more complex losses in
    \cref{ssec:multiclass},
    the model output function will involve both $x$ and $y$.
}
The final parameters of the model can thus be written as
\begin{equation}
    \label{eq:loss_min}
    \thetastar(S) = \arg\min_\theta \sum_{(x_i, y_i) \in S} \log\left[
        1 + \exp\lr{-y_i \cdot \modeleval{z_i}{\theta}}
    \right].
\end{equation}
Note that unlike in \cref{ssec:logistic}, we do not assume that the model itself is
linear---e.g., the model might be a deep neural network parameterized by
weights $\theta$.

We implement \trak as a sequence of five steps:
\begin{enumerate}
    \item Linearizing the model output function (via Taylor approximation),
    which reduces the model of interest to a linear function in parameter space.
    Prior work (around, e.g., the empirical neural tangent kernel)
    suggests that this approximation
    can be relatively accurate, especially for overparameterized neural networks
    \citep{jacot2018neural,wei2022more,long2021properties,malladi2022kernel}.
    \item Reducing the dimensionality of
    the linearized model using random projections.
    Specifically, we take advantage of the Johnson-Lindenstrauss lemma
    \citep{johnson1984extensions},
    which guarantees that this projection preserves the model-relevant information.
    \item Estimating attribution scores by leveraging the attribution method
    described in \cref{lem:formal}.
    \item Ensembling results over several
    models, each trained on a random subset of the original training
    set $S$.
    \item Sparsifying the attribution scores using soft-thresholding.
\end{enumerate}

\noindent We discuss these steps in more depth below.

\paragraph{(Step 1) Linearizing the model.}
Recall that our goal here is to apply the data attribution method $\tau_\text{NS}$
from \cref{lem:formal}.
The main roadblock to applying \cref{lem:formal} in our setting
is that we are studying
a {\em non-linear} model---that is, our model output function may not be a
linear function of $\theta$.
We address this issue by approximating $\modeleval{z}{\theta}$
with its Taylor expansion centered
around the final model parameters $\thetastar$.
In particular, for any $\theta$,
we replace $\modeleval{z}{\theta}$ with
\begin{align}
    \label{eq:taylor}
    \approxmodeleval{z}{\theta} &\coloneqq
    \modeleval{z}{\thetastar} +
    \nabla_\theta \modeleval{z}{\thetastar}^\top (\theta - \thetastar).
\end{align}
This approximation
suggests a change in perspective---rather than viewing
$\modeleval{z}{\theta}$ as a non-linear model acting on
inputs $x$, we can view it as a {\em linear} model
acting on inputs $\nabla_\theta \modeleval{z}{\thetastar}$.
In particular,
rewriting the loss minimization \eqref{eq:loss_min} while
replacing $\modeleval{z}{\theta}$ with $\approxmodeleval{z}{\theta}$
yields
\begin{equation}
    \label{eq:rewritten_loss_min}
    \thetastar(S) =
    \arg\min_\theta \sum_{(x_i, y_i) \in S} \log\left[
        1 + \exp\lr{-y_i \cdot
        \lr{
            \theta^\top \nabla_\theta \modeleval{z_i}{\thetastar}
            + \modeleval{z_i}{\thetastar} - \nabla_\theta \modeleval{z_i}{\thetastar}^\top \thetastar
        }}\right].
\end{equation}
Now, Equation \eqref{eq:rewritten_loss_min} should look familiar---specifically,
if we define the variables $g_i \coloneqq \nabla_\theta \modeleval{z_i}{\thetastar}$
and $b_i \coloneqq \modeleval{z_i}{\thetastar} - \nabla_\theta\modeleval{z_i}{\thetastar}^\top \thetastar$,
then \eqref{eq:rewritten_loss_min} becomes
\begin{equation}
    \label{eq:rewritten_loss_min_2}
    \thetastar(S) =
    \arg\min_\theta \sum_{(g_i, b_i, y_i)} \log\left[
        1 + \exp\lr{-y_i \cdot
        \lr{ \theta^\top g_i + b_i }}\right].
\end{equation}
Comparing \eqref{eq:rewritten_loss_min_2}
to \eqref{eq:binary_loss} (from \cref{ssec:logistic})
makes it clear that we can view
$\thetastar$ as the solution to a (generalized) logistic regression,
in which the inputs $x_i$ are gradients
$g_i \coloneqq \nabla_\theta \modeleval{z_i}{\thetastar}$ of the
model,
the bias terms are
$b_i \coloneqq \modeleval{z_i}{\thetastar} - \nabla_\theta\modeleval{z_i}{\thetastar}^\top \thetastar$
 and the labels $y_i$ remain the same.

\medskip \noindent {\em \textbf{Note}:}
In the context of neural networks, we can view
Step 1 as replacing the binary classifier with its empirical neural tangent kernel
(eNTK) approximation \citep{jacot2018neural,atanasov2022neural,wei2022more}.
We discuss how \trak connects to the eNTK in more detail in \Cref{sec:related}.

\paragraph{(Step 2) Reducing dimensionality with random projections.}
The linear approximation from Step~1
dramatically simplifies our model class of interest
from a highly non-linear classifier to simple logistic regression.
Still,
the resulting logistic regression can be extremely high dimensional.
In particular, the input dimension of the linear model \eqref{eq:taylor}
is the number of {parameters} of the original model
(which can be on the order of millions or billions),
not the dimensionality of the inputs $x_i$.

To reduce the dimensionality of this problem,
we leverage a classic result of \citet{johnson1984extensions}.
This result guarantees that multiplying each
gradient $\grad_i = \nabla_\theta \modeleval{z_i}{\thetastar} \in \mathbb{R}^p$
by a random matrix $\mathbf{P} \sim \mathcal{N}(0, 1)^{p \times k}$
for $k \ll p$
preserves inner products $g_i^\top g_j$ with high probability\footnote{
In Appendix \ref{app:theory_jl} we discuss why preserving inner products
suffices to preserve the structure of the logistic regression.}
(while significantly reducing the dimension).
Thus,
we define the ``feature map''
$\phi: \mathcal{Z} \to \mathbb{R}^k$ as
\begin{equation}
    \label{eq:featurizer}
    \phi(z) \coloneqq \mathbf{P}^\top \nabla_\theta \modeleval{z}{\thetastar},
\end{equation}
i.e., a function taking an example $z$ to its corresponding projected gradient,
and
from now on
replace $\grad_i$ with
\begin{equation}
    \label{eq:projgrad}
    \projgrad_i \coloneqq \phi(z_i) = \mathbf{P}^\top \grad_i = \mathbf{P}^\top \nabla_\theta \modeleval{z_i}{\thetastar}.
\end{equation}

\paragraph{(Step 3) Estimating influences.}
Now that we have simplified our original model of interest
to a logistic regression problem of tractable dimension,
we can finally adapt \cref{lem:formal}.

To this end, recall that the training ``inputs'' are now the
(projected) gradients $\projgrad_i$ (see \eqref{eq:projgrad}).
We thus replace the matrix $X$ in \eqref{eq:loo_infl} with the matrix
$\Phi \coloneqq [\phi_1; \ldots, \phi_n] \in \mathbb{R}^{n \times k}$ of stacked projected gradients.
We also find empirically that both the denominator in \eqref{eq:loo_infl}
and the diagonal matrix $R$ have little effect
on the resulting estimates, and so we omit them from our adapted estimator.
Our estimator for attribution scores for an example of interest $z$ thus becomes:
\begin{equation}
    \label{eq:single_model_trak}
    \tau(z, S) \coloneqq \phi(z)^\top (\Phi^\top \Phi)^{-1} \Phi^\top \mathbf{Q},
\end{equation}
where we recall from \eqref{eq:featurizer} that $\phi(z) = \mathbf{P}^\top \nabla_\theta \modeleval{z}{\thetastar}$,
and where we define
\begin{equation}
    \label{eq:q_mat}
    \mathbf{Q}
    \coloneqq \text{diag}(\{1-p_i^\star\})
    = \text{diag}\left(
        \left\{(1 + \exp(y_i \cdot \modeleval{z_i}{\thetastar}))^{-1}\right\}\right)
\end{equation}
to be the $n \times n$ diagonal matrix of ``one minus correct-class
probability'' terms.\footnote{
    Note that in our linearization \eqref{eq:rewritten_loss_min_2},
    the predicted probability is also a function of the bias
    terms $b_i$. We can avoid
    having to compute these bias terms by simply using the
    predicted probability from the true model
    (i.e., the neural network) instead of the linearized one.
}

\medskip \noindent {\em \textbf{Remark.}} An alternative way to motivate our single-model estimator (\Cref{eq:single_model_trak})
is to compute the influence function \cite{koh2017understanding} using the generalized Gauss-Newton approximation to the Hessian  \cite{sagun2017empirical,martens2020new,bae2022if}. As noted in prior works \cite{teso2021interactive,bae2022if}, this approximation is a more convenient choice than the full Hessian as it is guaranteed to be positive semi-definite.

\paragraph{(Step 4) Ensembling over independently trained models.}
So far, our analysis ignores the fact that in many modern settings,
training is {non-deterministic}.
That is, applying the same learning algorithm to the same training dataset
(i.e., changing only the random seed)
can yield models with (often significantly) differing behavior
\citep{nguyen2021wide,damour2020underspecification}.
Non-determinism poses a problem for data attribution because
by definition, we cannot explain such seed-based differences in
terms of the training data.

To ``smooth out'' the impact of such seed-based differences,
we aggregate the estimator \eqref{eq:single_model_trak}
across multiple trained models
(for computational efficiency, one can also use different checkpoints from
the same model---see \Cref{app:proxies}).
In particular, we adopt the natural idea of just averaging $\tau(z, S)$ from
\eqref{eq:single_model_trak} directly, with two small modifications:
\begin{itemize}
    \item[(a)]
    Rather than computing $M$ copies of \eqref{eq:single_model_trak}
    and averaging the results, we separately compute and average $M$ copies of
    $\mathbf{Q}$ (i.e., \eqref{eq:q_mat})
    and $M$ copies of $\phi(z)^\top (\Phi^\top \Phi)^{-1} \Phi^\top$
    (i.e., the remaining terms in \eqref{eq:single_model_trak}).
    We then take the product of these averaged matrices.
    \item[(b)]
    Rather than training $M$ copies of the same model $\thetastar(S)$,
    we sample $M$ random subsets of $S$ ($S_1,\ldots,S_M$),
    and use the resulting models $\thetastar(S_1),\ldots,\thetastar(S_M)$
    to compute attribution scores.
\end{itemize}
The first modification (a) is mainly for numerical stability, while the second
modification (b) is meant to better handle duplicated training examples (and,
more generally, features that are highly ``redundant'' in the training data). We
study the effect of these modifications empirically in \cref{app:more_ablation}.
At this point, our estimator is of the form:
\begin{equation}
    \label{eq:multi_model_trak}
    \tau_M(z, S) \coloneqq
    \left(\frac{1}{M} \sum_{i=1}^M \mathbf{Q}_m \right) \cdot
    \left(\frac{1}{M} \sum_{i=1}^M \phi_m(z)^\top (\Phi_m^\top \Phi_m)^{-1} \Phi_m^\top \right),
\end{equation}
where $S_1,\ldots,S_M$ are $M$ randomly selected subsets of the training set $S$;
$\Phi_m$ are the corresponding projected gradients from the model
$\thetastar(S_m)$;
$\phi_m(z)$ is the featurized example $z$ under model $\thetastar(S_m)$;
and $\mathbf{Q}_m$ is the corresponding matrix of probabilities as defined in \Cref{eq:q_mat}.

\paragraph{(Step 5) Inducing sparsity via soft-thresholding.}
In the last step, we post-process the attribution scores from Step 4 via
{\em soft thresholding},
a common denoising method in statistics \cite{donoho1995noising}
for when an underlying signal is known to be sparse.
Within our particular context, \citet{ilyas2022datamodels} find that
for neural networks
attribution scores are often sparse---that is,
each test example depends on only a few examples from the training set.
Motivated by this observation, we apply the soft thresholding operator
$S(\cdot; \lambda)$
defined for any $\tau \in \mathbb{R}^n$ as:
\begin{equation}
    \mathcal{S}(\tau; \lambda) =
        (\tau_i - \lambda) \cdot \bm{1}\{\tau_i > \lambda\}
        + (\tau_i + \lambda) \cdot \bm{1}\{\tau_i < -\lambda\}.
      \label{eqn:soft}
\end{equation}
We choose the soft threshold parameter $\lambda$ via cross-validation.
That is, given a set of trained models,
we first estimate attribution scores \eqref{eq:multi_model_trak},
then sample a range of values for $\lambda$,
compute corresponding attribution scores by applying \eqref{eqn:soft},
and finally
select the value of $\lambda$
that yields that highest linear datamodeling score (\Cref{def:datamodeling}) on
the set of trained models.
After applying soft-thresholding,
our final estimator becomes
\begin{equation}\label{eq:trak}
    \tau_\trak(z, S) \coloneqq
    \mathcal{S}\lr{
    \left(\frac{1}{M} \sum_{i=1}^M \mathbf{Q}_m \right) \cdot
    \left(\frac{1}{M} \sum_{i=1}^M \phi_m(z)^\top (\Phi_m^\top \Phi_m)^{-1} \Phi_m^\top \right),
    \widehat{\lambda}
    }
\end{equation}
where, again, $\widehat{\lambda}$ is selected via cross-validation (see \Cref{app:trak_hparams} for details).

\subsection{Extending to multi-class classification}
\label{ssec:multiclass}
In the previous section, we instantiated \trak for binary classifiers;
we now show how to extend \trak to the multi-class setting.
Recall that our key insight in the binary case was to
linearize the model output function $\modeleval{z}{\theta}$ around the optimal parameters
$\thetastar(S)$ (see \eqref{eq:taylor}).
Our choice of output function
(i.e., the raw logit of the classifier)
allowed us to then cast the original (non-convex) learning problem of interest
as an instance of binary logistic regression with inputs $\nabla_\theta
\modeleval{z}{\thetastar}$. That is, we made the approximation
\begin{equation}
    \label{eq:binary_extending}
    \thetastar(S) \approx \arg\min_\theta \sum_{z_i \in S} \log\left[1 + \exp\lr{-y_i \cdot
    \lr{
        \nabla_\theta \modeleval{z_i}{\thetastar}^\top \theta
        + b_i
    }
    }\right],
\end{equation}
and then leveraged \cref{lem:formal}.

To apply this same approach to the $c$-class setting (for $c > 2$), one possibility is
to first transform the problem into $\smash{c^2}$ binary classification problems,
then apply the approach from \cref{sec:estimator_algo} directly.
(For example, \citet{malladi2022kernel} use this transformation to apply the neural
tangent kernel to $c$-way classification problems.)
In large-scale settings, however, it is often expensive or infeasible to study
of all $c^2$ subproblems, e.g., ImageNet has $c = 1000$ classes.

We thus take a different approach.
In short, we leverage the fact that we always have labels available
(even for test examples)
to reduce the multi-class classification problem to a {\em single} logistic regression.
More specifically, for an example $z = (x, y)$, we define the model output function
\begin{equation}
    \label{eq:modelout_mc}
    \modeleval{z}{\theta} \coloneqq \log\left(\frac{p(z; \theta)}{1 - p(z; \theta)}\right),
\end{equation}
where $p(z;\theta)$ is the softmax probability assigned to the {\em correct} class.

A crucial property of the model output function \eqref{eq:modelout_mc} is that
it allows us to rewrite the loss function for $c$-way classification as
\begin{align}
    \loss{z} &= -\log(p(z;\theta)) \\
             &= \log\left[1 + \exp\lr{-\modeleval{z}{\theta}}\right],
    \label{eq:multi_to_binary}
\end{align}
where the first line is the definition of cross-entropy loss,
and the second line comes from \eqref{eq:modelout_mc}.
As a result,
if we linearize $\modeleval{z}{\theta}$ as in
Step 1 above (\cref{sec:estimator_algo}),
we can make the approximation
\[
    \thetastar(S) \approx \arg\min_\theta \sum_{z_i \in S} \log\left[1 + \exp\lr{- \nabla_\theta \modeleval{z_i}{\thetastar}^\top \theta + b_i} \right].
\]
This approximation is identical to the one we made
for the binary case (see \eqref{eq:binary_extending}).
We can thus treat the multi-class problem as a single binary logistic regression
with inputs $\nabla_\theta \modeleval{z_i}{\thetastar}$\footnote{
    Note that the corresponding ``labels'' for this logistic regression
    are actually identically equal to one---to see this, compare
    \eqref{eq:multi_to_binary} to \eqref{eq:binary_extending}.
    This does not change the resulting attributions, however, as
    \cref{lem:formal} only depends on labels through its
    dependence on the correct-class probability $p_i^*$.
}
and then apply Steps 2-5 from
\cref{sec:estimator_algo} directly to this binary problem.

\subsection{Implementing \trak}
\label{sec:pseudocode}
We summarize our final algorithm for computing the data attribution method
$\tau_\trak$ in the general multi-class case (see also
\cref{eq:trak}) in \cref{alg:estimator_pseudo}.
The output of the algorithm is an attribution matrix $\mathbf{T}$, whose rows are given by $\tau_{\trak}(z,S)$.
To make \cref{alg:estimator_pseudo} efficient even for very large models, we implemented a
highly optimized random projector, which we discuss in \cref{app:impl}.
\begin{algorithm}[H]
\caption{\trak for multi-class classifiers (as implemented)}
\label{alg:estimator_pseudo}
\begin{algorithmic}[1]
    \State {\bfseries Input:} Learning algorithm $\mathcal{A}$,
    dataset $S$ of size $n$,
    sampling fraction $\alpha \in (0,1]$,
    correct-class likelihood function $p(z;\theta)$, projection dimension $k \in \mathbb{N}$
    \State {\bfseries Output:} Matrix of attribution scores $\mathbf{T} \in \mathbb{R}^{n \times n}$
    \State $\modeleval{z}{\theta} \coloneqq \log(\frac{p(z; \theta)}{1 - p(z; \theta)})$
    \hfill$\triangleright$ Margin function $f_\theta$
    \For{$m \in \{1,\ldots,M\}$}
    \State Sample random $S' \subset S$ of size  $\alpha \cdot n$
    \State $\thetastar_m \gets \mathcal{A}(S')$ \hfill$\triangleright$ Train a model on $S'$
    \State $\mathbf{P} \sim \mathcal{N}(0, 1)^{p \times k}$ \hfill$\triangleright$ Sample projection matrix
    \State $\mathbf{Q}^{(m)} \gets \bm{0}_{n \times n}$
    \For{$i \in \{1,\ldots,n\}$}
    \State $\phi_i \gets \mathbf{P}^\top\nabla_\theta \modeleval{z_i}{\thetastar_m}$ \hfill$\triangleright$ Compute gradient at $\thetastar_m$ and project to $k$ dimensions
    \State $\mathbf{Q}^{(m)}_{ii} \gets 1 - p(z_i; \thetastar)$ \hfill$\triangleright$ Compute weighting term
    \EndFor
    \State $\Phi_m \gets [{\phi}_1; \cdots; {\phi}_n]^\top$
    \EndFor
    \State
    $\mathbf{T} \gets \left[\frac{1}{m}\sum\limits_{m=1}^M
    \Phi_m(\Phi_m^\top \Phi_m)^{-1} \Phi_m^\top
     \right]\left[\frac{1}{m}\sum\limits_{m=1}^M\mathbf{Q}^{(m)}\right]$
     \State \textbf{return} $\textsc{Soft-Threshold}(\mathbf{T})$
\end{algorithmic}
\end{algorithm}

\section{Evaluating \trak}
\label{sec:eval}
We now evaluate \trak (see \cref{eq:trak} and \cref{alg:estimator_pseudo} in
\cref{sec:pseudocode}) in a variety of vision and natural language settings. To
this end, we compare \trak with existing data attribution methods and show that
it achieves significantly better tradeoffs between efficacy and computational
efficiency.

\subsection{Experimental setup}
We evaluate and study \trak with the following experimental setup.

\vspace*{-1em}
\paragraph{Datasets, models, and baselines.}
We use ResNet-9 classifiers trained on
the \cifar dataset (\cifarten, and a two-class subset called \cifartwo);
ResNet-18 \citep{he2015deep} classifiers trained on the 1000-class
ImageNet \citep{russakovsky2015imagenet} dataset,
and pre-trained \bert \citep{devlin2019bert} models
finetuned on the \qnli (Question-answering Natural
Language Inference) classification task from the \glue benchmark \citep{wang2018glue}.
We provide further details on these choices of dataset and task in \Cref{app:datasets_models}.

To put \trak's performance into context, we also evaluate a variety of existing attribution methods,
including influence functions \citep{koh2017understanding};
a variant based on the Arnoldi iteration \citep{schioppa2022scaling};
\tracin \citep{pruthi2020estimating};
gradient aggregated similarity (\gas) \citep{hammoudeh2022training};
representation similarity \citep{hanawa2021evaluation};
empirical influences \citep{feldman2020neural};
and datamodels \citep{ilyas2022datamodels}.
(See \Cref{app:baselines} for more details.)

\paragraph{Evaluation with linear datamodeling scores.}
For each method and each dataset we consider,
we compute its linear datamodeling score (LDS) as described in \Cref{def:datamodeling}.
Specifically, let $\tau$ be a given data attribution method
(as framed in \cref{def:attribution}),
and let $g_\tau(z, S'; S)$ be its corresponding attribution-derived prediction function
(see \cref{def:attr_output}).
Then, to evaluate $\tau$:

\begin{enumerate}
\item We sample $100$ different random subsets $\{S_j \subset S: j \in [100]\}$
of the training set $S$,
and train five models on each one of these subsets.
Each subset $S_j$ is sampled to be 50\% of the size of $S$,
but we also consider
other subsampling ratios in \Cref{app:more_results}.

\item For each example of interest $z$ (i.e., for each example in the test set
of the dataset we are studying),
we approximate the expectation of the model output
$\mathbb{E}[\modeleval{z}{\thetastar_i(S_j)}]$ for each training subset $S_j$
(where the expectation is taken over the learning algorithm's randomness)
by averaging across the corresponding five models $\{\thetastar_i(S_j)\}_{i=1}^5$.

\item We then compute the linear datamodeling score for each example of interest
$z$ as the Spearman rank correlation
\citep{spearman1904proof} between the averaged model outputs computed in the previous step
and the attribution-derived predictions $g_\tau(z, S_j; S)$ of model outputs.
That is, we compute:
    \[
        \text{Spearman-}\rho\bigg(
            \underbrace{\left\{\frac{1}{5} \sum_{i=1}^5 \modeleval{z}{\thetastar_i(S_j)}: j \in [100] \right\}}_\text{averaged model outputs}, \underbrace{\{g_\tau(z, S_j; S): j \in [100]\}}_{\substack{\text{attributed-derived predictions}\\\text{of model outputs}}}
        \bigg)
    \]

\item Finally, we average the LDS (\cref{def:datamodeling}) across 2,000 examples of interest,
sampled uniformly at random from the validation set,
and report this score along with the $95\%$ bootstrap confidence intervals corresponding to the random
re-sampling from the subsets $S_j$.
\end{enumerate}

\paragraph{Computational cost.}
We quantify the computational cost of each attribution method using two metrics.
The first one is the {\em total wall-time} of computing attribution scores on a single A100 GPU.
This metric is intuitive and useful, but depends on implementation details and hardware.
We thus also study a second metric, namely,
the {\em total number of trained models used}.
This metric is hardware and implementation-agnostic; it is motivated by an
observation that for large models,
the time it takes to compute attribution scores will be dominated by
the time it takes to {train} the models needed for
attribution.\footnote{For many data attribution methods, such as influence function-based
methods or \trak, there is an extra step of computing per-example gradients
through the model of interest.
However, this step is generally fully parallelizable,
and usually bounded by the time it takes to train a model from scratch.}
We find that for both metrics, our results lead to similar conclusions.

\subsection{Results}
Across all models and datasets that we consider, \trak attains a significantly
better tradeoff between efficacy (as measured by the LDS) and computational
efficiency than all the other attribution methods that we examine (see
\Cref{fig:headline,fig:headline_full} and \Cref{tab:all_best}). Indeed, \trak attains
efficacy comparable to datamodels (which achieves the best performance among existing methods when unconstrained)
with a computational footprint that is
(on average) over 100x smaller.

\begin{figure}[!htb]
    \centering
    \input{figures/main_figure_full.tex}
\caption{{\em \trak achieves state-of-the-art tradeoffs between
    attribution efficacy and efficiency.} We use \trak to attribute ResNet-9
    classifiers trained on \cifartwo and \cifarten; ResNet-18 classifiers trained
    on ImageNet; and BERT-base models finetuned on \qnli.  The $x$-axis indicates
    the computational cost measured as the number of trained models that a given
    method uses to compute attribution scores. The $y$-axis indicates the
    method's efficacy as measured by the linear datamodeling score (LDS). Error
    bars indicate 95\% bootstrap confidence intervals.}
\label{fig:headline_full}
\end{figure}

\begin{figure}[!hb]
    \centering
    \begin{tabular}{p{0.21\textwidth}p{0.34\textwidth}p{0.35\textwidth}}
    \toprule
    \textbf{Example} & \textbf{Highest \trak score (+)} & \textbf{Lowest \trak score (-)} \\
    \midrule
    \scriptsize {\bf \scriptsize {\bf Q:}} What genre of music is Lindisfarne classified as? {\bf A:} Lindisfarne are a
    folk-rock group with a strong Tyneside connection. {\bf (Yes)} & \scriptsize {\bf Q:} What
    genre of music is featured at Junk? {\bf A:} The nightclub, Junk, has been
    nominated for the UK's best small nightclub, and plays host to a range of
    dance music's top acts. {\bf (Yes)} & \scriptsize {\bf Q:} Which genre did Madonna started
    out in? {\bf A:} Stephen Thomas Erlewine noted that with her self-titled debut
    album, Madonna began her career as a disco diva, in an era that did not have
    any such divas to speak of. {\bf (No)} \\
    \hline
    \scriptsize {\bf Q:} What can rubisco do by mistake? {\bf A:} It can waste up to half the carbon
    fixed by the Calvin cycle. {\bf (No)} & \scriptsize {\bf Q:} What can clothing provide
    during hazardous activities? {\bf A:} Further, they can provide a hygienic
    barrier, keeping infectious and toxic materials away from the body. {\bf (No)}
    & \scriptsize {\bf Q:} Quantum Dot LEDs can do what special skill? {\bf A:} This allows
    quantum dot LEDs to create almost any color on the CIE diagram. {\bf (Yes)}
    \\
    \bottomrule
\end{tabular}

    \vspace{0.5em}
    \includegraphics[width=\linewidth,trim={0 2.2cm 0 0},clip]{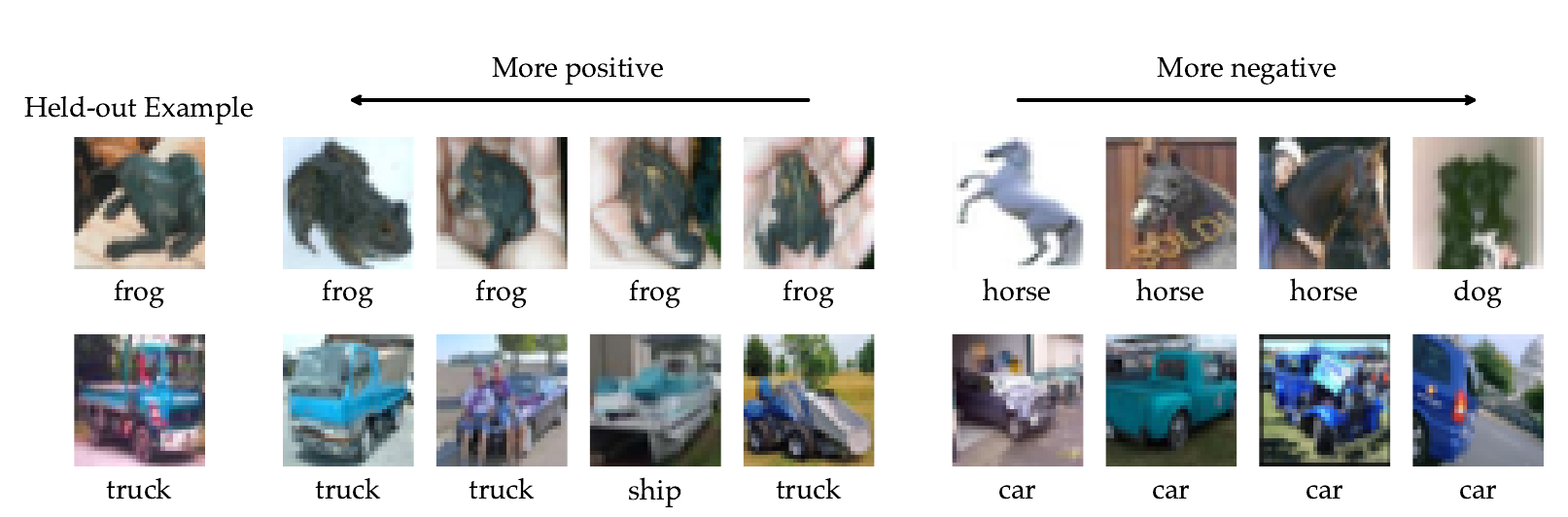}
    \vspace{0.5em}
    \includegraphics[width=\linewidth,trim={0 2.4cm 0 0},clip]{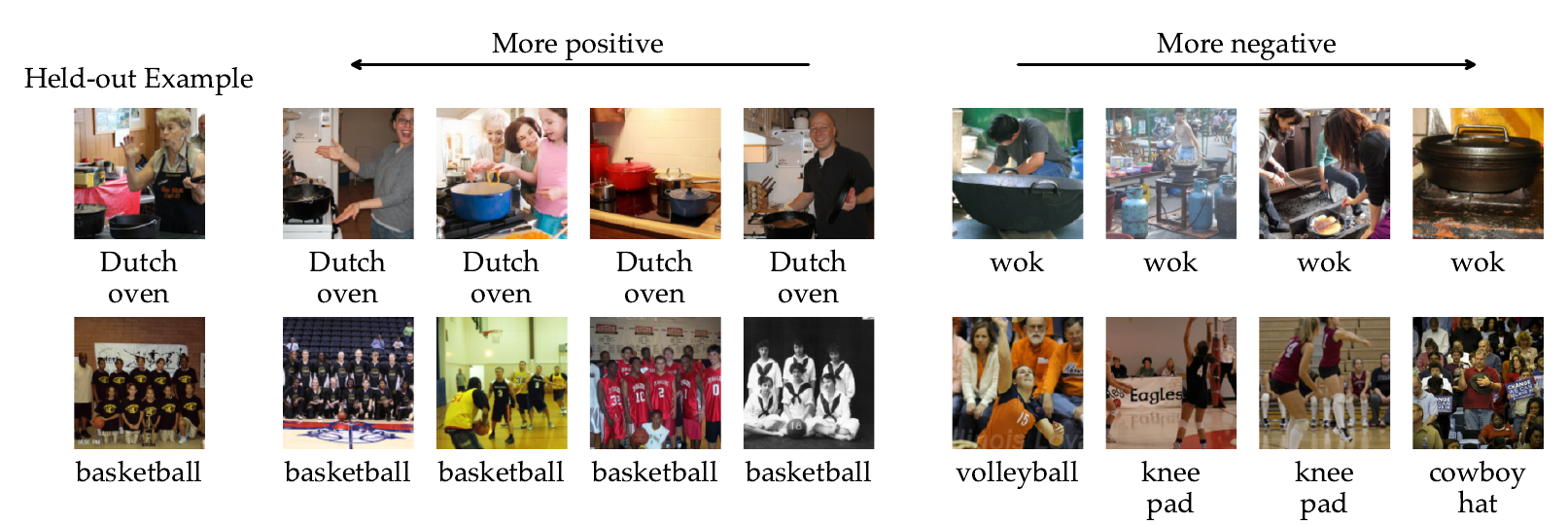}
\caption{
    We present two randomly selected test examples and their corresponding
    most helpful (highest-scoring) and most detracting (lowest-scoring)
    training examples as identified by \trak,
    for \bertbase classifiers trained on \qnli (top);
    ResNet-9 classifiers trained on \cifarten (middle);
    and ResNet-18 classifiers trained on ImageNet (bottom).
    We observe that \trak-identified training examples are
    semantically similar to the corresponding target examples,
    and that the vast majority of helpful (detracting) examples are of
    the same (different) class as the target.
    We present more such examples in \Cref{app:more_examples}
    and at \url{trak.csail.mit.edu}.
}
\label{fig:imagenet_nns}
\end{figure}

\paragraph{Inspecting \trak-identified examples.}
In \Cref{fig:imagenet_nns} we also display, for two randomly chosen test examples
from \qnli, \cifarten, and ImageNet datasets, the training examples corresponding
to the most positive and negative \trak scores.

\paragraph{Comparing \trak and datamodel scores.}
Recall from \cref{subsec:oracle} that one can view datamodels \citep{ilyas2022datamodels}
as an ``oracle'' of sorts for the linear datamodeling score (LDS) objective.
It turns out, as we show in \cref{tab:corr_between}, that \trak scores
correlate with datamodel scores, while scores of other attribution methods do not.
(We define correlation here as the Spearman rank correlation between the
vectors $\tau_\trak(z)$ and $\tau_\text{DM}(z)$, averaged over multiple examples
of interest $z$.)

\begin{table}[h]
    \centering
    \begin{tabular}{ccccccc}
        \toprule
        Method &   $\trak_{100}$ & $\trak_{20}$ & \tracin \citep{pruthi2020estimating} &  IF \citep{koh2017understanding}  &  GAS \citep{hammoudeh2022identifying} &  random \\
        \midrule
        $\rho(\tau, \tau_{\text{DM}})$  &  0.26 & 0.19 & 0.00 &  0.03 &  0.03 &  -0.03 \\
        \bottomrule
    \end{tabular}
    \caption{{\em Correlation with datamodel scores.} We measure the
    correlation between the attribution scores computed by different methods
    $\tau$ and those given by datamodels $\tau_{\text{DM}}$
    \citep{ilyas2022datamodels} on the \cifarten dataset. Specifically, for each
    test example of interest $z$, we compute the Spearman rank correlation
    $(\rho)$ between $\tau(z)_i$ and $\tau_{\text{DM}}(z)_i$ over training
    examples $i$ that have  nonzero datamodel weight $\tau_{\text{DM}}(z)_i$ and
    then average the resulting correlation over $1000$ randomly chosen examples of interest.
    $\trak_{N}$ indicates a version of $\trak$ that uses $N$ trained models in
    its estimator.}
    \label{tab:corr_between}
\end{table}

\paragraph{Understanding the roots of \trak's performance.}
In \cref{app:more_ablation}, we study the roots of \trak's performance through
an extensive ablation study. We vary, for example, how we linearize the model of
interest (Step 1 in \cref{sec:estimator_algo}), the dimension $k$ of the random
projection we use (Step 2 in \cref{sec:estimator_algo}), how we apply the Newton
step attribution from \cref{lem:formal} (Step 3 in \cref{sec:estimator_algo}),
and how we aggregate information from independently trained models (Step 4 in
\cref{sec:estimator_algo}).

As a byproduct of this investigation, we find two ways of computing \trak at
even lower cost: (a) leveraging models that have not been trained to
convergence, and (b) taking advantage of multiple checkpoints from the same
model, rather than multiple models from independent training runs.  We find (see
\cref{tab:ablation_epoch_used,tab:num_independent_runs},
explained further and reproduced in \cref{app:more_ablation})
that both of these
optimizations can dramatically reduce \trak's computational cost without
significantly degrading its performance.

\begin{table}[htbp]
    \centering
    \begin{minipage}{.46\textwidth}
      \centering
      \begin{tabular}{cc}
      \toprule
      \# training epochs & LDS \\
      \midrule
      1 & 0.100 \\
      5 & 0.204 \\
      10 & 0.265 \\
      15 & 0.293 \\
      25 & 0.308 \\
      \bottomrule
      \end{tabular}
      \caption{The performance of \trak on \cifarten as a function of the epoch at which we
      terminate model training. In all cases, \trak scores are computed with projection dimension $k = 1000$ and $M=100$ independently trained models.}
      \label{tab:ablation_epoch_used}
    \end{minipage}\hfill
    \begin{minipage}{.50\textwidth}
      \centering
      \begin{tabular}{cc}
      \toprule
      \# independent models & LDS \\
      \midrule
      5 & 0.329 \\
      6 & 0.340 \\
      10 & 0.350 \\
      100 & 0.355 \\
      \bottomrule
      \end{tabular}
      \caption{{\trak maintains its efficacy when we use multiple checkpoints from different epochs of
      the same training run instead of checkpoints from
      independently-trained models (\cifarten).} In all cases, $M=100$ checkpoints and projection dimension $k = 4000$ are used to compute \trak scores. }
      \label{tab:num_independent_runs}
    \end{minipage}
\end{table}

\section{Applications of \trak}
In \Cref{sec:eval}, we evaluated our data attribution method \trak on standard
image classification and NLP tasks and compared its performance to existing
attribution methods. We now illustrate the usefulness of \trak through three
additional applications:

\paragraph{Attributing \clip models.} In \cref{sec:CLIP}, we
use \trak to study image-text embeddings of models
trained with the \clip contrastive loss \citep{radford2021learning}. In
particular, we show how leveraging \trak allows us to identify small subsets of
the training set that, when removed, cause the resulting \clip embeddings to fail
to capture a given image-caption pair association.

\paragraph{Fact tracing language models.}
Next, in \cref{subsec:fact_trace}, we use \trak to provide data attribution for
language models \cite{vaswani2017attention}. In particular, we apply \trak to
{\em fact tracing}: the problem of tracing a language model's factual assertion
back to the corresponding training examples.
On the \ftracetrex  fact tracing benchmark, \trak significantly outperforms
the best gradient-based baseline (\tracin) used in prior work. Furthermore,
while \trak performs worse than an information retrieval baseline (BM25
\citep{robertson1995okapi}), we demonstrate that this is likely a shortcoming of the
benchmark rather than of \trak.
In particular, removing training examples traced by \trak (and re-training the
model) reduces that model's accuracy on the corresponding facts {\em more} than
removing training examples traced by BM25---and, in fact, more than
removing the {\em ground-truth} training examples as indicated by \ftracetrex.

\paragraph{Accelerating datamodel applications.} Finally, in
\cref{subsec:datamodel_apps}, we use \trak to accelerate two downstream
applications that leverage datamodel scores. That is, first, we look at the
problem of estimating {\em prediction brittleness} using datamodel scores
\cite{ilyas2022datamodels}. Then, we revisit the \modeldiff algorithm
\cite{shah2022modeldiff}, which leverages datamodel scores for {\em learning algorithm
comparison}, i.e., the task of distinguishing two learning algorithms based on
feature priors they instill. For both applications, using \trak scores in
place of datamodel scores reduces the total computational cost by at least a
factor of 100 while retaining the same effectiveness.

\subsection{Attributing CLIP models}
\label{sec:CLIP}
Recent works have found that one can leverage natural language supervision to
help models learn a rich joint image-text embedding space. In particular, CLIP (Contrastive Language-Image Pre-training)
\cite{radford2021learning} representations have become a versatile primitive
bridging visual and language domains and is used, for example, for zero-shot
classification \cite{radford2021learning} and as text encoders for latent
diffusion models \cite{rombach2022high}. While the
quality of these representations---as measured by aggregate metrics such as
downstream zero-shot accuracy---appears to be driven largely by the properties and scale of
the training datasets \cite{fang2022data,santurkar2022caption,cherti2022reproducible},
we lack a fine-grained understanding of how the composition of the training data contributes to learning well-aligned representations.
To that end, we use \trak to investigate how training data influences the resulting
CLIP embeddings at a {\em local} level. That is, we want to be able to pin-point
training examples that cause a model to learn a given {\em specific}
image-caption pair association.

\subsubsection{Computing \trak for CLIP}
Similarly to the classification setting we were considering so far, we need to
first choose an appropriate model output function (see, e.g.,
\cref{eq:modelout_mc})
to compute attribution scores with \trak.
This choice will be motivated by the CLIP training loss (which we review below)
and will reduce our setting back to the classification case.

\paragraph{The CLIP loss.}
A CLIP model with parameters $\theta$ takes in an image-caption pair $(x,y)$ and
outputs an image embedding $\imagemb{x}$ and a text embedding $\textemb{y}$.
Given a (random) batch of training examples $B = \{(x_1,y_1),...,(x_n,y_n)\}$,
the CLIP training loss computes all $n \times n$  pairwise cosine similarities
between the image and text embeddings
\[
    S_{ij} \coloneqq \imagemb{x_i} \cdot \textemb{y_j},
\]
and aims to maximize the cosine similarities $S_{ii}$ of correct pairs while
minimizing the cosine similarities $S_{ij}$, for $i\ne j$, of incorrect pairs.
More specifically, the training loss of example $(x_i,y_i) \in B$ is defined as
the following symmetric cross entropy over the similarity scores $S_{ij}$:
    \begin{equation}
        \loss{x_i,y_i} =  - \log \frac{\exp(S_{ii})}{\sum\limits_{1\le j \le n} \exp(S_{ij})} - \log \frac{\exp(S_{ii})}{\sum\limits_{1\le j \le n} \exp(S_{ji})}, \label{eq:clip_loss}
    \end{equation}
where the first term corresponds to matching each image $x_i$ to its correct
caption $y_i$, and the second term corresponds to matching each caption to its
correct image. In effect, we are solving two classification problems: one where
the images are inputs and captions (from the same batch) are labels, and vice
versa.

\paragraph{Reducing to classification.}
Recall that in the classification setting we trained the model with the cross
entropy loss (i.e., $-\log p(z;\theta)$, where $p(z;\theta)$ is the
correct-class probability), and used the model output function
$\modeleval{z}{\theta} = \log p(z;\theta)/(1-p(z;\theta))$
(\Cref{eq:modelout_mc}), i.e., the logit transform of the correct-class
probability to compute \trak scores.

To take advantage of the same formula in the CLIP setting, note that our loss
(\ref{eq:clip_loss}) can be viewed as having the form
\begin{align*}
    \loss{x_i,y_i} = -\log{p_1(x_i,y_i; \theta)} - \log{p_2(x_i,y_i; \theta)},
\end{align*}
where $p_1(x_i,y_i; \theta)$ corresponds to the probability of matching an image
to its corresponding caption based on the cosine similarity, and likewise for
$p_2(x_i,y_i;\theta)$. A natural choice of model output function in this case,
then, is using the sum of the model output functions corresponding to the
two classification problems:
    \begin{align*}
        \modeleval{x_i,y_i}{\theta} &\coloneqq \log\left(\frac{p_1(x_i,y_i; \theta)}{1 - p_1(x_i,y_i; \theta)}\right) + \log\left(\frac{p_2(x_i,y_i; \theta)}{1 - p_2(x_i,y_i; \theta)}\right) \\
         &= -\log\sum\limits_{1 \le j \le n}\exp(S_{ij} - S_{ii}) - \log\sum\limits_{1 \le j \le n}\exp(S_{ji}-S_{ii}).
    \end{align*}
Indeed, this choice allows us once again (see \Cref{ssec:multiclass}) to reduce
our problem to an instance of logistic regression and apply the same formula for
influence approximation (\Cref{lem:formal}) as before.
We can then also compute \trak scores following the same approach
(i.e., using \Cref{alg:estimator_pseudo} in \cref{sec:pseudocode}).

\begin{figure*}[t]
    \includegraphics*[width=\textwidth]{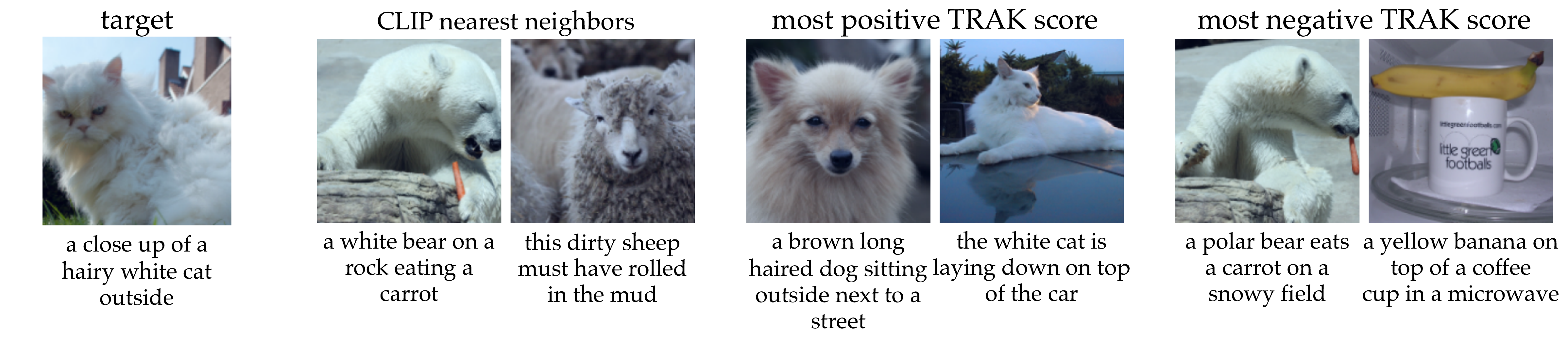}
    \includegraphics*[width=\textwidth]{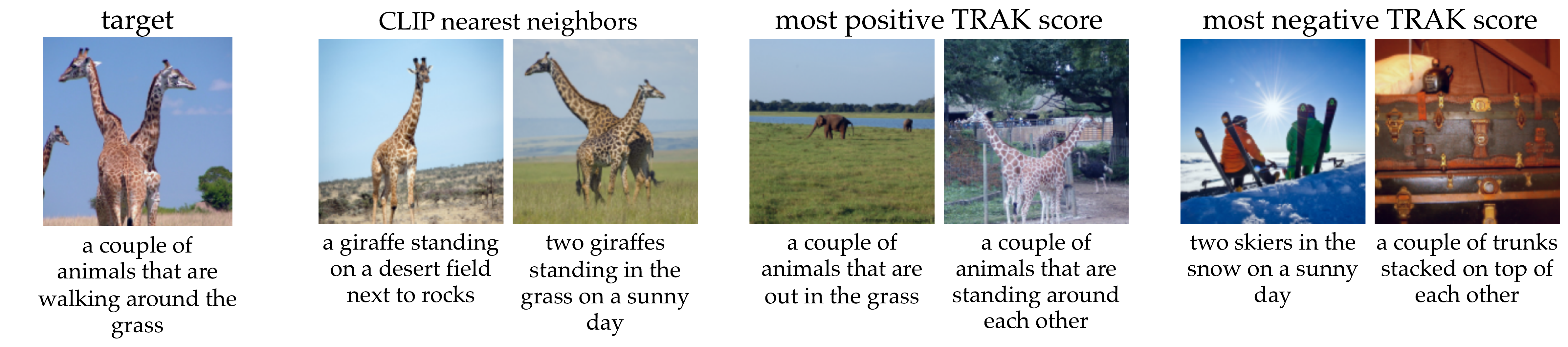}
    \caption{
        {\em Attributing CLIP trained on \mscoco.} The first column shows two
        target image-caption pairs from the validation set of \mscoco. The
        second two columns display the nearest neighbors to the target in CLIP
        embedding space (using the average of image and text cosine
        similarities). The next two columns show the train set samples that,
        according to \trak, are most helpful for aligning the image embedding to
        the caption embedding. Similarly, the last two columns display the train
        samples that are the most detracting from aligning the image and caption
        embeddings. In \Cref{app:more_examples}, we display more examples and
        also compare to \tracin.
    }
    \label{fig:CLIP_nearest_neighbors}
\end{figure*}

\subsubsection{Results}
We train image-text models (with a ResNet-50 \cite{he2015deep} as the image
encoder and a Transformer \cite{vaswani2017attention} as the text encoder) using
the \clip objective on \mscoco \citep{lin2014microsoft}. To evaluate the
effectiveness of \trak applied to such \clip models, we perform a qualitative
(visual) analysis; and a quantitative (counterfactual) evaluation. In both cases, we
compare \trak with \tracin and \clip similarity distance\footnote{We use the average of cosine similarities between the image
embeddings and between the text embeddings.} baselines.

\paragraph{Visual analysis.}
\cref{fig:CLIP_nearest_neighbors} displays two target examples of interest along
with the corresponding training examples having the highest attribution scores
(according to \trak and \clip similarity distance---see \Cref{app:more_examples} for the analysis corresponding to \tracin). For the first example, the
nearest neighbor in the \clip space (the polar bear) turns out to have a {\em
negative} attribution score according to \trak. For the second example, the most
helpful \trak examples are the ones for which the captions contain the
phrase ``a couple of animals'' but where the images do not necessarily feature
giraffes (possibly because the target caption does not mention ``giraffe'' either). On
the other hand, the most helpful examples according to \clip
similarity distance all feature giraffes. These differences suggest that \trak
attribution scores may capture significantly different traits from \clip
similarity distance.

\paragraph{Counterfactual evaluation.}
We next investigate to what extent training examples identified by each
attribution method affect the \clip model's ability to learn a given
image-caption association.
Specifically, we say that a \clip model has {\em learned} a given association between
an image and a caption whenever their corresponding image and caption embeddings
have high cosine similarity relative to other image-caption pairs. To
evaluate each attribution method (i.e., \trak, \tracin, and \clip similarity
distance), for a given target image-caption pair, we remove from the training
set the $k$ examples with the most positive attribution scores a given
attribution method produces, and then re-train a model from scratch (averaging
over ten training runs to reduce stochasticity). Finally, we examine the decrease in
cosine similarity between the embeddings of target image and caption pair, and average this
result over different target pairs.

Our results (\Cref{fig:CLIP_barplot_counterfactuals}) indicate that removing
training inputs identified by \trak can significantly degrade the model's
ability to learn the target image-caption pair. Indeed, removing just $k=400$
target-specific training puts (i.e., less than $0.5\%$ of the train set)
decreases the (average) \clip similarity distance between the target image and
caption embeddings by 0.36. In contrast, removing the same number of nearest
neighbors in \clip space results in a much smaller effect size (a 0.11 decrease),
while removing training examples identified by \tracin has no
significant effect.

\begin{figure}[h]
    \centering
    \pgfplotstableread[col sep=comma]{data/CLIP_barplot_data_bootstrap.csv}\datatable

\begin{tikzpicture}
    \definecolor{darkgray141160203}{RGB}{141,160,203}
    \definecolor{dimgray85}{RGB}{85,85,85}
    \definecolor{gainsboro229}{RGB}{229,229,229}
    \definecolor{lightgray204}{RGB}{204,204,204}
    \definecolor{mediumaquamarine102194165}{RGB}{102,194,165}
    \definecolor{orchid231138195}{RGB}{231,138,195}
    \definecolor{salmon25214198}{RGB}{252,141,98}
    \definecolor{amethyst}{rgb}{0.6, 0.4, 0.8}
    \definecolor{bleudefrance}{rgb}{0.19, 0.55, 0.91}
    \definecolor{blush}{rgb}{0.87, 0.36, 0.51}
    \definecolor{brilliantrose}{rgb}{1.0, 0.33, 0.64}
    \definecolor{trakorange}{rgb}{0.87, 0.6, 0.48}
    \definecolor{trakgreen}{rgb}{0.51, 0.705, 0.635}
    \definecolor{trakblue}{rgb}{0.60, 0.64, 0.753}

    \begin{axis}[
        axis background/.style={fill=gainsboro229},
        axis line style={white},
        width=0.6\textwidth,
        height=5.5cm,
        ybar,
        tick align=outside,
        tick pos=left,
        enlarge x limits=0.5,
        legend cell align={left},
        legend style={
            fill opacity=0.8,
            draw opacity=1,
            text opacity=1,
            row sep=0.25cm,
            at={(1.05,.8)},
            anchor=north west,
            draw=gainsboro229,
            fill=gainsboro229,
            /tikz/every even column/.append style={column sep=0.5cm},
            /tikz/every odd column/.append style={column sep=0.1cm}
        },
        ylabel={Drop in cosine similarity},
        xlabel={Number of training examples removed},
        xtick={50, 400},
        xticklabels={$k=50$, $k=400$},
        nodes near coords={
            \pgfmathprintnumber[fixed]{\pgfplotspointmeta}
        },
        nodes near coords style={
            yshift=0.25cm,
            text=gray,
            font=\small
        },
        nodes near coords align={vertical},
        ymin=0,
        ymax=1, %
        error bars/y dir=both,
        error bars/y explicit,
        error bars/error bar style={gray},
        bar width={1cm},
        y grid style={white},
        ymajorgrids,
        ymin=0, ymax=0.5,
        ytick style={color=dimgray85}
        ]
        \addplot+[fill=trakorange,
                  draw=trakorange, 
                  error bars/.cd, 
                  y dir=both, 
                  y explicit]
            table[
                x expr={\thisrow{k}}, 
                y=value, 
                y error plus=upper_error, 
                y error minus=lower_error, 
                col sep=comma, 
                restrict expr to domain={\thisrow{method}}{1:1}]{\datatable}; 
        
        \addplot+[fill=trakblue, 
                  draw=trakblue,
                  error bars/.cd, 
                  y dir=both, 
                  y explicit]
            table[
                x expr={\thisrow{k}}, 
                y=value, 
                y error plus=upper_error, 
                y error minus=lower_error, 
                col sep=comma, 
                restrict expr to domain={\thisrow{method}}{0:0}]{\datatable}; 
        
        \addplot+[fill=trakgreen, 
                  draw=trakgreen,
                  error bars/.cd, 
                  y dir=both, 
                  y explicit]
            table[
                x expr={\thisrow{k}}, 
                y=value, 
                y error plus=upper_error, 
                y error minus=lower_error, 
                col sep=comma, 
                restrict expr to domain={\thisrow{method}}{2:2}]{\datatable}; 
        
        \legend{CLIP NNs, TracIn \citep{pruthi2020estimating}, TRAK (ours)}
    \end{axis}
\end{tikzpicture}
    \caption{
    {\em Which training inputs can we remove from the training set so as the
    resulting \clip model no longer associates a target image with its caption?}
    We measure how the cosine similarity between target image and caption
    embeddings is affected when we re-train a \clip model after removing the most
    influential training examples---as identified by \trak, \tracin, and \clip
    similarity distance. We report the {\em decrease} in cosine similarity, averaged over
    $100$ randomly selected image-caption pairs from the validation set. Error bars
    represent $95\%$ confidence intervals.
    }
    \label{fig:CLIP_barplot_counterfactuals}
\end{figure}
\subsection{Fact tracing for large language models (\mtfive)}
\label{subsec:fact_trace}
As large language models are deployed in a variety of contexts, e.g.,
as conversation agents \citep{thoppilan2022lamda}
or knowledge bases \cite{petroni2019language},
there is an emerging need to be able to
attribute models' outputs back to specific data sources
\cite{bohnet2022attributed}.
To that end, we study {\em fact tracing} \cite{akyurek2022towards}, i.e.,
the task of identifying the training examples that cause a language model to
generate a given ``fact.''

\paragraph{A benchmark for fact tracing.}
\citet{akyurek2022towards} develop a testbed for the fact tracing problem
by way of a dataset (and corresponding evaluation methodology) called \ftracetrex.
We provide a high-level overview of \ftracetrex here,
and describe it in more depth in \cref{app:fact_tracing:dataset}.
The \ftracetrex dataset consists of a set of ``abstracts'' and a set
of ``queries,'' both of which pertain to the same database of
``facts.''
\citet{akyurek2022towards} annotate each abstract
with a set of facts it expresses, and
each query with the (single) fact that it asks about.
As a part of the task setup, one finetunes a pre-trained language model on the set
of abstracts using {\em masked language modeling},\footnote{
In masked language modeling \citep{raffel2020exploring}, the language model is asked to predict the tokens
corresponding to a masked-out portion of the input. In \ftracetrex, either a subject or object in the abstract is masked out.}
and then evaluates this model's correctness on each query in
the query set.
This step defines a set of ``novel facts,'' i.e.,
queries that the model answers correctly {\em only after} finetuning.

With the above setup in place, we can define the \ftracetrex fact tracing benchmark.
\citet{akyurek2022towards} reason that each novel fact (as identified above)
should have been learned (during finetuning) from the abstracts that express the same fact.
The benchmark thus evaluates a given data
attribution method's ability to retrieve, for each novel fact,
the abstracts in the training set that express the same fact.
(Such abstracts are called the {\em ground-truth proponents} of the query.)

In particular, observe that applying a data attribution method $\tau(\cdot)$ to
a particular query (treating the set of abstracts as the training set) yields
scores that we can use as a ranking over the set of the abstracts.
\citet{akyurek2022towards} compute the {\em mean reciprocal rank} (MRR)
of the ground-truth proponents in this ranking (see \cref{app:fact_tracing:dataset}),
a standard metric from information retrieval,
to quantify the efficacy of $\tau(\cdot)$ at fact tracing.
We evaluate \trak on this benchmark,
along with two baselines from
\citep{akyurek2022towards}, \tracin \cite{pruthi2020estimating}
and the information retrieval method BM25 \cite{robertson1995okapi}.

\paragraph{Computing \trak scores for language models.}
To apply \trak to this setting, we need to choose an appropriate model output function,
as we did before for the classification setting (see \Cref{ssec:multiclass}) and for \clip (see \Cref{sec:CLIP}).
To this end, we observe that the  masked language modeling objective
has a natural interpretation as a sequence of $v$-way classification problems over the masked tokens,
where $v$ is the vocabulary size. %
Thus, inspired by our analysis of the multi-class classification setting
from \cref{ssec:multiclass}, %
we choose the model output function for this setting to be
the sum of the ``canonical'' model output function \eqref{eq:modelout_mc}
for each of the $v$-way classification problems
(see \cref{app:fact_tracing:trak} for more details).

\subsubsection{Results and discussion}
We find that while \trak significantly outperforms \tracin on
the \ftracetrex benchmark (0.42 vs. 0.09 using the aforementioned MRR score),
neither method matches the
performance of the information retrieval baseline BM25 (0.77 MRR).\footnote{Note that
while our finding that BM25 outperforms \tracin
matches that of \citet{akyurek2022towards}, our exact numbers are incomparable
due to the mismatch in model classes.}

To understand the possible roots of \trak's underperformance
relative to BM25 on \ftracetrex,
we carry out a counterfactual analysis.\footnote{
    See \cref{app:fact_tracing:cfx} for a detailed account of our experiment.
}
Specifically, for a subset $S^\star$ of the \ftracetrex query set, we
create three corresponding {\em counterfactual training sets}.
Each such training set corresponds to {\em removing} one of three collections of abstracts
from the \ftracetrex abstract set:
\begin{enumerate}
    \item[(a)] the most important abstracts for model performance on $S^\star$, as estimated by \trak;
    \item[(b)] the abstracts that are most similar to the queries in $S^\star$ according to BM25;
    \item[(c)] the corresponding ``ground-truth proponents'' for the queries in $S^\star$ as per \ftracetrex.
\end{enumerate}
We then measure the average {\em decrease} in performance on $S^\star$ when a model is finetuned on these
counterfactual datasets compared finetuning on the full training set.
Intuition would suggest that performance would decrease the most when models are
trained on the counterfactual training set~(c); in particular,
there is ostensibly {\em no} direct evidence for {\em any} of the facts corresponding to the queries in $S^\star$
anywhere in that set.

We find (see \cref{fig:nlp_counterfactual}), however, that it is
only the \trak-{\em based counterfactual training set} that
causes a large change in model behavior.
That is, removing abstracts identified with \trak
leads to a 34\% decrease in accuracy,
significantly more than the decreases induced by removing abstracts according to
BM25 (10\%) or even removing {\em ground-truth proponents} (12\%).

\begin{figure}[h]
    \centering
    \begin{tikzpicture}
    \definecolor{darkgray141160203}{RGB}{141,160,203}
    \definecolor{dimgray85}{RGB}{85,85,85}
    \definecolor{gainsboro229}{RGB}{229,229,229}
    \definecolor{lightgray204}{RGB}{204,204,204}
    \definecolor{mediumaquamarine102194165}{RGB}{102,194,165}
    \definecolor{orchid231138195}{RGB}{231,138,195}
    \definecolor{salmon25214198}{RGB}{252,141,98}
    \definecolor{amethyst}{rgb}{0.6, 0.4, 0.8}
    \definecolor{bleudefrance}{rgb}{0.19, 0.55, 0.91}
    \definecolor{blush}{rgb}{0.87, 0.36, 0.51}
    \definecolor{brilliantrose}{rgb}{1.0, 0.33, 0.64}
    \definecolor{trakorange}{rgb}{0.87, 0.6, 0.48}
    \definecolor{trakgreen}{rgb}{0.51, 0.705, 0.635}
    \definecolor{trakblue}{rgb}{0.60, 0.64, 0.753}
    \begin{axis}[
        axis background/.style={fill=gainsboro229},
        axis line style={white},
        width=0.6\textwidth,
        height=5.5cm,
        ybar,
        tick align=outside,
        tick pos=left,
        enlarge x limits=0.5,
        ylabel={Accuracy drop on targets},
        xlabel={Method},
        xtick={1,2,3},
        xticklabels={Ground-truth, BM25, TRAK},
        nodes near coords={
            \pgfmathprintnumber[fixed]{\pgfplotspointmeta}
        },
        nodes near coords style={
            yshift=0.3cm,
            text=gray,
            font=\small
        },
        nodes near coords align={vertical},
        ymin=0,
        ymax=1, %
        error bars/y dir=both,
        error bars/y explicit,
        error bars/error bar style={gray},
        bar width={1cm},
        y grid style={white},
        ymajorgrids,
        ymin=0, ymax=50,
        ytick style={color=dimgray85}
        ]
        \addplot[fill=trakorange,
                  draw=trakorange, 
                  bar shift=0cm,
                  error bars/.cd, 
                  y dir=both, 
                  y explicit]
            coordinates{
                (1, 16.5) +- (0, 1)%
            }; 
        
        \addplot[fill=trakblue, 
                  draw=trakblue,
                  error bars/.cd, 
                  y dir=both, 
                  y explicit]
            coordinates {
                (2, 12) +- (3.0, 3.0)
            }; 

        \addplot[fill=trakgreen, 
                  draw=trakgreen,
                  bar shift=0cm,
                  error bars/.cd, 
                  y dir=both, 
                  y explicit]
            coordinates {
                (3, 34) +- (4.8, 4.6)
            }; 
    \end{axis}
\end{tikzpicture}
\caption{{\em Identifying counterfactually important examples for learning facts on \ftracetrex.}
        Given a set of queries that the language model (\texttt{mt5-small}) originally answers correctly after training, we compare how three different interventions---removing abstracts with the highest \trak scores, removing the most similar abstracts according to BM25, and removing the ground-truth proponents as indicated by \ftracetrex---affect the resulting model's accuracy on the queries.
        The $y$-axis shows the {\em decrease} in accuracy (on the query set, relative to the original model) after each intervention; results are averaged over 50 queries and eight independent models.
        Error bars represent $95\%$ confidence intervals.
    }
    \label{fig:nlp_counterfactual}
\end{figure}

\paragraph{Discussion.}
Our results demonstrate that while \trak may not be effective at identifying abstracts
that directly express the same fact as a given query
(i.e., the ground-truth proponents as defined by \ftracetrex),
it {\em can} successfully identify the abstracts that are most
responsible for the finetuned model {\em learning} a given fact.
In particular, \trak's subpar performance on the attribution benchmark
is an artifact of the \ftracetrex benchmark rather than a flaw of \trak itself.

There are several potential explanations for this phenomenon, many of which
\citet{akyurek2022towards} already discuss in their work:
\begin{itemize}
    \item There may be errors in the \ftracetrex benchmark. (Although,
    given the drastic difference between the \trak scores
    and the ground-truth labels in their ability to identify counterfactually important abstracts,
    such data errors are unlikely to be the sole culprit.)
    \item Models may be answering queries by
    {\em combining} facts from the training set.
    For example, neither ``The largest pyramid is in Giza'' nor ``Giza is a city in Egypt'' would be
    ground-truth proponents for the query ``Which country is home to the largest pyramid?'' in \ftracetrex,
    but a model that learns both of these facts
    may still be able to correctly answer that query.
    \item Alternatively, models may be learning from the syntactic rather than
    semantic structure of abstracts. For example, a model may correctly answer
    that a person from Korea is called a ``Korean'' by learning from an
    abstract which says ``A person from Bulgaria is Bulgarian.''
\end{itemize}

More broadly,
our results highlight a difference between {\em fact tracing}
and {\em behavior tracing}.
In other words,
finding a data source that supports a given model-generated text is a different task
than identifying the actual data sources that {\em caused} the model to generate this text in
the first place.
While we may be able to address the former problem with { model-independent}
techniques such as information retrieval or web search,
the latter requires methods that remain faithful to (and thus, dependent on)
the model being studied. Our results here indicate that \trak can be an effective tool for the latter problem.

\subsection{Accelerating datamodel applications}
\label{subsec:datamodel_apps}
Our evaluation thus far has demonstrated that data attribution scores computed with
\trak  can {\em predict} how a given model's output changes as a function of the composition of the corresponding model's training set.
While the capability to make such predictions is useful in its own right, prior
work has shown that this primitive also enables many downstream applications
\cite{koh2017understanding,jia2019towards,alaa2020discriminative}.  For example,
prior works leverage datamodel scores to identify brittle predictions
\cite{ilyas2022datamodels} and to compare different learning algorithms
\cite{shah2022modeldiff}. We now show that using \trak in place of datamodel
scores can significantly speed up these downstream applications too.

\subsubsection{Estimating prediction brittleness}
\citet{ilyas2022datamodels} use datamodel scores to
provide {\em lower bounds} on the {\em brittleness} of a given example---that
is, given an example of interest $z$, they identify a subset of the training set
whose removal from the training data causes the resulting re-trained model to
misclassify $z$.
The brittleness estimation algorithm that \citet{ilyas2022datamodels} leverage
hinges on the fact that the datamodel attribution function $\tau_\text{DM}(z)$
can accurately predict model outputs, i.e., achieve high LDS. Motivated by
\trak's good performance on the linear datamodeling task (see, e.g.,
\cref{fig:headline_full}), we examine estimating the brittleness of \cifarten examples
using \trak scores in place of datamodel ones (but otherwise following the
procedure of \citet{ilyas2022datamodels}). Our results (see
\Cref{fig:brittle}) indicate that \trak scores computed from an ensemble of just
100 models are about as effective at estimating brittleness as datamodel scores
computed from 50,000 models. Thus, \trak scores can be a viable (and orders of
magnitude faster) alternative to datamodels for estimating
prediction brittleness.

\begin{figure*}[hb]
    \centering
    \begin{tikzpicture}
    \centering
    \definecolor{darkgray141160203}{RGB}{141,160,203}
    \definecolor{dimgray85}{RGB}{85,85,85}
    \definecolor{gainsboro229}{RGB}{229,229,229}
    \definecolor{lightgray204}{RGB}{204,204,204}
    \definecolor{mediumaquamarine102194165}{RGB}{102,194,165}
    \definecolor{orchid231138195}{RGB}{231,138,195}
    \definecolor{salmon25214198}{RGB}{252,141,98}
    \definecolor{amethyst}{rgb}{0.6, 0.4, 0.8}
    \definecolor{bleudefrance}{rgb}{0.19, 0.55, 0.91}
    \definecolor{blush}{rgb}{0.87, 0.36, 0.51}
    \definecolor{brilliantrose}{rgb}{1.0, 0.33, 0.64}
    \begin{groupplot}[
        group style={group size= 1 by 1, horizontal sep=0cm, vertical sep=0cm},
        height={6cm},
        width=.55\linewidth]
    \nextgroupplot[
        xlabel={\# training examples removed to flip prediction},
        ylabel={Frac. of CIFAR-10 test set},
        legend pos={north east},
        axis background/.style={fill=gainsboro229},
        axis line style={white},
        legend cell align={left},
        legend columns=1,
        legend style={
        fill opacity=0.8,
        draw opacity=1,
        text opacity=1,
        at={(1.05,0.9)},
        anchor=north west,
        draw=gainsboro229,
        fill=gainsboro229,
        },
        ymajorgrids=true,
        no marks,
        xmin=0, xmax=1000,
        xmajorgrids, xminorgrids,
        x grid style={white},
        y grid style={white},
        tick align=outside,
        tick pos=left,
    ]

    \addplot[ultra thick, color=mediumaquamarine102194165] table [x=x, y=y, col sep=comma] {data/brittleness_methods/Datamodel300k.csv};
    \addlegendentry{Datamodels ($300$k models)}

    \addplot[ultra thick, color=black, dashed] table [x=x, y=y, col sep=comma] {data/brittleness_methods/TRAK.csv};
    \addlegendentry{TRAK (100 models)}

    \addplot[ultra thick, color=mediumaquamarine102194165] table [x=x, y=y, col sep=comma] {data/brittleness_methods/Datamodel50k.csv};
    \addlegendentry{Datamodels ($50$k models)}

    \addplot[ultra thick, color=orchid231138195] table [x=x, y=y, col sep=comma] {data/brittleness_methods/TracIn.csv};
    \addlegendentry{TracIn ($100$ models)}

    \addplot[ultra thick, color=blush] table [x=x, y=y, col sep=comma] {data/brittleness_methods/RepresentationSimilarity.csv};
    \addlegendentry{Representation Sim. ($100$ models)}

    \addplot[ultra thick, color=bleudefrance] table [x=x, y=y, col sep=comma] {data/brittleness_methods/InfluenceFunction.csv};
    \addlegendentry{Influence Function ($100$ models)}
    
    \end{groupplot}
\end{tikzpicture}
    \caption{{\em Using \trak scores to identify brittle model predictions.}
    Following the methodology of \citet{ilyas2022datamodels}, we apply
    different data attribution methods to estimate the brittleness of model predictions on
    examples from the \cifarten validation set. The number of models used by each attribution method is specified in parentheses, e.g., \trak
    (100) indicates that \trak scores were computed using an ensemble of 100 trained
    models.}
\label{fig:brittle}
\end{figure*}

\subsubsection{Learning algorithm comparisons}
A useful way to leverage datamodels is to view them as {\em data representations}.
More specifically, following \citet{ilyas2022datamodels}, for an example of interest $z$, one can view the datamodel
attribution $\tau_\text{DM}(z)$ as an embedding of $z$ into $\mathbb{R}^{n}$,
where $n$ is the size of the training dataset.  Analyzing examples in such
induced {\em datamodel representation spaces} turns out to enable uncovering
dataset biases and model-specific subpopulations
\cite{ilyas2022datamodels}.
Furthermore, this representation space is not specific to a particular model
instance or architecture---it is {\em globally aligned} in the sense that for
the same example $z$, the attribution score $\tau_\text{DM}(z)_i$ of a given
train example $i$ has a consistent interpretation across {\em different}
learning pipelines.
\citet{shah2022modeldiff} leverage the properties of the datamodel
representation space to perform model-agnostic {\em learning algorithm comparison}
(called \modeldiff): given two learning algorithms, they show how to
use datamodels to identify {\em distinguishing features}, i.e., features that
are used by one learning algorithm but not the other.

Once again, motivated by \trak's good performance on the LDS metric,
we investigate whether \trak scores can substitute for datamodel scores in this
context. To this end, we revisit one of the case studies from
\citet{shah2022modeldiff}---the one that compares image classifiers trained {with}
and {without} data augmentation, and identifies features that distinguish these
two classes of models.
When applied to this case study, \modeldiff computed with \trak scores recovers similar distinguishing features
to the ones originally found by \citet{shah2022modeldiff} (using datamodel scores)---see \Cref{fig:modeldiff} for more details.
Also, employing \trak scores in place of datamodel scores reduces the total
computational cost by a factor of 100, showing, once again, that \trak can dramatically
accelerate downstream tasks that rely on accurate attribution scores.

\section{Related work}
\label{sec:related}
In this section, we highlight and discuss how \trak connects to prior works on training data attribution, the neural tangent kernel, and kernel approximation.

\paragraph{Training data attribution.}
There is a sizable body of work on data attribution methods.
Here we discuss approaches most similar to ours, but we refer the reader back to \Cref{sec:prelim} for an overview of prior work on data attribution methods and to \citep{hammoudeh2022training} for an even more
extensive survey.

In the setting of generalized linear models, \citet{wojnowicz2016influence} speed up classical influence estimation
(\Cref{lem:formal})
by leveraging random projections. Also, \citet{khanna2019interpreting} employ a similar estimator based on the Fisher
matrix for data attribution and subset selection. Their experiments are limited though
to small neural networks and linear models.
 Most similarly to our approach, \citet{achille2021lqf} leverage the linearized model for approximating influence functions
(among other applications). However,
their approach introduces several changes to the model of interest (such as modifying activations, loss, and
regularization) and focuses on finetuning in smaller-scale settings, whereas \trak can be applied directly to the original model (and at scale).

Similarly to us, prior works also investigate the tradeoffs between scalability and efficacy of
data attribution methods. For instance, \citet{jia2021scalability} study these tradeoffs by proposing new metrics and comparing according to them leave-one-out methods (e.g., influence functions) and Shapley values.
They put forth, in particular, a new estimator for Shapley values that is based on approximating the original model with a $k$-nearest neighbors model over the pre-trained embeddings---this can be viewed as an alternative to working with the linearized model.

As discussed in \Cref{sec:prelim}, a major line of work uses {\em Hessian-based influence functions} for data attribution \cite{koh2017understanding,koh2019accuracy,basu2021influence}.
In particular, the influence function effectively computes---up to an error that can be bounded---the one-step Newton approximation  with respect to the full model parameters \cite{koh2019accuracy}.
Recall that \trak also leverages the one-step Newton approximation in order to estimate  leave-one-out influences for logistic regression (see \Cref{sec:method}).
However, in contrast to the influence function approach, the Hessian matrix we leverage (the matrix $X^\top R X$ in \Cref{lem:formal}) is positive semi-definite as it is computed with respect to the {\em linearized model} rather than the original model.
As a result, computing \trak does not require the use of additional regularization (beyond the one implicitly induced by our use of random projections), which is practically necessary in the influence function approach.
Prior works also leverage a similar Hessian matrix based on the generalized Gauss-Newton matrix \cite{bae2022if} or the equivalent Fisher information matrix \cite{teso2021interactive}, which are guaranteed to be positive semi-definite.

\paragraph{Neural tangent kernel.}
The neural tangent kernel (NTK) \cite{jacot2018neural} and its generalizations \cite{yang2021tensor} are
widely studied as a tool for theoretically analyzing generalization
\citep{arora2019fine}, optimization \citep{wei2019regularization}, and
robustness \citep{gao2019convergence} of (overparameterized) neural networks.  While these works focus on neural networks in the their
large or infinite-width limit, a line of recent works
\citep{mu2020gradients,achille2021lqf,long2021properties,atanasov2022neural,wei2022more,malladi2022kernel,atanasov2023onset,ma2022behind}
studies instead the
finite-width {\em empirical NTK} (eNTK).
Our \trak estimator is partly motivated by the observation from this line of work that kernel regression with the eNTK provides a good approximation to the original model.

While we leverage the eNTK approximation for data attribution, prior works leveraged the NTK and eNTK for
various other applications, such as studying generalization
\citep{bachmann2022generalization}, sample selection for active
learning \citep{holzmuller2022framework}, model selection
\citep{deshpande2021linearized}, federated learning
\citep{yu2022tct}, and fast domain adaptation \citep{maddox2021fast}.  Our
reduction to the linear case (Step 1 in \Cref{sec:estimator_algo}) is analogous to the approach of
\citet{bachmann2022generalization} that leverages formulas for the leave-one-out error of kernel methods coupled with the NTK approximation to
estimate the generalization error.
Another related work is that of \citet{zhang2022rethinking}, who theoretically characterize the accuracy of the Hessian-based influence function
in the NTK regime (i.e., large-width limit).

Finally, although the work on NTK popularized the idea of leveraging gradients as features,
similar ideas can be traced back to works on the Fisher kernel and related ideas
\cite{zinkevich2017holographic}.

\paragraph{Kernel methods and random projections.}
Our application of random projections to improve computational efficiency of kernel approximation is a widely used idea in
kernel methods
\citep{blum2006random,rahimi2007random}. Aside from computational advantages, this technique can also provide insight into empirical phenomena. For example, \citet{malladi2022kernel} use the
kernel view along with random projections as a lens to explain the efficacy of subspace-based finetuning methods.

\section{Discussion \& Conclusion}
In our work, we formalize the problem of data attribution and introduce a new
method, \trak, that is effective and efficiently scalable. We then demonstrate the usefulness of \trak in a variety of
large-scale settings: image classifiers trained on \cifar and ImageNet, language models (\bert
and \mtfive), and image-text models (\clip).

Still, \trak is not without limitations: in particular, it requires the model to be differentiable, and its effectiveness also depends
on the suitability of the linear approximation.
That said, the success of the applying the NTK on language modeling tasks \cite{malladi2022kernel} as well as our own experiments both suggest that this approximation is likely to continue to work for larger models.
\trak presents a unique opportunity to reap the benefits of data attribution in previously untenable domains, such as large generative models.
In \cref{app:future_work}, we further discuss possible avenues for future work.

\clearpage
\section*{Acknowledgements}
We thank Ekin Akyurek for help 
installing and using the \ftracetrex benchmark.

Work supported in part by the NSF grants CNS-1815221 and DMS-2134108, and Open Philanthropy. This material is based upon work supported by the Defense Advanced Research Projects Agency (DARPA) under Contract No. HR001120C0015.

Research was sponsored by the United States Air Force Research Laboratory and the United States Air Force Artificial Intelligence Accelerator and was accomplished under Cooperative Agreement Number FA8750-19-2-1000. The views and conclusions contained in this document are those of the authors and should not be interpreted as representing the official policies, either expressed or implied, of the United States Air Force or the U.S. Government. The U.S. Government is authorized to reproduce and distribute reprints for Government purposes notwithstanding any copyright notation herein.

\printbibliography

\clearpage
\appendix
\addcontentsline{toc}{section}{Appendix} %
\renewcommand\ptctitle{Appendices}
\part{}
\parttoc
\clearpage

\counterwithin{figure}{section}
\counterwithin{table}{section}
\counterwithin{algorithm}{section}

\section{Experimental Setup}
\label{app:exp_setup}
\subsection{Datasets and models}
\label{app:datasets_models}

\paragraph{\cifar.}
We construct the \cifartwo dataset as the subset of \cifarten \citep{krizhevsky2009learning} consisting
of only the ``cat'' and ``dog'' classes. We initially used \cifartwo as the main test bed when designing \trak, as it is a binary classification task and also smaller in size.
On both \cifartwo and \cifarten, we train a ResNet-9 architecture.\footnote{\url{https://github.com/wbaek/torchskeleton/blob/master/bin/dawnbench/cifar10.py}}
For \cifartwo, we use (max) learning rate $0.4$, momentum $0.9$, weight decay $5\text{e-}4$, and train for $100$ epochs using a cyclic learning rate schedule with a single peak at epoch 5.
For \cifarten, we replace the learning rate with $0.5$ and train for $24$ epochs.

Our code release includes a notebook\footnote{\url{https://github.com/MadryLab/trak/blob/main/examples/cifar2_correlation.ipynb}} that can reproduce the \cifartwo results end-to-end.
\paragraph{ImageNet.}
We use the full 1000-class ImageNet dataset and train a modified ResNet-18 architecture.
Models are trained from scratch for 15 epochs,
cyclic learning rate with peak at epoch 2 and initial learning rate $5.2$,
momentum $0.8$, weight decay $4\text{e-}5$, and label smoothing $0.05$.

\paragraph{\qnli.}
We finetune a pre-trained \bert model (\texttt{bert-base-cased}\footnote{\url{https://huggingface.co/bert-base-cased}}) on the \qnli (Question-answering Natural Language Inference) task from the \glue benchmark.
We use the default training script\footnote{\url{https://github.com/huggingface/transformers/blob/main/examples/pytorch/text-classification/run_glue.py}} from HuggingFace with a few modifications: we use SGD (20 epochs, learning rate starting at $1\text{e-}3$) instead of AdamW, and we remove the last \texttt{tanh} non-linearity before the classification layer. Removing the last non-linearity prevents the model outputs in saturating, resulting in higher LDS. (That said, we find that \trak scores can be still computed on the models with non-linearity; this was only for improving evaluation.)
We restrict the training set to 50,000 examples, approximately half of the full training set.

\paragraph{\clip on \mscoco.}
We use an open-source implementation\footnote{\url{https://github.com/mlfoundations/open_clip}} of \clip.
The model uses a ResNet-50 for the image encoder and a Transformer for the text encoder (for captions).
We train for $100$ epochs using the
Adam optimizer with batch size $600$, a cosine learning rate schedule with starting learning rate $0.001$, weight decay $0.1$, and momentum $0.9$. All images are
resized to a resolution of $224\times 224$. We use random resize crop, random
horizontal flip, and Gaussian blur as data augmentations.

In the counterfactual evaluation, we consider a normalized
notion of cosine similarity, $\bar{r} = r/(r_\text{95} - r_\text{5})$, where $r$ is the raw correlation between image and caption embeddings and $r_{\alpha}$ is the $\alpha$-percentile of image-caption similarities across the entire dataset. Results remain similar with other choices of metric.

\paragraph{Fact tracing \mtfive on \ftracetrex.}
We follow the setup exactly as in \citet{akyurek2022towards} as we describe in \Cref{subsec:fact_trace}, other than using a smaller architecture (\texttt{mt5-small}). See \Cref{app:lexical} for more details.

\paragraph{\textsc{ModelDiff} on \textsc{Living17}.}
The \textsc{Living17} dataset~\cite{santurkar2021breeds} is an image classification dataset derived from the ImageNet dataset and consists of 17 classes, each comprised of four original ImageNet classes.

We train the standard ResNet-18 architecture on the above dataset, either using standard data augmentation (random resized cropping and random horizontal flips) or with no data augmentation (only center cropping, same as used on when evaluating). The goal of the case study from \citet{shah2022modeldiff} is to distinguish the above two learning algorithms in terms of the feature priors of the resulting trained models.
To run \textsc{ModelDiff}, follow the setup in \citet{shah2022modeldiff} exactly; we refer to the work for more details of the case study and implementation details.

\subsection{\trak hyperparameters}
\label{app:trak_hparams}

\trak only has two hyperparmeters: the projection dimension $k$ and the number of models $M$.
The following hyperparameters were used unless specified otherwise:
\begin{table}[!htbp]
    \centering
    \begin{tabular}{llrr}
        \toprule
        Dataset & Model & Number of models ($M$) & Projection dimension ($k$) \\
        \midrule
        \cifartwo & ResNet-9 & - & 4,000 \\
        \cifarten & ResNet-9 & - & 20,000 \\
        \qnli & \bertbase & - & 4,000 \\
        ImageNet & ResNet-18 & - &  15,000 \\
        \mscoco & ResNet-50 (\clip) & 100 & 20,000 \\
        \ftracetrex & \texttt{mt5-small} & 10 & 4,000 \\
        \textsc{Living-17} & ResNet-18 & 100 & 1,000 \\
        \bottomrule
    \end{tabular}
    \caption{\trak hyperparameters used for different experiments. Blank indicates that different numbers were used depending on the experiment.}
    \label{tab:proj_dim}
\end{table}

\paragraph{Soft-thresholding.} An optional hyperparameter is needed if we use soft-thresholding (Step 5).
Among the four tasks we evaluate the LDS on, we find that soft-thresholding is only helpful for the non-binary classification tasks (i.e., CIFAR-10 and ImageNet, but not CIFAR-2 and QNLI); intuitively, this may be due to the fact that the underlying model output function depends on fewer examples (i.e., the attribution vector is sparser) when there are more classes.

For both CIFAR-10 and ImageNet, we use a single sparsity threshold---i.e., for each test example, we choose the soft-thresholding parameter $\lambda$ s.t. the resulting \trak score vector has exactly $k$ non-zero entries, and use the same $k$ for all test examples. To choose $k$, for CIFAR-10 we cross-validate using the same $M$ models that we used to compute \trak scores, when $M \ge 20$; in other words, we avoid ``cheating'' by using additional models for cross-validation.
For ImageNet, we simply choose $k=1000$ since there are on average 1,300 training examples per class.

\subsection{Baselines}
\label{app:baselines}

We provide details on baselines used in our evaluation in \Cref{sec:eval}.
Though most of the existing approximation-based methods only use a single model checkpoint in their original formulation, we average the methods over multiple independent checkpoints to help increase its performance.

\paragraph{Influence functions.}
The standard Hessian-based influence functions yield the attribution scores
\[ \tau(z_j)_i = \nabla L(z_j;\thetastar) \; H_{\thetastar}^{-1} \; \nabla L(z_i; \thetastar), \]
where $H_{\thetastar}$ is the empirical Hessian w.r.t. the training set.
We use an existing PyTorch implementation\footnote{\url{https://github.com/alstonlo/torch-influence}} that uses the stochastic approximation of inverse-Hessian-vector products using the \textsc{LiSSA} \cite{agarwal2017second} algorithm as done in \citet{koh2017understanding}.
As in the original work, we compute the gradients only with respect to the last linear layer; using additional layers caused the inversion algorithm to either diverge or to run out of memory.
For hyperparameters, we use similar values as done in prior work; we use $r=1$, $d=5000$, and damping factor of 0.01. We find that additional repeats ($r$, the number of independent trials to average each iHvp estimate) does not help, while increasing the depth ($d$, the number of iterations used by \textsc{LiSSA}) helps significantly.

\paragraph{Influence functions based on the Arnoldi iteration.}
This variant of influence functions from \citet{schioppa2022scaling} is based on approximating the top eigenspace of the Hessian using the Arnoldi iteration \cite{arnoldi1951principle}. We use the original implementation in \textsc{JAX}.\footnote{\url{https://github.com/google-research/jax-influence}}  We normalize the gradients as recommended in the original paper.
While much faster than the original formulation in \citet{koh2017understanding}, we find that the attribution scores not very predictive (according to the LDS).

\paragraph{TracIn.}
We use the \tracincp estimator from \citep{pruthi2020estimating}, defined as
\[ \tau(z_j)_i = \sum\limits_{t=1}^T \eta_t \cdot \nabla L(z_j;\theta_t) \cdot  \nabla L(z_i; \theta_t), \]
where $\theta_t$ is the checkpoint from the epoch $t$ and $\eta_t$ is the corresponding learning rate $\eta_t$.
We also average over trajectories of multiple independently trained models, which increases its performance. We approximate the dot products using random projections of dimensions 500-1000 as we do for \trak, as the estimator is intractable otherwise.
We found that increasing the number of samples (epochs) from the training trajectory does not lead to much improvement.

\paragraph{Gradient Aggregated Similarity (\gas).}
This is a ``renormalized'' version of the TracInCP \cite{hammoudeh2022training} based on using the cosine similarity instead of raw dot products.
In general, its performance is indistinguishable from that of \tracin.

\paragraph{Representation similarity.}
We use the {\em signed} $\ell_2$ dot product in representation space (feature embeddings of the penultimate layer), where the sign indicates whether the labels match.
We also experimented with cosine similarity but the resulting performance was similar.

\paragraph{Empirical influences.}
We use the subsampling-based approximation to leave-one-out influences as used by \cite{feldman2020neural}, which is a difference-in-means estimator given by
\[ \tau(z_j)_i = \mathbb{E}_{S \ni z_i} \modeleval{z_j}{\theta}
- \mathbb{E}_{S \not\ni z_i} \modeleval{z_j}{\theta} \]
where the first (second) expectation is over training subsets that include (exclude) example $z_i$.

\paragraph{Datamodels.}
We use the $\ell_1$-regularized regression-based estimators from \citet{ilyas2022datamodels}, using up to 60,000 models for \cifartwo and 300,000 models for \cifarten (trained on different random 50\% subsets of the full training set).

\subsection{Hardware and wall-time measurements}
\label{app:wall_time}
For all of our experiments, we use NVIDIA A100 GPUs each with 40GB of memory and 12 CPU cores.
We evaluate the computational cost of attribution methods using two metrics, {\em total wall-time} and the {\em total number of trained models used}; see \Cref{sec:eval} for motivation behind these metrics.
For most attribution methods, one or more of the following components dominate their total runtime:
\begin{itemize}
    \item \texttt{TRAIN\_TIME}: the time to train one model (from scratch)
    \item \texttt{GRAD\_TIME}: the time to compute gradients of one model (including computing random projections) for the entire dataset under consideration (both train and test sets). This time may vary depending on size of the projection dimension, but our fast implementation (\Cref{app:impl}) can handle dimensions of up to 80,000 without much increase in runtime.
\end{itemize}

\noindent The total compute time for each method was approximated as follows, where $M$ is the number of models used:
\begin{itemize}
\item {\bf \trak:} $M \times (\texttt{TRAIN\_TIME} + \texttt{GRAD\_TIME})$, as we have to compute gradients for each of the trained models.
\item {\bf Datamodel \cite{ilyas2022datamodels} and Empirical Influence \cite{feldman2020neural}:} $M \times \texttt{TRAIN\_TIME}$. The additional cost of estimating datamodels or influences from the trained models (which simply involves solving a linear system) is negligible compared to the cost of training.
\item {\bf \textsc{LiSSA} based influence functions \cite{koh2017understanding}:} These approaches are costly because they use thousands of Hessian-vector product iterations to approximate a single inverse-Hessian-vector product (which is needed for each target example).
Hence, we computed these attribution scores for a much smaller sample of validation set (50 to 100). We measured the empirical runtime on this small sample and extrapolated to the size of the entire (test) dataset.
\item {\bf Influence function based on the Arnoldi iteration \cite{schioppa2022scaling}:} We ran the authors' original code\footnote{\url{https://github.com/google-research/jax-influence}} on \cifar models of the same architecture (after translating them to \texttt{JAX}) and measured the runtime.
\item {\bf \tracin \cite{pruthi2020estimating} and \gas \cite{hammoudeh2022identifying}:} $M \times (\texttt{TRAIN\_TIME} + \texttt{GRAD\_TIME} \times T)$, where $T$ is the number of checkpoints used per model.
\end{itemize}

\clearpage
\section{\trak implementation}
\label{app:impl}
We release an easy-to-use library, \texttt{trak},\footnote{\url{https://github.com/MadryLab/trak}}, which computes \trak scores using \Cref{alg:estimator_pseudo}.
Computing \trak involves the following four steps:
(i) training models (or alternatively, acquiring checkpoints), (ii) computing gradients, (iii) projecting gradients with a random projection matrix (Rademacher or Gaussian), and (iv) aggregating into the final estimator (\Cref{eq:multi_model_trak}).

Step (i) is handled by the user, while steps (ii)-(iv) are handled automatically by our library. Step (ii) is implemented using the \texttt{functorch} library to compute per-example gradients. Step (iii) is either implemented using matrix multiplication on GPU or by a faster custom CUDA kernel, which is described below.
Step (iv) just involves a few simple matrix operations.

\subsection{Fast random projections on GPU}
One of the most costly operation of \trak is the random projection of the
gradients onto a smaller, more manageable vector space. While CPUs are not
equipped to handle this task on large models (e.g., LLMs) at sufficient
speed, at least on paper, GPUs have more than enough raw compute.

In practice, however, challenges arise. First, storing the projection matrix
entirely is highly impractical. For example, a matrix for a model with
300 million weights and an output of 1024 dimensions would require in excess of
1TB of storage. One solution is to generate the projection in blocks (across the output dimension). This solution is possible (and offered in our
implementation) but is still radically inefficient. Indeed, even if the
generation of the matrix is done by block it still has to be read and written
once onto the GPU RAM. This severely limits the performance as memory throughput
becomes the bottleneck.

\paragraph{Our approach.}
Our solution is to generate the coefficients of the projection as needed (in
some situations more than once) and never store them. As a result, the
bandwidth of the RAM is solely used to retrieve the values of the gradients and
write the results at the end. This forces us to use pseudo-randomness but this is actually preferrable since a true random matrix would make experiments impossible to
reproduce exactly.

Our implementation is written in C++/CUDA and targets NVIDIA GPUs of compute
capability above or equal 7.0 (V100 and newer). It supports (and achieve better
performance) batches of multiple inputs, and either normally distributed
coefficients or {-1, 1} with equal probabilities.

\paragraph{Implementation details.}
We decompose the input vectors into $K$ blocks, where each block is projected independently to increase parallelism. The final result is obtained by summing each partial projection. To reduce memory usage, we keep $K$ to roughly 100.

We further increase parallelism by spawning a thread for each entry of the output blocks, but this comes at the cost of reading the input multiple times. To mitigate this issue, we use Shared Memory offered by GPUs to share and reduce the frequency of data being pulled from global memory. We also use Shared Memory to reduce the cost of generating random coefficients, which can be reused for all the inputs of a batch.

Finally, we take advantage of Tensor Cores to maximize throughput and efficiency, as they were designed to excel at matrix multiplications. These interventions yield a fast and power-efficient implementation of random projection. On our hardware, we achieved speed-ups in excess of 200x compared to our ``block-by-block'' strategy.

\clearpage
\section{Theoretical Justification}
\label{app:theory_extra}
\subsection{The one-step Newton approximation for leave-one-out influence}
\label{app:theory_newton}
The key formula we use in \trak is the estimate for the leave-one-out (LOO) influence in logistic regression (\Cref{lem:formal}).
Here, we reproduce the derivation of this estimate from \citet{pregibon1981logistic} then extend it to incorporate example-dependent bias terms. %

\paragraph{Convergence condition for logistic regression.}
Assume that we optimized the logistic regression instance via Newton-Raphson, i.e., the parameters are iteratively updated as
\begin{equation}
    \thetahat_{t+1} \leftarrow \thetahat_{t} + H_{\thetahat_t}^{-1} \nabla_\theta {L}(\thetahat_t) \label{eqn:newton_general}
\end{equation}
where $H_{\thetahat}$ is the Hessian and $\nabla_\theta {L}(\thetahat)$ is the gradient associated with the total training loss ${L}(\thetahat) = \sum_{z_i \in S} L(z_i;\theta)$.
In the case of logistic regression, the above update is given by
\begin{equation}
    \thetahat_{t+1} \leftarrow \thetahat_t + (X^\top R X)^{-1} X^\top \hat{{q}} \label{eqn:newton_logistic}
\end{equation}
where $\hat{{q}} = \vec{1} - \hat{{p}}$ is the vector of the probabilities for the {\em incorrect} class evaluated at $\thetahat_t$ and $R = \text{\normalfont diag}(\hat{p} (1 -\hat{p})$ is the corresponding matrix.
Upon convergence, the final parameters $\thetastar$ satisfy the following:
\begin{equation}
    (X^\top R X)^{-1} X^\top {q}^\star = 0 \label{eqn:opt}
\end{equation}
where ${q}^\star$ is the incorrect-class probability vector corresponding to $\thetastar$.

\paragraph{The one-step Newton approximation.}
We estimate the counterfactual parameters $\thetastar_{-i}$ that would have resulted from training on the same training set excluding example $i$ by simply taking a single Newton step starting from the same global optimum $\thetastar$:
\begin{equation}
\thetastar_{-i} = \thetastar + (X_{-i}^\top R_{-i} X_{-i})^{-1} X_{-i}^\top {q}_{-i}^\star,
\end{equation}
where the subscript $-i$ denotes the corresponding matrices and vectors without the $i$-th training example.
Rearranging and using (\ref{eqn:opt}),
\begin{align*}
    \thetastar - \thetastar_{-i} &= -  (X_{-i}^\top R_{-i} X_{-i})^{-1} X_{-i}^\top {q}_{-i}^\star\\
    \thetastar - \thetastar_{-i} &= (X^\top R X)^{-1} X^\top {q}^\star -  (X_{-i}^\top R_{-i} X_{-i})^{-1} X_{-i}^\top {q}_{-i}^\star
\end{align*}
\noindent Using the Sherman–Morrison formula to simplify above,\footnote{This is used also, for instance, to derive the LOO formulas for standard linear regression.} we have
\begin{equation}
    \thetastar - \thetastar_{-i} = \frac{(X^\top R X)^{-1} x_i}{1 - x_i^\top (X^\top R X)^{-1} x_i \cdot p_i^\star (1 - p_i^\star)} {q}^\star_i = \frac{(X^\top R X)^{-1} x_i}{1 - x_i^\top (X^\top R X)^{-1} x_i \cdot p_i^\star (1 - p_i^\star)} (1 - {p}^\star_i)
\end{equation}
The above formula estimates the change in the parameter vector itself. To estimate the change in prediction at a given example $x$, we take the inner product of the above expression with vector $x$ to get the formula in \Cref{lem:formal}.

The approximation here is in assuming the updates converge in one step. Prior works \cite{koh2019accuracy} quantify the fidelity of such approximation under some assumptions.
The effectiveness of \trak across a variety of settings suggests that the approximation is accurate in regimes that arise in practice.

\paragraph{Incorporating bias terms.}
The above derivation is commonly done for the case of standard logistic regression, but it also directly extends to the case where the individual predictions incorporate example-dependent bias terms $b_i$ that are independent of $\theta$. In particular, note that the likelihood function after linearization in Step 1 is given by
\begin{equation}
    p(z_i; \theta) = \sigma(-y_i \cdot (\nabla_\theta \modeleval{z_i}{\thetastar} \cdot \theta + b_i))
\end{equation}
where $\sigma (\cdot)$ is the sigmoid function.
Because the Hessian and the gradients of the training loss only depend on $\theta$ through $p(z_i; \theta)$, and because $b_i$'s are independent of $\theta$, the computation going from
\cref{eqn:newton_general} to \cref{eqn:newton_logistic} is not affected.
The rest of the derivation also remains identical as the bias terms are already incorporated into $p^\star$ and ${q}^\star$.

\paragraph{Generalization to other settings.}
\label{app:general_model_output}
While our derivations in this paper focus on the case of logistic regression, more generally, \trak can be easily adapted to any choice of model output function as long as the training loss $L$ is a convex function of the model output $f$.
The corresponding entries in the $\mathbf{Q}=\text{diag}(1-p^\star_i)$ matrix in \Cref{lem:formal} is then replaced by $\partial L / \partial f (z_i)$.
The $R$ matrix and the leverage scores also change accordingly, though we do not include them in our estimator (that said, including them may improve the estimator in settings beyond classification).

However, in general one needs care in choosing an appropriate model output function in order to maximize the performance on the linear datamodeling prediction task. If the chosen model output is not well approximated by a linear function of training examples, then that puts an upper bound on the predictive performance of {\em any} attribution method in our framework. We discuss appropriate choices of model output functions further in \Cref{app:linear}.

\subsection{Random projections preserve gradient flow}
\label{app:theory_jl}

In Step 2 of \trak, we use random projections to reduce the dimension of the gradient vectors. Here, we justify this approximation when our model is trained via gradient descent. Similar analysis has been used prior, e.g., by \citet{malladi2022kernel}.

In the limit of small learning rate, the time-evolution of model output $f(z;\theta)$ under gradient descent (or gradient flow) is captured by the following differential equation \citep{jacot2018neural}:
\begin{align}
    \frac{df(z;\theta)}{dt} = \sum\limits_i \frac{\partial \loss{z_i}}{\partial f(z_i; \theta)} \cdot (\nabla f(z_i;\theta) \cdot \nabla f(z;\theta)) \approx  \sum\limits_i \frac{\partial \loss{z_i}}{\partial f(z_i;\theta)} \cdot (g_i \cdot g(z)) \label{eqn:ode}
\end{align}
where $g_i$ and $g(z)$ are the gradients of the final model corresponding to examples $z_i$ and $z$ as before. The approximation is due to assuming that the gradients do not change over time.

If we treat the outputs $\{\hat{f}(z_i;\theta)\}_i$ as time-varying variables, then their time evolution is entirely described by the above system of differential equations (one for each $i$, replacing $z$ with $z_i$ above).
Importantly, the above equations only depend on the gradients through their inner products. Hence, as long as we preserve the inner products to sufficient accuracy, the resulting system has approximately the same evolution as the original one. This justifies replacing the gradient features with their random projections.

\subsection{Subsampling the training set}
\label{app:theory_alpha}

In Step 4 of our algorithm, we ensemble the attribution scores over multiple models.  As
we investigate in \Cref{app:ablation_num_models}, this significantly improves \trak's performance. An important design choice is training each model on a different random
subset of the training set.

This choice is motivated by the following connection between \trak scores and empirical influences \cite{feldman2020neural}.
Recall that we designed \trak to optimize the linear datamodeling score.
As we discuss in \Cref{sec:prelim}, datamodels can be viewed as an ``oracle'' for optimizing the same metric.
Further, as \citet{ilyas2022datamodels} observes, datamodels can be viewed as a regularized version of empirical influences \cite{feldman2020neural}, which are defined as a difference-in-means estimator,
\begin{align}
    \tau(z_j)_i &= \mathbb{E}_{S' \sim \mathcal{D}}[f(z_j;\thetastar(S'))|
    z_i \in S']
    - \mathbb{E}_{S' \sim \mathcal{D}}[f(z_j;\thetastar(S'))|
    z_i \not\in S']
\end{align}
where $\mathcal{D}$ is the uniform distribution over $\alpha$-fraction subsets of training set $S$. Assuming the expectation over $\alpha$-fraction subsets is identical to that over subsets of one additional element, we can rearrange the above expression as
\begin{align}
    \tau(z_j)_i &= \mathbb{E}_{S' \sim \mathcal{D}}[f(z_j;\thetastar(S' \cup \{z_i\})) - f(z_j;\thetastar(S'))].
\end{align}
The above expression is simply the expectation of leave-one-out influence over different random subsets. As the estimate from step 3 of our algorithm is specific to a single training set, we need to average over different subsets in order to approximate the above quantity.

In principle, the estimates computed from $\thetastar(S')$ only apply to the training examples included in the subset $S'$, since the underlying formula (\Cref{lem:formal}) concerns examples that were included for the original converged parameter $\thetastar$. Hence, when averaging over the models, each model should only update the \trak scores corresponding to examples in $S'$. However, we found that the estimates are marginally better when we update the estimates for the entire training set $S$ (i.e., even those that were not trained on).

\paragraph{Generalization across different $\alpha$'s.}
A possible concern is that we overfit to a particular regime of $\alpha$ used in evaluating with the LDS. In \Cref{fig:jointplot}, we evaluate \trak scores (computing using $\alpha=0.5$) in other regimes and find that they continue to be highly predictive (though with some degradation in correlation).
More generally, our various counterfactual evaluations using the full training set (CIFAR-10 brittleness estimates in \Cref{fig:brittle}, the CLIP counterfactuals in \Cref{fig:CLIP_barplot_counterfactuals}) indicate that \trak scores remain predictive near the $\alpha=1$ regime.

\subsection{Linearity and model output function}
\label{app:linear}

We study linear predictors derived from attribution scores, as linearity is a latent assumption for many popular attribution methods.
Linearity also motivates our choices of model output functions.

\paragraph{Latent assumption of linearity.}
Our evaluation of data attribution methods cast them as linear predictors.
While not always immediate, linearity is a latent assumption behind most of the prior methods that we evaluate in this paper.
Datamodels and Shapley values satisfy additivity by construction \cite{ghorbani2019data,jia2019towards}.
The approach based on influence functions \cite{koh2017understanding,koh2019accuracy} typically uses the sum of LOO influences to estimate influences for groups of examples.
Similarly, empirical (or subsampled) influences \cite{feldman2020neural} also correspond to a first-order Taylor approximation of the model output function. The \tracin estimator also implicitly assumes linearity \citep{pruthi2020estimating}.

That said, others works also incorporate additional corrections beyond the first order linear terms \cite{basu2019second} and find the resulting predictions better approximate the true influences.

\paragraph{Choice of model output function $f$.}
In our experiments, we choose the model output function suitable for the task at hand: for classification and language modeling, we used a notion of margin that is equivalent to the logit function, while for CLIP, we used a similar one based on the CLIP loss.

Our particular choice of the logit function ($\log p/(1-p)$) in the multi-class classification case was motivated by theoretical \cite{saunshi2023understanding} and empirical \cite{ilyas2022datamodels} observations from prior works.
In particular, this choice of model output function is well approximated by {\em linear} datamodels, both in practice and in theory.
A slightly different definition of margin used in \citet{ilyas2022datamodels}---where the margin is computed as the logit for the correct class minus the second highest class---can also be viewed as an approximation to the one used here.

More generally, choosing a good $f$ boils down to linearizing (w.r.t. $\theta$) as much of the model output as possible, but not too much. On one extreme, choosing $f(z) = z$ (i.e., linearizing nothing, as there is no dependence on $\theta$)
means that the one-step Newton approximation has to capture all of the non-linearity in both the model and the dependence of $L$ on $f$; this is essentially the same approximation used by the Hessian-based influence function.  On the other extreme, if we choose $f = L$, we linearize too much, which does not work well as $L$ in general is highly non-linear as a function of $f$.

\clearpage
\section{Additional Results}
\label{app:more_results}

\subsection{Correlation distribution}

\begin{figure}[!htbp]
    \centering
    \includegraphics[width=0.45\linewidth]{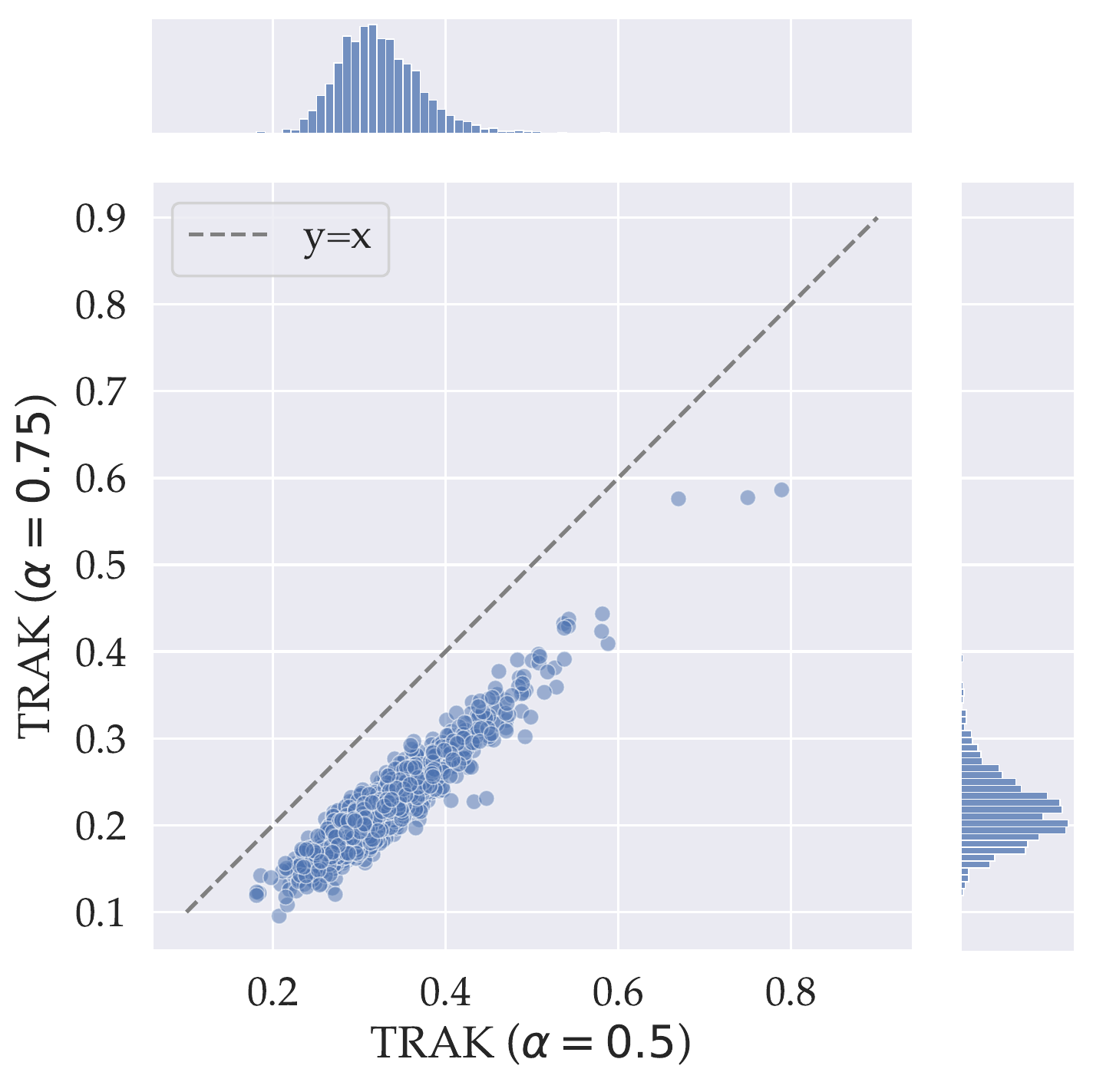}
    \includegraphics[width=0.45\linewidth]{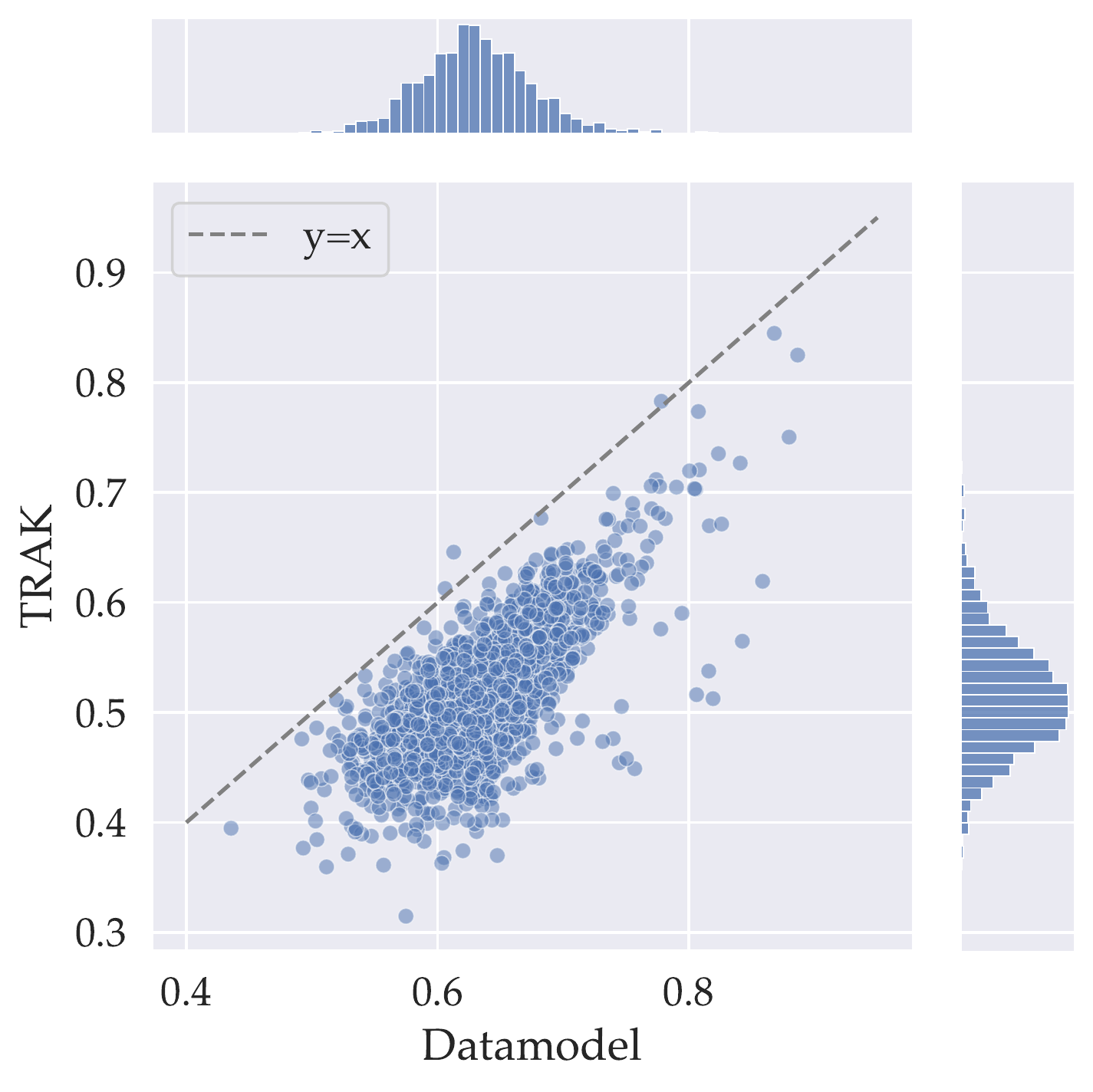}
\caption{{\bf (Left)} The LDS of \cifartwo \trak scores computed with $\alpha=0.5$ models then evaluated on either models trained with either $\alpha=0.5$ or $\alpha=0.75$. Each point corresponds to a validation example. {\bf (Right)} The LDS of \cifartwo datamodel scores compared with that of \trak. Here, the LDS is measured on two different estimators.}
\label{fig:jointplot}
\end{figure}

\paragraph{Generalization across $\alpha$'s.} In \Cref{fig:jointplot} left, we compare the linear datamodeling scores (LDS) evaluated on $\alpha=0.5$ sub-sampled training sets to those evaluated on $\alpha=0.75$.
(The numbers are overall lower as these are evaluated on data where only one model was trained on each subset,instead of averaging over 5 models; hence, there is more noise in the data.) As we observe, the LDS scores on different $\alpha$'s are highly correlated, suggesting that \trak scores computed on a single $\alpha$ generalize well.

\paragraph{LDS correlation between \trak and datamodels.} In \Cref{fig:jointplot} right, we compare the LDS correlations of datamodels to that of \trak and find that they are correlated across examples; in general, \trak also performs better on examples on which datamodels perform better.

\clearpage
\subsection{Table for LDS evaluation}

\begin{table}[h]
    \centering
    \begin{tabular}{llrrrrrrrrr}
        \toprule
        Dataset & & TRAK & TracIn \citep{pruthi2020estimating} & Infl. \citep{koh2017understanding} & Datamodels \citep{ilyas2022datamodels} \\
        \midrule
        \cifartwo & \# models & 5 & 100 & - & 1,000 \\
        & Time (min.) & 3 & 100 & - & 500  \\
        & LDS & {\bf 0.203(3)} & 0.056(2) & - & 0.162(5)  \\
        \midrule
        \cifarten & \# models & 20 & 20 & 1 & 5,000 \\
        & Time (min.) & 20 & 60 & 20,000 & 2,500 \\
        & LDS & {\bf 0.271(4)} & 0.056(7) & 0.037(13) & 0.199(4) \\
        \midrule
        \qnli & \# models & 10 & 1 &  1 & 20,000 \\
        & Time (min.) & 640 & 284 & 18,000 & 176,000 \\
        & LDS & {\bf 0.416(10)} & 0.077(29) & 0.114(43) & 0.344(32) \\
        \midrule
        ImageNet & \# models & 100 & 1 &  20 & 30,000 \\
        & Time (min.) & 2920 & 76 & $>$100,000 &   525,000  \\
        & LDS & {\bf 0.188(6)} & 0.008(6) &   0.037(6) & 0.1445(6) \\
        \bottomrule
        \end{tabular}
        \caption{{\em Comparison of different data attribution methods.} We quantify various data attribution methods in terms of both their {\em predictiveness}---as
        measured by the linear datamodeling score---as well as their {\em
        computational efficiency}---as measured by either the total computation
        time (wall-time measured in minutes on a single A100 GPU; see
        \Cref{app:wall_time} for details) or the number of trained models used
        to compute the attribution scores. The errors indicate 95\%
        bootstrap confidence intervals.
        Sampling-based methods (datamodels and
        empirical influences) can outperform \trak when allowed to use more
        computation, but this leads to a significant
        increase in computational cost.
        }
        \label{tab:all_best}
\end{table}

\clearpage
\subsection{\trak examples}
\label{app:more_examples}
\begin{figure}[!b]
    \centering
    \includegraphics[width=.9\linewidth,trim={0 0 0 0},clip]{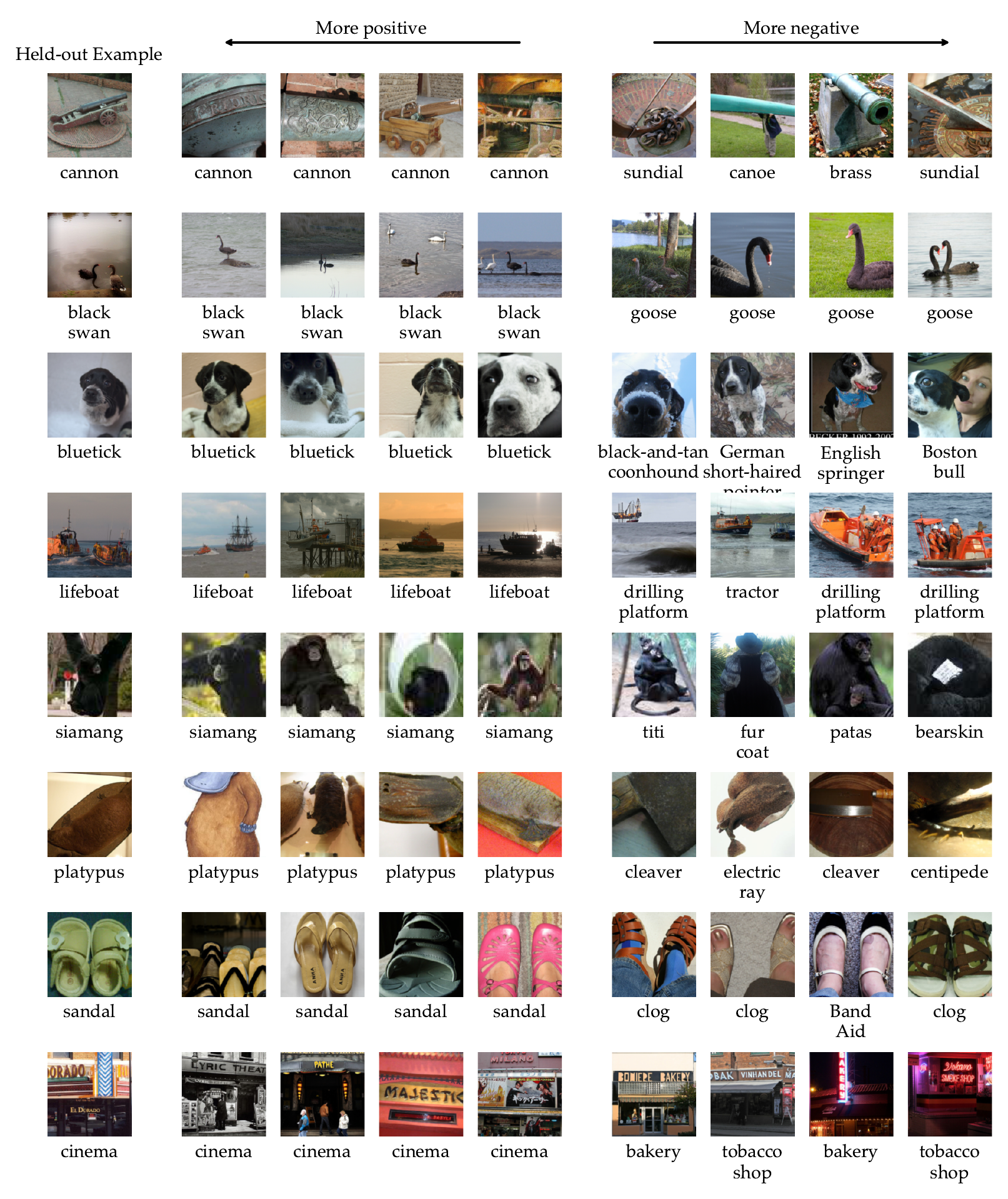}
\caption{
     {\em \trak attributions for ResNets trained on ImageNet.}
    We display random test examples and their corresponding
    most helpful (highest-scoring) and most detracting (lowest-scoring)
    training examples according to \trak.
}
\label{fig:imagenet_nns_extra}
\end{figure}

\begin{figure}[!t]
    \centering
    \includegraphics[width=\linewidth,trim={0 0 0 0},clip]{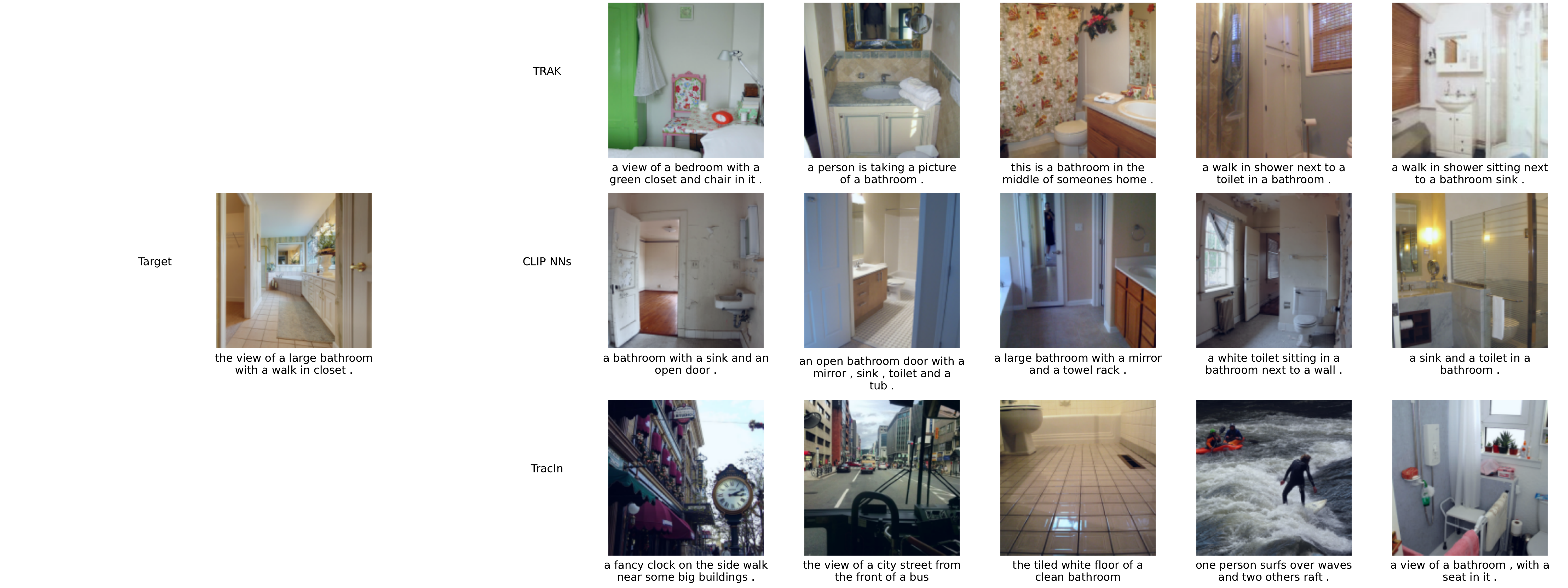}
    \includegraphics[width=\linewidth,trim={0 0 0 0},clip]{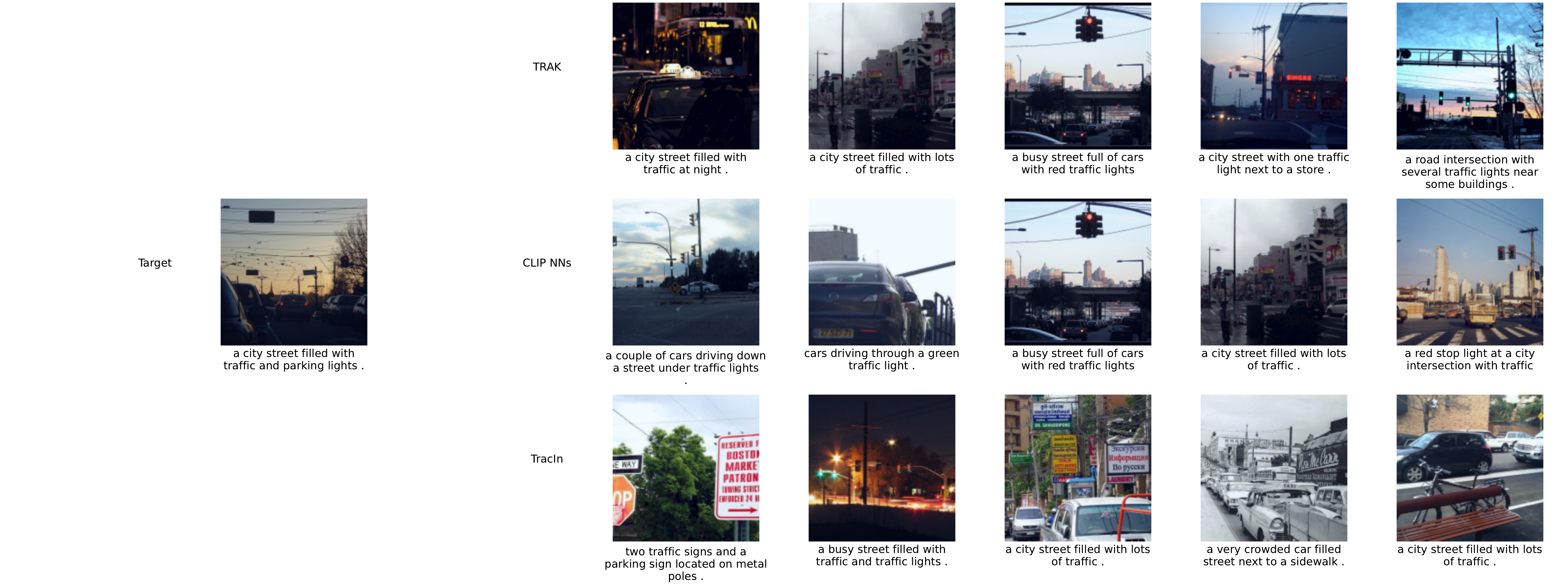}
    \includegraphics[width=\linewidth,trim={0 0 0 0},clip]{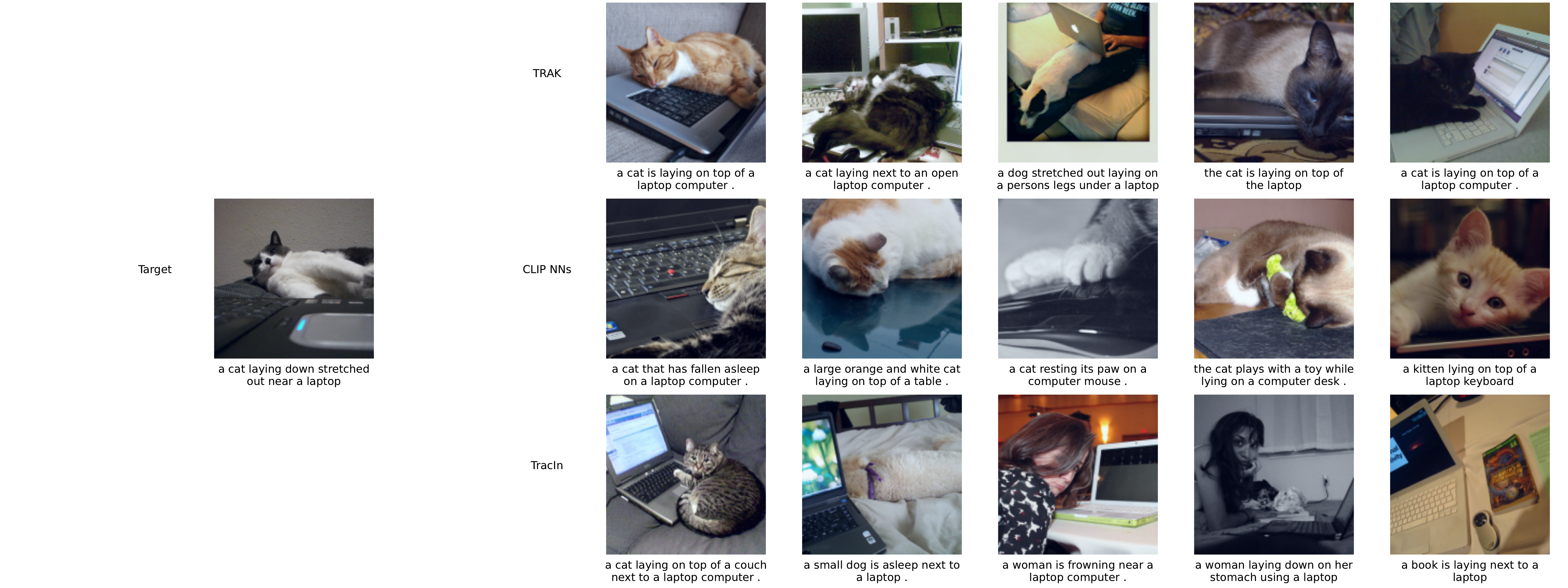}
\caption{
     {\em Top attributions for \clip models trained on \mscoco.}
    We display random test examples and their corresponding
    most helpful (highest-scoring) and most detracting (lowest-scoring)
    training examples according to \trak, \clip similarity distance, and \tracin.
    }
\label{fig:clip_examples_extra}
\end{figure}

We display more examples identified with \trak scores in \Cref{fig:imagenet_nns_extra} (ImageNet), \Cref{tab:qnli_more} (\qnli), and \Cref{fig:clip_examples_extra} (\clip on \mscoco).

\clearpage
\begin{figure}
    \centering
    \begin{tabular}{p{0.33\textwidth}p{0.30\textwidth}p{0.30\textwidth}}
    \toprule
    \textbf{Example} & \textbf{Highest \trak score (+)} & \textbf{Lowest \trak score (-)} \\
    \midrule
    \scriptsize {\bf Q:} What was a major success, especially in rebuilding Warsaw? {\bf A:} Like many cities in Central and Eastern Europe, infrastructure in Warsaw suffered considerably during its time as an Eastern Bloc economy – though it is worth mentioning that the initial Three-Year Plan to rebuild Poland (especially Warsaw) was a major success, but what followed was very much the opposite. {\bf (Yes)} & \scriptsize {\bf Q:} In 1998, the deal was renewed for what amount over four years? {\bf A:} Television money had also become much more important; the Football League received £6.3 million for a two-year agreement in 1986, but when that deal was renewed in 1988, the price rose to £44 million over four years. {\bf (Yes)} & \scriptsize {\bf Q:} Who was a controversial figure due to a corked-bat incident? {\bf A:} Already a controversial figure in the clubhouse after his corked-bat incident, Sammy's actions alienated much of his once strong fan base as well as the few teammates still on good terms with him, (many teammates grew tired of Sosa playing loud salsa music in the locker room) and possibly tarnished his place in Cubs' lore for years to come. {\bf (No)} \\
    \midrule
    \scriptsize {\bf Q:} What is the name associated with the eight areas that make up a part of southern California? {\bf A:} Southern California consists of one Combined Statistical Area, eight Metropolitan Statistical Areas, one international metropolitan area, and multiple metropolitan divisions. {\bf (Yes)} & \scriptsize {\bf Q:} Was was the name given to the Alsace provincinal court? {\bf A:} The province had a single provincial court (Landgericht) and a central administration with its seat at Hagenau. {\bf (Yes)} & \scriptsize {\bf Q:} What do six of the questions asses? {\bf A:} For each question on the scale that measures homosexuality there is a corresponding question that measures heterosexuality giving six matching pairs of questions. {\bf (No)} \\
    \midrule
    \scriptsize {\bf Q:} What words are inscribed on the mace of parliament? {\bf A:} The words There shall be a Scottish Parliament, which are the first words of the Scotland Act, are inscribed around the head of the mace, which has a formal ceremonial role in the meetings of Parliament, reinforcing the authority of the Parliament in its ability to make laws. {\bf (No)} & \scriptsize {\bf Q:} Whose name is on the gate-house fronting School Yard? {\bf A:} His name is borne by the big gate-house in the west range of the cloisters, fronting School Yard, perhaps the most famous image of the school. {\bf (No)} & \scriptsize {\bf Q:} What kind of signs were removed form club Barcelona? {\bf A:} All signs of regional nationalism, including language, flag and other signs of separatism were banned throughout Spain. {\bf (Yes)} \\
    \midrule
    \scriptsize {\bf Q:} What was the percentage of a female householder with no husband present? {\bf A:} There were 158,349 households, of which 68,511 (43.3\%) had children under the age of 18 living in them, 69,284 (43.8\%) were opposite-sex married couples living together, 30,547 (19.3\%) had a female householder with no husband present, 11,698 (7.4\%) had a male householder with no wife present. {\bf (Yes)} & \scriptsize {\bf Q:} What percent of household have children under 18? {\bf A:} There were 46,917 households, out of which 7,835 (16.7\%) had children under the age of 18 living in them, 13,092 (27.9\%) were opposite-sex married couples living together, 3,510 (7.5\%) had a female householder with no husband present, 1,327 (2.8\%) had a male householder with no wife present. {\bf (Yes)} & \scriptsize {\bf Q:} Roughly how many same-sex couples were there? {\bf A:} There were 46,917 households, out of which 7,835 (16.7\%) had children under the age of 18 living in them, 13,092 (27.9\%) were opposite-sex married couples living together, 3,510 (7.5\%) had a female householder with no husband present, 1,327 (2.8\%) had a male householder with no wife present. {\bf (No)} \\
        \midrule
        \scriptsize {\bf Q:} What did Warsz own? {\bf A:} In actuality, Warsz was a 12th/13th-century nobleman who owned a village located at the modern-day site of Mariensztat neighbourhood. {\bf (Yes)} & \scriptsize {\bf Q:} What company did Ray Kroc own? {\bf A:} It was founded in 1986 through the donations of Joan B. Kroc, the widow of McDonald's owner Ray Kroc. {\bf (Yes)} & \scriptsize {\bf Q:} What did Cerberus guard? {\bf A:} In Norse mythology, a bloody, four-eyed dog called Garmr guards Helheim. {\bf (No)} \\
        \midrule
        \scriptsize {\bf Q:} What words are inscribed on the mace of parliament? {\bf A:} The words There shall be a Scottish Parliament, which are the first words of the Scotland Act, are inscribed around the head of the mace, which has a formal ceremonial role in the meetings of Parliament, reinforcing the authority of the Parliament in its ability to make laws. {\bf (No)} & \scriptsize {\bf Q:} Whose name is on the gate-house fronting School Yard? {\bf A:} His name is borne by the big gate-house in the west range of the cloisters, fronting School Yard, perhaps the most famous image of the school. {\bf (No)} & \scriptsize {\bf Q:} What kind of signs were removed form club Barcelona? {\bf A:} All signs of regional nationalism, including language, flag and other signs of separatism were banned throughout Spain. {\bf (Yes)} \\
    \bottomrule
\end{tabular}
\caption{{\em Top \trak attributions for \qnli examples.} Yes/No indicates the label (entailment vs. no entailment).}
\label{tab:qnli_more}
\end{figure}

\clearpage

\clearpage
\subsection{\modeldiff with \trak}
\begin{figure}[h]
    \centering
    \includegraphics[width=\linewidth,trim={0 0 0 0},clip]{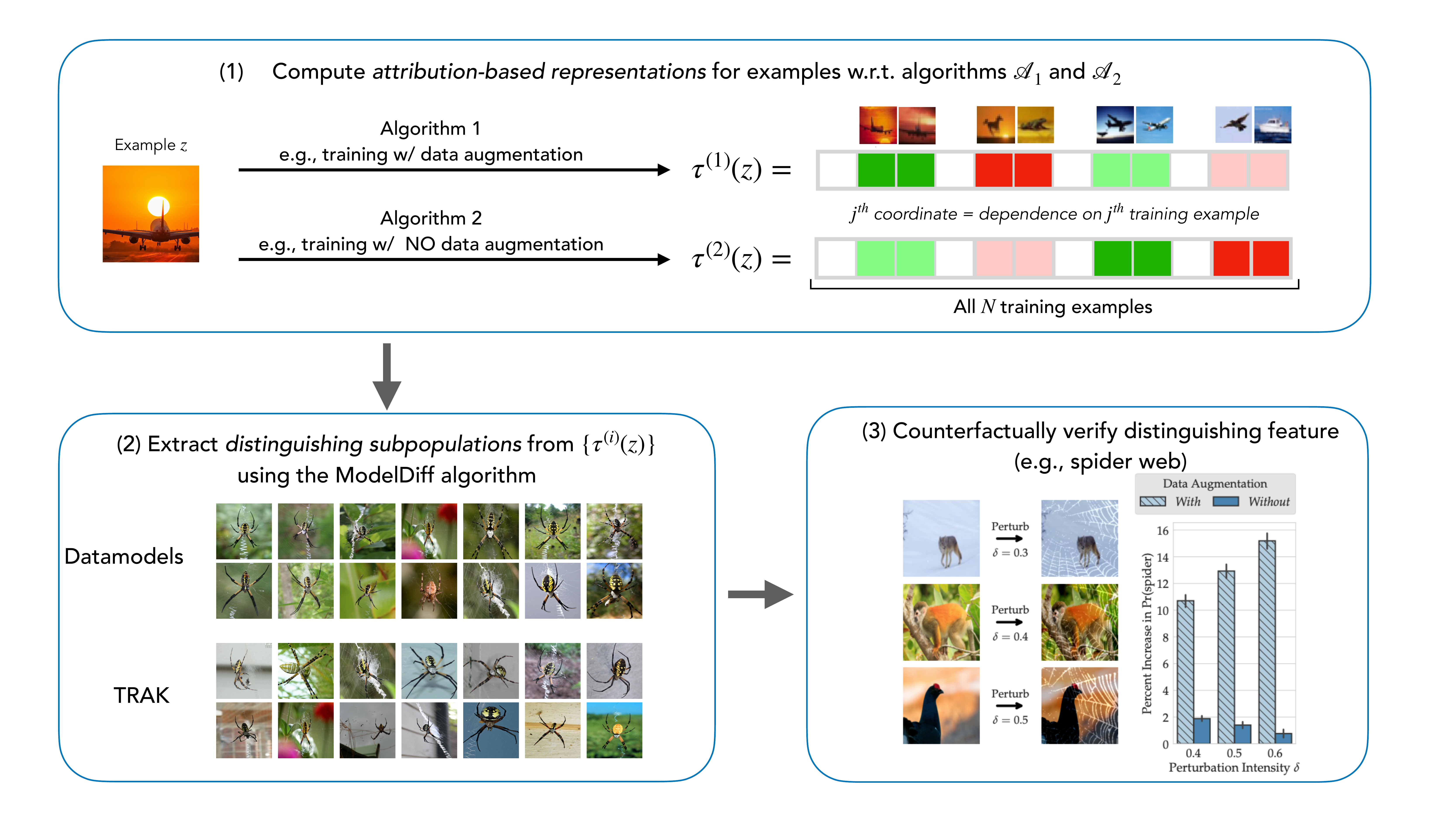}
\caption{
{\em Accelerating learning algorithm comparisons with \trak.}
The \modeldiff framework from \citep{shah2022modeldiff} uses datamodel
representations to surface features that distinguish two learning algorithms. In
the case study here, we compare models trained on the \textsc{Living17} dataset {\em with} and {\em without} data
augmentation. Applying \modeldiff involves three stages: (1) computing datamodel
representations; (2) applying the \modeldiff algorithm to extract {\em
distinguishing subpopulations} of inputs on which two model classes behave
differently; (3) counterfactually testing the inferred feature associated
with the subpopulation. \citet{shah2022modeldiff} find that models trained with
data augmentation latch onto the presence of spider webs as a spurious
correlation to predict the class spider. Here, we recover their result by using
\trak scores instead of datamodel scores in step (1); doing so reduces the
computational cost of \modeldiff by 100x.
}
\label{fig:modeldiff}
\end{figure}

\Cref{fig:modeldiff} shows how we apply \trak to dramatically accelerate the
\modeldiff algorithm.

\clearpage
\section{Ablation Studies}
\label{app:more_ablation}
We perform a number of ablation studies to understand how different
components of \trak affect its performance.
Specifically, we study the following:
\begin{itemize}
    \item The dimension of the random projection, $k$. %
    \cref{sec:estimator_algo}).
    \item The number of models ensembled, $M$. %
    \cref{sec:estimator_algo}).
    \item Proxies for ensembles to further improve \trak's computational
    efficiency.
    \item The role of different terms in the influence estimation formula (\cref{eq:trak}).
    \item Alternative choice of the kernel (using last layer representations).
    \item Alternative methods of ensembling over models.
\end{itemize}

\noindent As in \Cref{sec:eval}, we evaluate the linear datamodeling score (LDS) on models trained on the \cifartwo, \cifarten,
and \qnli datasets. Note that the LDS is in some cases lower than the counterparts in \Cref{fig:headline_full} as we use a smaller projected dimension ($k$) and do not use soft-thresholding in these experiments.

\subsection{Dimension of the random projection}

Recall that when we compute \trak we reduce the dimensionality of the gradient
features using random projections (Step 2 of \cref{sec:estimator_algo}).
Intuitively, as the resulting dimension $k$ increases, the corresponding
projection better preserves inner products, but is also more expensive to
compute. We now study how the choice of the projection dimension $k$ affects
\trak's attribution performance.

\Cref{fig:jl_dim} (Left) shows that as we increase the dimension, the
LDS initially increases as expected; random projections to a higher dimension
preserve the inner product more accurately, providing a better approximation of
the gradient features. However, beyond a certain point, increasing projection
dimension {\em decreases} the LDS. We hypothesize that using random projections
to a lower dimension has a regularizing effect that competes with the increase
in approximation error.\footnote{Indeed, we can view our approach of first
projecting features to a lower dimension and then performing linear regression
in the compressed feature space, as an instance of {\em compressed linear
regression} \cite{maillard2009compressed} and also related to principal
components regression \cite{thanei2017random}. These approaches are known to
have a regularizing effect, so \trak may also benefit from that effect.} Finally,
the dimension at which LDS peaks {\em increases} as we increase the number of
models $M$ used to compute \trak.

\begin{figure}[!htbp]
    \centering
    \pgfplotsset{scaled x ticks=false}
\begin{tikzpicture}

    \definecolor{darkgray141160203}{RGB}{141,160,203}
    \definecolor{dimgray85}{RGB}{85,85,85}
    \definecolor{gainsboro229}{RGB}{229,229,229}
    \definecolor{lightgray204}{RGB}{204,204,204}
    \definecolor{mediumaquamarine102194165}{RGB}{102,194,165}
    \definecolor{orchid231138195}{RGB}{231,138,195}
    \definecolor{salmon25214198}{RGB}{252,141,98}
    \definecolor{amethyst}{rgb}{0.6, 0.4, 0.8}
    \definecolor{bleudefrance}{rgb}{0.19, 0.55, 0.91}
    \definecolor{blush}{rgb}{0.87, 0.36, 0.51}
    \definecolor{brilliantrose}{rgb}{1.0, 0.33, 0.64}

    \begin{groupplot}[
      group style={group size= 2 by 1},
      height={5cm},
      width=.5\linewidth]
    \nextgroupplot[
    axis background/.style={fill=gainsboro229},
    axis line style={white},
    legend cell align={left},
    legend columns=4,
    legend style={
      fill opacity=0.8,
      draw opacity=1,
      text opacity=1,
      at={(-0.2,1.6)},
      anchor=north west,
      draw=gainsboro229,
      fill=gainsboro229,
      /tikz/every even column/.append style={column sep=0.5cm},
      /tikz/every odd column/.append style={column sep=0.1cm}
    },
    tick align=outside,
    tick pos=left,
    x grid style={white},
    align=center,
    title={ResNet-9 on CIFAR-2},
    xlabel={Random projection dimension $k$},
    xmajorgrids,
    xminorgrids,
    xmin=0, xmax=9500,
    xtick style={color=dimgray85},
    xticklabel style={/pgf/number format/fixed},
    xtick={0,2000,4000,6000,8000},
    y grid style={white},
    ylabel={Correlation (LDS) \\ (more accurate \(\displaystyle \rightarrow\))},
    ymajorgrids,
    ymin=0, ymax=0.6,
    ytick style={color=dimgray85},
    ]
    \addplot [draw=black,
             fill=black,
             fill opacity=0, draw opacity=0.1,
             mark options={scale=0.7, line width=1pt},
             mark=star] coordinates {
        (1000, 0.275)
        (2000, 0.289)
        (3000, 0.284)
        (4000, 0.272)
        (5000, 0.254)
        (6000, 0.232)
        (7000, 0.204)
        (8000, 0.170)
        (9000, 0.122)
    };
    \node[opacity=0.3] at (axis cs: 2600,.23) {10 models};
    \node[opacity=1.0] at (axis cs: 6000,.55) {100 models};

    \addplot [draw=black,          fill=black,          fill opacity=0, draw opacity=0.2,          mark options={scale=0.7, line width=1pt},          mark=star] coordinates {
        (1000, 0.342)
        (2000, 0.363)
        (3000, 0.362)
        (4000, 0.350)
        (5000, 0.332)
        (6000, 0.306)
        (7000, 0.274)
        (8000, 0.231)
        (9000, 0.169)
    };

    \addplot [draw=black,          fill=black,          fill opacity=0, draw opacity=0.3,          mark options={scale=0.7, line width=1pt},          mark=star] coordinates {
        (1000, 0.379)
        (2000, 0.406)
        (3000, 0.408)
        (4000, 0.398)
        (5000, 0.381)
        (6000, 0.355)
        (7000, 0.321)
        (8000, 0.275)
        (9000, 0.203)
    };

    \addplot [draw=black,
              fill=black,
              fill opacity=0,
              draw opacity=0.4,
              mark options={scale=0.7, line width=1pt},
              mark=star] coordinates {
        (1000, 0.401)
        (2000, 0.432)
        (3000, 0.437)
        (4000, 0.430)
        (5000, 0.414)
        (6000, 0.389)
        (7000, 0.356)
        (8000, 0.307)
        (9000, 0.230)
    };

    \addplot [draw=black,
              fill=black,
              draw opacity=0.5,
              fill opacity=0.,
              mark options={scale=0.7, line width=1pt},
              mark=star] coordinates {
        (1000, 0.417)
        (2000, 0.451)
        (3000, 0.459)
        (4000, 0.453)
        (5000, 0.439)
        (6000, 0.416)
        (7000, 0.383)
        (8000, 0.333)
        (9000, 0.253)
    };

    \addplot [draw=black,          fill=black,          fill opacity=0.0, draw opacity=0.6,          mark options={scale=0.7, line width=1pt},          mark=star] coordinates {
        (1000, 0.429)
        (2000, 0.465)
        (3000, 0.475)
        (4000, 0.471)
        (5000, 0.459)
        (6000, 0.437)
        (7000, 0.405)
        (8000, 0.355)
        (9000, 0.273)
    };

    \addplot [draw=black,          fill=black,          fill opacity=0.0, draw opacity=0.7,          mark options={scale=0.7, line width=1pt},          mark=star] coordinates {
        (1000, 0.438)
        (2000, 0.477)
        (3000, 0.488)
        (4000, 0.486)
        (5000, 0.475)
        (6000, 0.454)
        (7000, 0.423)
        (8000, 0.374)
        (9000, 0.290)
    };

    \addplot [draw=black,          fill=black,          fill opacity=0.0, draw opacity=0.8,          mark options={scale=0.7, line width=1pt},          mark=star] coordinates {
        (1000, 0.445)
        (2000, 0.485)
        (3000, 0.498)
        (4000, 0.497)
        (5000, 0.488)
        (6000, 0.469)
        (7000, 0.439)
        (8000, 0.390)
        (9000, 0.306)
    };

    \addplot [draw=black,          fill=black,          fill opacity=0.0, draw opacity=0.9,          mark options={scale=0.7, line width=1pt},          mark=star] coordinates {
        (1000, 0.451)
        (2000, 0.493)
        (3000, 0.507)
        (4000, 0.507)
        (5000, 0.499)
        (6000, 0.481)
        (7000, 0.452)
        (8000, 0.405)
        (9000, 0.320)
    };

    \addplot [draw=black,          fill=black,          fill opacity=0.0, draw opacity=1.0,          mark options={scale=0.7, line width=1pt},          mark=star] coordinates {
        (1000, 0.456)
        (2000, 0.499)
        (3000, 0.514)
        (4000, 0.515)
        (5000, 0.508)
        (6000, 0.491)
        (7000, 0.464)
        (8000, 0.418)
        (9000, 0.333)
    };

    \addplot [draw=mediumaquamarine102194165,
              fill=mediumaquamarine102194165,
              fill opacity=0.0,
              draw opacity=1.0,
              mark options={fill=mediumaquamarine102194165, fill opacity=1.0, scale=0.5, line width=1pt},
              mark=*,
              only marks] coordinates {
        (2000, 0.289)
        (2000, 0.363)
        (3000, 0.408)
        (3000, 0.437)
        (3000, 0.459)
        (3000, 0.475)
        (3000, 0.488)
        (4000, 0.498)
        (4000, 0.507)
        (4000, 0.515)
    };

    \nextgroupplot[
    axis background/.style={fill=gainsboro229},
    axis line style={white},
    tick align=outside,
    tick pos=left,
    x grid style={white},
    align=center,
    title={ResNet-9 on CIFAR-2},
    xlabel={Computation time (mins) on 1xA100 \\ (\(\displaystyle \leftarrow\) more efficient)},
    xmajorgrids,
    xminorgrids,
    xtick style={color=dimgray85},
    xlabel={Number of Models},
    xmin=-5, xmax=110,
    error bars/y dir=both,
    error bars/y explicit,
    legend pos=north west,
    y grid style={white},
    ymajorgrids,
    ymin=0, ymax=0.7,
    ytick style={color=dimgray85}
    ]
    \addplot+[
        draw=mediumaquamarine102194165,
        thick,
        mark=star,
        mark options={draw=black, fill=black, fill opacity=0., draw opacity=0.},
        error bars/.cd,
        y dir=both,
        y explicit,
        error bar style={black} %
    ] coordinates {
        (1, 0.096450)    +- (0.00198164013630789, 0.00198164013630789)
        (3, 0.162533)    +- (0.00250681872192129, 0.00250681872192129)
        (5, 0.203170)    +- (0.00285016339769890, 0.00285016339769890)
        (10, 0.271743)   +- (0.00335648047362187, 0.00335648047362187)
        (20, 0.350416)   +- (0.00367534416506587, 0.00367534416506587)
        (30, 0.398201)   +- (0.00371800863084005, 0.00371800863084005)
        (40, 0.429772)   +- (0.00367970543552132, 0.00367970543552132)
        (50, 0.453237)   +- (0.00361153959465393, 0.00361153959465393)
        (60, 0.471247)   +- (0.00354877381531748, 0.00354877381531748)
        (70, 0.485665)   +- (0.00348402233077449, 0.00348402233077449)
        (80, 0.497404)   +- (0.00341587862141950, 0.00341587862141950)
        (90, 0.507326)   +- (0.00336738271801295, 0.00336738271801295)
        (100, 0.515456)  +- (0.00330819475979240, 0.00330819475979240)
    };
    \end{groupplot}
\end{tikzpicture}
\caption{
\textbf{Left:}
  {\em The impact of the dimension of random projection on \trak's
  performance on \cifartwo.} Each line corresponds to a different value of $M \in \{10,20,...,100\}$ (the number of models \trak is averaged over); darker lines correspond to higher $M$. As we increase the projected dimension, the LDS initially increases. However, beyond a certain dimension, the LDS
  begins to decrease. The ``optimal'' dimension (i.e., the peak in the above
  graph) increases with higher $M$.
\textbf{Right:}
  {\em The impact of ensembling more models on \trak's performance on \cifartwo.} The
  performance of \trak  as a function of the number of models used in the
  ensembling step. \trak scores are computed with
  random projections of dimension $k=4000$.
}
\label{fig:jl_dim}
\end{figure}

\subsection{Number of models used in the ensemble}
\label{app:ablation_num_models}
An important component of computing \trak is ensembling over multiple independently trained models (Step 4 in \cref{sec:estimator_algo}).  In our experiments, we average
\trak's attribution scores over ensembles of size ranging from $1$ to $100$.
Here, we quantify the importance of this procedure on \trak's performance.

\Cref{fig:jl_dim} (Right) shows that
\trak enjoys a significantly
better data attribution performance with more models.
That said,
even without ensembling (i.e., using a single model), \trak still performs
better (e.g., LDS of 0.096 on \cifartwo) than all prior gradient-based methods
that we evaluate.

\subsection{Proxies for model ensembles in compute-constrained settings}
\label{app:proxies}
In \cref{app:ablation_num_models} we saw that ensembling leads to significantly
higher efficacy (in terms of LDS).
In many settings, however, it is computationally expensive to
train several independent models to make an ensemble.
Hence, we study whether there is a cheaper alternative to
training multiple independent models that does not significantly sacrifice
efficacy.
To this end, we explore two avenues of approximating the full ensembling step while
dramatically reducing the time required for model training.
In particular, we investigate:
\begin{enumerate}
    \item using multiple checkpoints from each training trajectory;
    \item using checkpoints from early training, long before the model has
    converged.
\end{enumerate}

\paragraph{Multiple checkpoints from each training trajectory.}
We compute \trak scores
using a {\em fixed} number of checkpoints, but while varying the number of
independently-trained models.
For example, for 100 checkpoints, we can use the
final checkpoints from $100$ independently-trained models, the last two
checkpoints from $50$ independently-trained models, etc. We observe (see
\cref{tab:num_independent_runs_app}) that \trak achieves comparable LDS when we use
last $T$ checkpoints along the trajectory of the same models as a proxy for
independently-trained models in the ensembling step.

\paragraph{Using checkpoints from early training.}
We explore whether each of the models in the ensemble has to be fully trained to
convergence. In particular, we study the effect of using checkpoints from early
epochs on the LDS. While \trak benefits from using later-epoch gradient features, it
maintains its efficacy even when we use gradient features from training runs
long before reaching convergence (see \Cref{tab:ablation_epoch_used_app}). Leveraing this can further improve the computational efficiency of \trak.

\begin{table}[htbp]
    \centering
    \begin{minipage}{.46\textwidth}
      \centering
      \begin{tabular}{cc}
      \toprule
      \# training epochs & LDS ($M = 100$) \\
      \midrule
      1 & 0.100 \\
      5 & 0.204 \\
      10 & 0.265 \\
      15 & 0.293 \\
      25 & 0.308 \\
      \bottomrule
      \end{tabular}
      \caption{The performance of \trak on \cifarten as a function of the epoch at which we
      terminate model training. In all cases, \trak scores are computed with projection dimension $k = 1000$ and $M=100$ independently trained models.}
      \label{tab:ablation_epoch_used_app}
    \end{minipage}\hfill
    \begin{minipage}{.46\textwidth}
      \centering
      \begin{tabular}{cc}
      \toprule
      \# independent models & LDS \\
      \midrule
      5 & 0.329 \\
      6 & 0.340 \\
      10 & 0.350 \\
      100 & 0.355 \\
      \bottomrule
      \end{tabular}
      \caption{{\trak maintains its efficacy when we use multiple checkpoints from different epochs of the same training run instead of checkpoints from
      independently-trained models (\cifarten).} In all cases, $M=100$ checkpoints and projection dimension $k = 4000$ are used to compute \trak scores.}
      \label{tab:num_independent_runs_app}
    \end{minipage}
\end{table}

\subsection{Role of different terms.}
The \trak estimator (\Cref{eq:trak}) has a number of different components. We label each component (of the single model estimator) as follows:
\[
    \tau(z)_i = \frac{\phi(z)^\top \overbrace{(\Phi^\top R \Phi)^{-1}}^\text{reweighting} \phi(z_i) \cdot \overbrace{\frac{1}{1+e^{f(z_i)}}}^\text{loss gradient}}{1 - \underbrace{h_i}_\text{leverage score}}
\]

\noindent We ablate each of the terms above and re-evaluate the resulting variant of \trak on \cifartwo.
Our results in \Cref{tab:terms_ablate} indicate the following:
\begin{itemize}
\item {\bf Reweighting:} Experiment 6 shows that this matrix is a critical part of \trak's performance. Conceptually, this matrix distinguishes our estimator from prior gradient based similarity metrics such as \tracin.
\item {\bf Diagonal term $R$:} The full reweighting matrix includes a diagonal term $R$. Although it is theoretically motivated by \Cref{lem:formal}, including this term results in lower LDS, so we do not include it (Experiments 2,4).
\item {\bf Loss gradient}: This term corresponds to the $\mathbf{Q}$ matrix (\Cref{eq:q_mat}) and encodes the probability of the incorrect class, $1-p_i$; the name is based on the derivation in \Cref{app:theory_newton}, where this term corresponds to scalar associated with the gradient of the loss. Intuitively, this term helps reweight training examples based on on models' confidence on them.
 Experiment 5 shows that this term improves the performance substantially.
\item {\bf Leverage score:} This term does not impact the LDS meaningfully, so we do not include it (Experiments 1,2).
\item {\bf Averaging ``out'' vs ``in'':} Averaging the estimator and the loss gradient term separately, then re-scaling by the average loss gradient results in higher LDS (Experiment 3).
\end{itemize}

\begin{table}[!bht]
  \begin{tabular}{llllllr}
      \toprule
      Experiment & Reweighting & Loss &   Diagonal $R$ & Leverage & Averaging &  Correlation \\
      \midrule
      0 &         \checkmark &  \checkmark &   \xmark    &      \xmark    &     out &        0.499 \\
      1 &         \checkmark &  \checkmark &   \xmark    &      \checkmark &     out &        0.499 \\
      2 &         \checkmark &  \checkmark &  \checkmark &      \checkmark &     out &        0.430 \\
      3 &         \checkmark &  \checkmark &   \xmark    &       \xmark    &      in &        0.416 \\
      4 &         \checkmark &  \checkmark &  \checkmark &       \xmark    &     out &        0.403 \\
      5 &         \checkmark &   \xmark    &   \xmark    &       \xmark    &     out &        0.391 \\
      6 &          \xmark    &  \checkmark &   \xmark    &       \xmark    &     out &        0.056 \\
      \bottomrule
      \end{tabular}
  \centering
  \caption{{\em Ablating the contribution of each term in the \trak estimator.} For these experiments, we use random projections of dimenseion $k=2000$.}
  \label{tab:terms_ablate}
\end{table}

\subsection{Choice of the kernel}
To understand how the choice of the kernel impacts the performance of \trak,
we also compute a version of \trak using feature representations of the penultimate layer in place of the projected gradients.
This choice is equivalent to restricting the gradient features to those of the last linear layer.
As \Cref{tab:kernel_choice} shows, this method significantly improves on all existing baselines based on gradient approximations,\footnote{Note that as with the eNTK, the use of multiple models here is crucial: only using a single model gives a correlation of 0.006.} but still underperforms significantly relative to \trak. This gap suggests that the eNTK is capturing additional  information that is not captured by penultimate layer representations.
Moreover, the larger gap on \cifarten compared to \cifartwo and \qnli (both of which are binary classificaiton tasks) hints that the gap will only widen on more complex tasks.

We note that \trak applied only to the last layer is almost equivalent to the influence function approximation. Indeed, they perform similarly (e.g., the influence function approximation also achieves a LDS of 0.19 on \qnli).

\begin{table}[h]
    \centering
    \begin{tabular}{lrr}
        \toprule
              Dataset &  Kernel representation &  Linear Datamodeling Score (LDS) \\
        \midrule
          \cifartwo &  eNTK  &     {\bf 0.516} \\
          \cifartwo &  penultimate layer   &     0.198 \\
        \midrule
          \cifarten & eNTK &     {\bf 0.413} \\
          \cifarten & penultimate layer   &     0.120 \\
        \midrule
          \qnli & eNTK & {\bf 0.589} \\
          \qnli & penultimate layer & 0.195  \\
        \bottomrule
    \end{tabular}
    \caption{{\em Choice of the kernel in \trak}. We compare \trak computed using the eNTK (i.e., using features derived from full gradients) with \trak computed using the kernel derived from last layer feature representations. The attribution scores are ensembled over $M=100$ models.}
    \label{tab:kernel_choice}
    \end{table}

\subsection{Ensembling vs. Averaging the eNTK}
There are different ways to ensemble a kernel method given multiple kernels $\{K_i\}_i$: (i) we can average the Gram matrices corresponding to each kernel first and then predict using the averaged kernel (i.e., work with $\overline{K} = \frac{1}{n} \sum K_i$),  (ii) we can average their induced features (with respect to some fixed basis of functions) and use the corresponding kernel, or (iii) we can average the predictions derived from each kernel \citep{atanasov2023onset}.
\trak's algorithm follows the third approach (Step 4).

Here we ensemble using the first approach instead (i.e., using the averaged eNTK). We do this by first averaging the Gram matrices corresponding to each models' eNTK, using the Cholesky decomposition to extract features from the averaged Gram matrix ($G=LL^\top$), then using resulting features $L$ into the same influence formula (Step 3).
We find that computing \trak with this average eNTK gives a significantly underperforming estimator  (LDS of 0.120 on \cifartwo) than averaging {\em after} computing the estimator from each eNTK (LDS of 0.499).
This gap suggests that the underlying model is better approximated as an ensemble of kernel predictors rather than a predictor based on a single kernel.

\subsection{Summary}
To summarize the results of our ablation, \trak performs best when averaging over a
sufficient number of models (though computationally cheaper alternatives also work); gradients computed at later epochs; and random
projections to sufficiently high---but not too high---dimension. Using the reweighting matrix in \cref{eq:trak}, as well as deriving
the features from the full model gradient are also both critical to \trak's
predictive performance.

\clearpage
\section{Fact Tracing}
\label{app:lexical}
\subsection{The \ftracetrex Dataset}
\label{app:fact_tracing:dataset}
The training set of \ftracetrex is sourced from the \trex dataset \citep{elsahar2018t},
with each training example excerpted from a DBPedia abstract \citep{hellmann2013integrating} and
annotated with a list of facts it expresses.\footnote{See \citep{akyurek2022towards}
for more details on the annotation methodology.}
The test set of \ftracetrex is sourced from the \texttt{LAMA} dataset \citep{petroni2019language},
and each test example is a sentence that expresses a single fact---every training
example that expresses the same fact is called a ``proponent'' of this test example.
Now, given a test example expressing some fact,
the goal of fact tracing (as defined by the \ftracetrex benchmark)
is to correctly identify the corresponding
proponents from the training set.

More precisely, \citet{akyurek2022towards} propose the following evaluation methodology,
which we follow exactly
(with the exception that, due to computational constraints,
we use a smaller 300M-parameter \texttt{mt5-small} model
instead of the 580M-parameter \texttt{mt5-base}).
We first finetune the pretrained language model \cite{raffel2020exploring}
on the training set of \ftracetrex.
Then, we iterate through the \ftracetrex test set and find the examples
on which the pre-trained model is incorrect
and the finetuned model is correct,\footnote{
    To decide whether a model is ``correct'' on a given test example,
    we use MT5 as a conditional generation model. That is, we
    feed in a masked version of the query, e.g.,
    ``\textunderscore\textunderscore\ is the capital of France,''
    and mark the model as ``correct'' if the conditional generation
    matches the masked word.
} which \citet{akyurek2022towards} refer to as the ``novel facts'' learned by the model after finetuning.
For each novel fact identified,
we collect a set of candidate training examples,
comprising all proponents as well as
300 ``distractors'' from the training set.
\citet{akyurek2022towards} propose to evaluate different attribution methods
based on how well they identify the ground-truth proponents among each candidate
set.

Concretely, given an attribution method $\tau(\cdot)$,
we compute attribution scores $\tau(z)$ for each of the
novel facts in the test set.
For each novel fact, we sort the corresponding candidate examples by their
score $\tau(z)_i$. Finally, we compute the mean reciprocal rank (MRR), a standard
information retrieval metric, of ground-truth proponents
across the set of novel facts, defined as
\[
    \text{MRR} = \sum_{z \in \parbox{1cm}{\tiny novel \\ facts}}\frac{1}{\min\limits_{i\, \in\, \text{proponents}(z)} \text{rank}(\tau(z), i)}.
\]

\subsection{Fine-tuning details}
\label{app:fact_tracing:loss}
We finetune the pre-trained language model using the masked language modeling
objective \citep{devlin2019bert}.
In particular, for each training example $z_i \in [K]^L$
(where $K$ is the vocabulary size and $L$ is the maximum passage length),
we mask out a subject or object within the passage.
(E.g., a training example ``Paris is the capital of France'' might become an
input-label pair [``\textunderscore\textunderscore\ is the capital of France'',
``Paris'']).
We then treat the language modeling problem as multiple separate
$K$-way classification tasks.
Each task corresponds to predicting a single token of the masked-out text,
given (as input) the entire passage minus the token being predicted.
The loss function is the average cross-entropy loss on this sequence of
classification tasks.

\subsection{Computing \trak for masked language modeling}
\label{app:fact_tracing:trak}
The model output function we use, more precisely, is given by:
\[
    \modeleval{z}{\theta} = \!\!\!\!\!\!\sum\limits_{\ \ \ j\ \in\ \parbox{0.7cm}{\centering\tiny masked \\ tokens}}
    \log \left(\frac{p(z^{j}|z^{-j};\theta)}{1-p(z^{j}|z^{-j};\theta)}\right).
\]
In particular, to compute this model output function,
we compute the model output function \eqref{eq:modelout_mc}
for each one of the $V$-way classification problems separately, then define
our model output function as the sum of these computed outputs.

\subsection{Counterfactual experiment setup}
\label{app:fact_tracing:cfx}
To understand the possible roots of \trak's underperformance
relative to BM25 on \ftracetrex,
we carry out a counterfactual analysis.
Specifically, for a subset of the \ftracetrex test set, we
create three corresponding {\em counterfactual training sets}.
Each training set corresponds to \underline{removing} one of three collections of examples
from the \ftracetrex training set:
\begin{enumerate}
    \item[(a)]
    the union (across all 50 selected novel facts) of the 500 most important
    training examples for each novel fact, as identified by \trak
    (this corresponds to removing $17,914$ total training examples, leaving $1,542,539$ remaining);
    \item[(b)]
    the union of the 500 most important
    training examples for each novel fact, as identified by BM25
    ($18,146$ total examples removed, and $1,542,307$ remaining);
    \item[(c)]
    the union of the proponents---as defined by \ftracetrex---for
    each novel fact ($10,780$ examples removed, and $1,549,673$ remaining)
\end{enumerate}
Then, starting from a pre-trained \texttt{mt5-small} model (the same model that we
finetuned in (B) above to identify novel facts),
we finetune several models on each counterfactual training set, and compute
their average accuracy on the selected subset of 50 novel facts.
Note that, by construction, we know that on this subset
(i) the pre-trained model has an
accuracy of 0\%; and (ii) finetuning on the
entire \ftracetrex training set (i.e., with no examples removed)
yields models with 100\% accuracy.\footnote{
    In particular, recall that in order for a test example to be categorized as a
    ``novel fact,'' it must be both (a) incorrectly handled by the pre-trained
    \texttt{mt5-small} model and (b) correctly handled by a finetuned model.
}
As for the counterfactual training sets, one should note that:
\begin{itemize}
    \item Counterfactual training set (c) is missing all of the
    proponents for our subset of 50 novel facts---we would thus expect the
    corresponding finetuned model to have very low accuracy.
    In particular, there is ostensibly no direct evidence for {\em any} of the novel facts
    of interest anywhere in this counterfactual training set.
    \item Being constructed with BM25, counterfactual training set (b)
    has high lexical overlap with the novel facts of interest.
    Since BM25 performs well on the \ftracetrex benchmark,
    we would also expect the resulting models to have low accuracy.
\end{itemize}
In \cref{fig:nlp_counterfactual}, we report the resulting models' average performance on
the set of 50 selected novel facts.
What we find is that, counter to the above intuition,
{\em only the} \trak-{\em based counterfactual training set is able to
significantly change model behavior}.
That is, the counterfactual effect of removing the most important images
as identified by \trak on the selected subset of novel facts
is significantly higher than both
(a) that of removing the most important images according to BM25; and
(b) that of removing the {\em ground-truth proponents} of the facts as indicated
by the \ftracetrex benchmark.

\clearpage
\section{Future Work}
\label{app:future_work}
\subsection{Further applications of \trak}
Prior works have demonstrated the potential of leveraging data attribution for a
variety of downstream applications, ranging from explaining predictions
\cite{koh2017understanding,kong2022resolving}, cleaning datasets
\cite{jia2019towards}, removing poisoned examples \cite{lin2022measuring} to quantifying uncertainty \cite{alaa2020discriminative}.
Given the effectiveness of \trak, we expect that using it in place of existing attribution
methods will improve the performance in many of these downstream applications.
Moreover, given its computational efficiency, \trak can expand the settings in
which these prior data attribution methods are feasible. Indeed, we already saw some examples in \Cref{subsec:datamodel_apps}. We highlight a few promising
directions in particular:

\paragraph{Fact tracing and attribution for generative models.}
Fact tracing, which we studied in \Cref{subsec:fact_trace}, is a problem of increasing relevancy as large language models are widely deployed. Leveraging \trak for fact tracing, or attribution more broadly, may
help understand the capabilities or improve the trustworthiness of recent models such as GPT-3
\cite{brown2020language} and ChatGPT,\footnote{\url{https://chat.openai.com/}}
by tracing their outputs back to sources in a way that is faithful to the actual model.
More broadly, attribution for generative models (e.g., stable diffusion \cite{ho2020denoising,rombach2022high}) is an interesting direction for future work.

\paragraph{Optimizing datasets.}
\trak scores allow one to quantify the impact of individual training examples on model predictions on a given target example.
By aggregating this information, we can optimize what data we train the models on,
for instance, to choose {\em coresets} or to select new data for {\em active learning}.
Given the trend of training models on ever increasing size of datasets \cite{hoffmann2022training}, filtering data based on their \trak scores can also help models achieve with the benefits of scale without the computational cost.

Another advantage of \trak is that it is fully differentiable in the input (note that the associated gradients are different from the gradients with respect to model parameters that we use when computing \trak).
One potential direction is to leverage this differentiability for {\em dataset distillation}. Given the effectiveness of the NTK for this problem \cite{nguyen2021dataset}, there is potential in leveraging \trak---which uses the eNTK---in this setting.

\subsection{Understanding and improving the \trak estimator}

\paragraph{Empirical NTK.}
\trak leverages the empirical NTK to approximate the original model. Better understanding of when this approximation is accurate may give insights into improving \trak's efficacy.
For example, incorporating higher order approximations \cite{huang2020dynamics,bai2020beyond} beyond the linear approximation used in \trak is a possible direction.

\paragraph{Training dynamics and optimization.}
Prior works \citep{leclerc2020two,lewkowycz2020large} suggest that neural network training can exhibit two stages or regimes: in the first stage, the features learned by the network evolve rapidly; in the second stage, the features remain approximately invariant and the overall optimization trajectory is more akin a convex setting. We can view our use of the final eNTK as modeling this second stage. %
Understanding the extent to which the first stage (which \trak does not model) accounts for the remaining gap between true model outputs and \trak's predictions may help us understand the limits of our method as well as improve its efficacy.
Another direction is to study whether properly accounting for other optimization components used during training, such as mini-batches, momentum, or weight decay, can improve our estimator.

\paragraph{Ensembles.} As we saw in \Cref{app:ablation_num_models}, computing \trak
over an ensemble of models significantly improves its efficacy.  In particular, our results suggest that the eNTK's derived from
independently trained models capture non-overlapping information. %
Better understanding of the role of ensembling here may
us better understand the mechanisms underlying ensembles in other contexts
and can also provide practical insights for improving \trak's efficiency. For
instance, understanding when model checkpoints from a single trajectory can
approximate the full ensemble (\Cref{app:proxies}) can be valuable in settings where it is expensive to
even finetune several models.

\end{document}